%% file: main.tex
\documentclass[10pt]{article} 
\usepackage[preprint]{tmlr}

\input{math_commands.tex}

\usepackage{graphicx}
\usepackage{hyperref}
\usepackage{url}
\usepackage{booktabs}
\usepackage{bbm}
\usepackage{multirow}
\usepackage{natbib}
\usepackage{adjustbox}
\usepackage{makecell}
\usepackage{amsthm}
\usepackage{tcolorbox}
\tcbuselibrary{most}
\usepackage{wrapfig}
\usepackage{tabularx}
\usepackage{ragged2e}
\usepackage{threeparttable}
\usepackage{makecell}
\usepackage{enumitem}
\usepackage{caption}
\usepackage{subcaption}
\usepackage{adjustbox}
\usepackage{lmodern}
\usepackage{microtype}
\usepackage{natbib}
\usepackage{cleveref}
\usepackage{multirow}
\usepackage{enumitem}
\usepackage[table]{xcolor}
\usepackage{siunitx}
\usepackage{array}

\hypersetup{
    colorlinks=true,
    linkcolor=blue,
    citecolor=blue,
    urlcolor=blue
}

\newtcolorbox{keytakeaway}[1][]{%
  enhanced,
  breakable,
  colback=blue!5!white,
  colframe=blue!75!black,
  boxrule=0.6pt,
  arc=2pt,
  left=4pt,
  right=4pt,
  top=4pt,
  bottom=4pt,
  fonttitle=\bfseries,
  coltitle=black,
  #1
}

\newcolumntype{L}[1]{>{\raggedright\arraybackslash}p{#1}}
\newcolumntype{C}[1]{>{\centering\arraybackslash}p{#1}}

\title{Discovery and Spatial Characterisation of Multiple Shortcut Groups for Auditing Vision Model Bias}

\author{\name Akshit Achara \email akshit.achara@kcl.ac.uk \\
      \\
      \name Vishnunarayan Manickam \email vishnunarayan.manickam@kcl.ac.uk \\
      \\
      \name Thomas Day \email thomas.day@kcl.ac.uk \\
      \\
      \name Esther Puyol Anton \email esther.puyol\_anton@kcl.ac.uk \\
      \\
      \name Alexander Hammers \email alexander.hammers@kcl.ac.uk\\
      \\
      \name Andrew P. King \email andrew.king@kcl.ac.uk \\ \\
      \addr School of Biomedical Engineering and Imaging Sciences,\\ King's College London, UK}

\newcolumntype{Y}{>{\RaggedRight\arraybackslash}X}

\def\month{MM}  
\def\year{YYYY} 
\def\openreview{\url{https://openreview.net/forum?id=XXXX}} 

\begin{document}

\maketitle

\begin{abstract}
Deep learning models trained on datasets with spurious correlations can achieve high average accuracy whilst relying on shortcut features that do not generalise out of distribution. Whilst out-of-distribution testing can help in highlighting subgroup performance disparities arising from shortcut learning, it does not localise the regions within images that are associated with it. Existing research mostly utilises attribution maps from interpretability methods for understanding the spatial nature of spurious correlations. For example, conditional alignment methods separate task-relevant evidence from evidence tied to spurious correlations by comparing attribution maps from a task model, a sensitive attribute model, and a bias-reduced reference model. This yields shortcut-aligned and task-aligned contribution maps for each image. However, existing methods aggregate these maps across the dataset, potentially masking recurring spatial shortcut patterns that occur only in subsets of images. We address this limitation by grouping per-image shortcut and task contribution maps into recurring spatial patterns using K-means and non-negative matrix factorisation, and visualising the resulting shortcut groups through contribution maps and representative examples. Across \textsc{CelebA}, \textsc{CheXpert}, \textsc{Waterbirds}, \textsc{Camelyon17}, and \textsc{ISIC2019}, and across ResNet and ViT models, the discovered shortcut groups reveal both shared and distinct spatial patterns of shortcut and task contribution, with varying subgroup composition and error rates, thereby enabling targeted inspection of image subsets with higher error rates. We perform input occlusion and internal test-time interventions to demonstrate that masking or suppression of task contribution regions substantially degrades the model classification performance and propose a combined shortcut suppression and task amplification feature intervention approach which generally reduces performance disparities.
\end{abstract}

\input{sections/introduction}

\input{sections/related_work}

\input{sections/method}

\input{sections/datasets}

\input{sections/experiments}

\input{sections/discussion}

\input{sections/conclusion}

\input{sections/acknowledgements}

\clearpage

\bibliography{main}
\bibliographystyle{tmlr}

\clearpage

\appendix

\input{sections/appendix}

\end{document}

%% file: math_commands.tex
\usepackage{amsmath,amsfonts,bm}

\def\eqref#1{equation~\ref{#1}}

\def\1{\bm{1}}

\DeclareMathAlphabet{\mathsfit}{\encodingdefault}{\sfdefault}{m}{sl}
\SetMathAlphabet{\mathsfit}{bold}{\encodingdefault}{\sfdefault}{bx}{n}



%% file: sections/introduction.tex
\section{Introduction}
\label{sec:introduction}

Computer vision models can achieve high average accuracy while remaining brittle under distribution shift~\citep{geirhos2020shortcut}. A common cause is shortcut learning under spurious correlations, where feature(s) present in the image are predictive of the target label in the training distribution but do not generalise at test time within particular subpopulations. In this case, a model can obtain high training accuracy by using cues linked to the shortcut feature(s) as opposed to evidence that supports the intended task. If and when the correlation changes at test time, the model can suffer from failures that are often concentrated in particular subpopulations.

While out-of-distribution testing can help reveal such shortcut-driven failures~\citep{geirhos2020shortcut}, practitioners often also want to identify the image regions associated with the shortcut features. For instance, in datasets such as \textsc{Waterbirds}~\citep{sagawa2019distributionally}, the spurious features are related to the background as intended during the creation of the dataset. However, shortcut learning is also relevant in more realistic tasks such as in medical imaging, where acquisition and workflow artefacts can correlate with clinical labels~\citep{winkler2019association,degrave2021ai,kushol2023effects,brown2023detecting}. Furthermore, if the shortcut is associated with a demographic attribute, the failures can manifest as biased performance against underrepresented groups~\citep{achara2025invisible}. Shortcut features linked to demographic bias can be diffuse and/or difficult to localise to consistent and specific image regions across an entire dataset \citep{gichoya2022ai, lee2025investigation}. In such settings, an auditor often wants to know: what shortcut cues is the model using, which image subgroups do they affect, where do they appear in the image, and how do they relate to task-relevant evidence?

Attribution-based approaches offer a natural way to inspect where a model finds evidence for its decisions. For example, Spectral Relevance Analysis (SpRAy)~\citep{lapuschkin2019unmasking} clusters downsampled Layer-wise Relevance Propagation (LRP)~\citep{binder2016layer} attribution maps to identify recurring explanation patterns across a dataset. This characterises heterogeneity in attribution maps and can reveal Clever Hans behaviour after manual inspection, but it does not define shortcut-specific or actionable information that can be used to address shortcut learning. Other work on demographic shortcut learning used ranked regional attributions to analyse alignment between task and sensitive attribute models~\citep{achara2025invisible}. OSCAR~\citep{achara2025localising} generalises this rank-based analysis by comparing dataset-level regional rank profiles from an audited task model, a sensitive attribute model, and a bias-reduced reference model, using conditional correlations to quantify shortcut and task alignment and localise their region-level contributions. However, OSCAR aggregates image-level rankings across the dataset before computing its scores and contribution maps, yielding dataset-level contribution maps. This can obscure cases in which shortcut regions vary between multiple image subgroups or interact differently with task-relevant evidence. In this work, we build upon OSCAR and compute image-level shortcut contribution maps and task contribution maps, and then group their joint shortcut and task contributions to obtain recurring patterns within a single dataset.

Briefly, in our method each attribution map is partitioned into mutually exclusive spatial regions and converted into a within-image ranking of regional attribution values; conditional alignment of these ranked representations then gives image-level shortcut and task alignment scores and their region-level contributions, as described in Sections~\ref{sec:method-lowres} and~\ref{sec:method-contribution-maps}.  We form image-level shortcut and task contribution maps along with a joint shortcut and task contribution representation for each image, as described in Section~\ref{sec:method-patterns}. Following this, we group these image-level joint shortcut and task contribution representations into recurring spatial patterns. We refer to these groups as \emph{shortcut groups}.

We then ask whether the shortcut group-level contribution maps are useful for auditing. Specifically, we study (i) whether shortcut group membership can help in identifying high-risk subsets for inspection from the overall test set, and (ii)
whether shortcut contribution maps and task contribution maps provide actionable spatial evidence for test-time intervention (Section~\ref{sec:method-interventions}).

\paragraph{Contributions.}
We make the following contributions:

\textbf{Spatial shortcut pattern discovery from image-level contribution maps.}
We compute image-level shortcut and task contribution maps and group their joint spatial representations using K-means and non-negative matrix factorisation (NMF). The resulting analysis includes shortcut group prototypes and representative examples across five datasets and two architectures, together with broader prototype collections across attribution methods, grouping approaches, and number of shortcut groups (\Cref{fig:regime_exemplars_vit_celeba,fig:chexpert_artifact_exemplars}; Appendix Section~\ref{sec:supp-contribution-map-figures}). These results show that several coherent spatial shortcut patterns, with different task contribution, subgroup composition, and error rates, can coexist within a dataset and are not represented by a single dataset-level map.

\textbf{Failure analysis and targeted auditing with shortcut groups.}
We test whether shortcut group membership of images/examples can help in identifying high-risk subsets for inspection. Across datasets and models, the shortcut group membership based risk score can be used to identify subsets of images that contain a large proportion of the audited model's total test errors (\Cref{fig:supp_error_strat_heldout}; Appendix
Sections~\ref{sec:supp-grid-var},~\ref{sec:supp-k},
and~\ref{sec:supp-inspection-budget}).

\textbf{Test-time spatial intervention study across architectures.}
To close the loop between spatial characterisation and prediction, we use input masking and internal test-time interventions to determine whether the identified shortcut and task regions contain evidence relevant to the predictions. Masking or suppressing task contribution regions degrades classification performance, whereas combined shortcut suppression and task amplification reduces subgroup performance disparities. Shortcut group-level maps provide the most effective intervention guidance across most datasets and architectures (\Cref{tab:input_occlusion_summary,tab:intervention_source}; Appendix Section~\ref{sec:supp-dataset-vs-other-level}).

%% file: sections/related_work.tex
\section{Related work}

\paragraph{Spurious correlations and shortcut learning.}
Models can exploit spurious signals that correlate with the label in the training distribution, leading to failure of generalisation when these correlations change at test time~\citep{geirhos2020shortcut,sagawa2019distributionally}. This behaviour is also a concern in real-world domains such as medical imaging, where acquisition or workflow artefacts can correlate with clinical labels~\citep{winkler2019association,degrave2021ai,brown2023detecting,kushol2023effects}. Methods have been proposed for detecting and quantifying shortcut learning \citep{brown2023detecting} but lack actionable information to reduce it. Other methods aim to reduce shortcut reliance during training, including reweighting~\citep{nam2020learning,liu2021just}, robust optimisation~\citep{sagawa2019distributionally}, invariance objectives~\citep{arjovsky2019invariant}, and data augmentation~\citep{yun2019cutmix}. Such methods come with unavoidable computational overheads due to the retraining required. In this work, we study fixed trained models and ask how shortcut reliance varies across images, and whether spatial summaries of this variation can support auditing and test time intervention.

\paragraph{Subgroup, slice, and failure set discovery.}
A related line of work identifies vulnerable subsets, or slices, where models underperform. These methods use signals such as model errors and target labels~\citep{eyuboglu2022domino,sohoni2020no}, available metadata or feature values~\citep{sagadeeva2021sliceline}, or unsupervised partitions in representation space~\citep{bissoto2025subgroup}. Such methods answer the question of which examples form a high-risk subset. However, they are not aimed at providing evidence of the shortcut features the model is using and where that evidence appears in the image. A discovered slice may identify a vulnerable subset without revealing whether its failures are associated with one spatial shortcut cue, several recurring shortcut cues, or the absence of task-supporting evidence. Additionally, slice discovery is primarily useful for identifying different slices of a dataset with unknown attributes, whereas in this work we focus on spatially characterising shortcuts using known sensitive attributes.

\paragraph{Attribution-based auditing of spurious reliance.}
Attribution methods~\citep{selvaraju2017grad,binder2016layer,chattopadhay2018grad,pmlr-v235-achtibat24a} are widely used to inspect which image regions support a model prediction. Early work on (partially) automating the analysis of attribution maps to audit models for shortcuts involved clustering of maps followed by human inspection~\citep{lapuschkin2019unmasking}. More recent work has gone further, automatically localising and quantifying shortcut reliance by comparing task model attributions with attributions from sensitive attribute or shortcut predictors~\citep{achara2025localising}. However, this approach aggregates attribution evidence across the dataset, which assumes that shortcut contribution is spatially consistent across examples. This can hide cases where shortcut features vary by location, co-occur with different task features, and/or appear only in particular subsets of images.

\paragraph{From local explanations to structured summaries.}
Prior work has used concepts, patches, and prototypes to give local explanations more structure. TCAV measures sensitivity to user defined concepts~\citep{kim2018interpretability}, concept bottleneck models predict through annotated concepts~\citep{koh2020concept}, ACE discovers concepts from image segments~\citep{ghorbani2019towards}, and prototype based models compare image regions with learned prototypical parts~\citep{chen2019looks}. These approaches are valuable when concepts, annotations, segments, or prototype based architectures are available. They are less useful for auditing a fixed trained model when the relevant shortcut cues are unknown and may vary across images. We instead use attribution based conditional alignment to obtain spatial contribution maps, which are then grouped to visualise recurring spatial patterns. Unlike general concept discovery or representation interpretation methods~\citep{ghorbani2019towards,zhou2018interpreting}, our analyses describe recurring spatial structures in shortcut and task contribution and are evaluated through representative examples, subgroup composition, error rates, and spatial interventions.

\paragraph{Spatial interventions and shortcut mitigation.}
Many shortcut mitigation methods act during training or retraining, including reweighting~\citep{nam2020learning}, distributionally robust optimisation~\citep{sagawa2019distributionally}, invariant learning~\citep{arjovsky2019invariant}, model retraining~\citep{liu2021just}, evidential alignment~\citep{ye2025improving}, and subspace-based correction~\citep{zheng2025shortcutprobe}. Other fairness interventions operate after training at the output level, for example through post-hoc thresholding for equality of opportunity~\citep{hardt2016equality} or reject option classification~\citep{kamiran2012decision}. In contrast, we study test-time spatial interventions inside fixed trained vision models, asking whether shortcut and task contribution maps identify regions that can change model performance when acted on. This connects to broader work on actionable interpretability~\citep{orgad2026interpretability} and to prior work showing that attribution maps can guide interventions~\citep{achara2025localising}. Existing work, however, leaves open how attribution-derived shortcut and task evidence can guide test-time intervention in fixed vision models, including whether to intervene in input or feature space, where within the architecture to apply them, and how the spatial evidence and intervention effects change over iterative interventions. We examine these questions through input masking, feature-space shortcut suppression, and task amplification across convolutional and transformer architectures.

%% file: sections/method.tex
\section{Method}
\label{sec:method}

We describe the method in four steps. First, similar to \cite{achara2025localising}, we convert attribution maps from three models into ranked vectors (Section~\ref{sec:method-lowres}). The three models are the audited task model ($f_{\mathrm{TS}}$), the bias-reduced baseline/reference model ($f_{\mathrm{BA}}$), and the sensitive attribute model ($f_{\mathrm{SA}}$). $f_{\mathrm{BA}}$ is trained to predict the target label using a training set that is approximately balanced by the sensitive attribute being tested for possible shortcuts. $f_{\mathrm{SA}}$ uses the same training set as $f_{\mathrm{BA}}$ but is trained to predict the sensitive attribute. Second, we use the ranked vectors of these three models to compute partial correlation-based conditional alignment and obtain image-level shortcut contribution maps and task contribution maps (Sections~\ref{sec:method-alignment},~\ref{sec:method-contribution-maps}). Intuitively, the hypothesis is that, if $f_{\mathrm{TS}}$ uses image regions associated with the sensitive attribute to predict the target label (i.e. it uses shortcut learning), then after removing the spatial structure explained by $f_{\mathrm{BA}}$, its ranked attribution vector should be more aligned with that of $f_{\mathrm{SA}}$. We quantify this residual agreement using the shortcut alignment partial correlation score.
Conversely, if $f_{\mathrm{TS}}$ uses task-relevant regions and is shortcut free, then after removing the spatial structure explained by $f_{\mathrm{SA}}$, its ranked attribution vector should be more aligned with that of  $f_{\mathrm{BA}}$. We quantify this residual agreement using the task alignment partial correlation score.
Third, we form a joint shortcut and task contribution representation and summarise recurring spatial structure using hard grouping and soft mixture approaches by grouping the image-level contribution maps (Section~\ref{sec:method-patterns}). Finally, we introduce the input and feature level intervention approaches (Section~\ref{sec:method-interventions}).

\subsection{Preliminaries}
\label{sec:preliminaries}

First, we summarise the preliminary shortcut and task alignment computation based on existing work~\citep{achara2025localising}.

\subsubsection{From attributions to low-resolution spatial vectors}
\label{sec:method-lowres}

Let $x$ be an input image, $f$ a neural network, and $\hat{y}$ the prediction used as the attribution target. Given an attribution method $\mathcal{L}$, we obtain a nonnegative attribution map $\mathcal{L}(f,x,\hat{y})\in\mathbb{R}^{H\times W}_{+}$, interpolated to the image dimensions.

We partition the image into $R$ mutually exclusive blocks, defined by a fixed $G\times G$ grid with
$R=G^2$ regions $\{\Omega_r\}_{r=1}^R$, and compute the mean attribution over all the pixels $p$ in each region:
\begin{align*}
s_r(f,x,\hat{y})
=
\frac{1}{|\Omega_r|}
\sum_{p\in\Omega_r}
\mathcal{L}(f,x,\hat{y})_p .
\end{align*}
We then rank the $R$ regional scores within the image:
\begin{align}
\mathbf{s}(f,x,\hat{y})
=
\operatorname{rank}
\left(
(s_r(f,x,\hat{y}))_{r=1}^R
\right)
\in\mathbb{R}^R .
\label{equation:rankedattrvector}
\end{align}

The result is a low-resolution spatial attribution vector for each image, with one ranked attribution value per grid region. In OSCAR~\citep{achara2025localising}, the ranked vectors are aggregated across images before dataset-level shortcut and task alignment are computed. In contrast, we keep the ranked vectors at the image level. This preserves image-to-image variation in shortcut and task evidence and provides the input representation for grouping recurring spatial shortcut patterns in Section~\ref{sec:method-patterns}.

\subsubsection{Shortcut and task alignment from three attribution views}
\label{sec:method-alignment}

Following OSCAR~\citep{achara2025localising}, given an input $x$, we perform inference on $f_{\mathrm{TS}}$,  $f_{\mathrm{SA}}$, and $f_{\mathrm{BA}}$ and compute the low-resolution spatial attribution vectors as described in Eq. (\ref{equation:rankedattrvector}) for each test set image:
\begin{align*}
\mathbf{u}=\mathbf{s}(f_{\mathrm{TS}},x,\hat{y}_{\mathrm{TS}}),\qquad
\mathbf{v}=\mathbf{s}(f_{\mathrm{SA}},x,\hat{y}_{\mathrm{SA}}),\qquad
\mathbf{w}=\mathbf{s}(f_{\mathrm{BA}},x,\hat{y}_{\mathrm{BA}}),
\end{align*}
where $\hat{y}_{\mathrm{TS}},\hat{y}_{\mathrm{SA}},\hat{y}_{\mathrm{BA}}$ are the predicted labels used as attribution targets.

Now departing from OSCAR’s dataset-level formulation, to measure shortcut-aligned attribution, we remove the linear effect of the baseline model vector $\mathbf{w}$ from both the audited task model vector $\mathbf{u}$ and the sensitive attribute model vector $\mathbf{v}$, using ordinary least squares with an intercept. Let $\mathbf{r}_{u\mid w}\in \mathbb{R}^{R}$ and $\mathbf{r}_{v\mid w} \in \mathbb{R}^{R}$  denote the resulting residuals. We define the shortcut alignment score as
\begin{align}
\rho_{\mathrm{sc}}(x)
=
\mathrm{corr}
\left(
\mathbf{r}_{u\mid w},
\mathbf{r}_{v\mid w}
\right).
\label{eq:rho_sc}
\end{align}
A high value of $\rho_{\mathrm{sc}}(x)$ means that the audited task model and the sensitive attribute model emphasise similar spatial regions after accounting for the baseline model.

We also compute a task alignment score by comparing the vector for the audited task model with that of the baseline model while controlling for the sensitive attribute model. Let $\mathbf{r}_{u\mid v}$ and $\mathbf{r}_{w\mid v}$ be the residuals obtained after regressing $\mathbf{u}$ and $\mathbf{w}$ on $\mathbf{v}$. We define
\begin{align}
\rho_{\mathrm{task}}(x)
=
\mathrm{corr}
\left(
\mathbf{r}_{u\mid v},
\mathbf{r}_{w\mid v}
\right).
\label{eq:rho_task}
\end{align}
This captures spatial attribution structure shared by the audited task model and the baseline model that is not explained by the sensitive attribute model.

\subsubsection{Region contribution maps}
\label{sec:method-contribution-maps}

To localise which regions support the shortcut and task scores, we use a region-wise contribution decomposition. Let $\mathbf{r}_1,\mathbf{r}_2\in\mathbb{R}^R$ denote the two residual vectors whose correlation defines an alignment score. For shortcut alignment these are $\mathbf{r}_{u\mid w}$ and $\mathbf{r}_{v\mid w}$, whilst for task alignment they are $\mathbf{r}_{u\mid v}$ and $\mathbf{r}_{w\mid v}$.
We standardise each residual vector across the $R$ regions. For
$j\in \{1,2\}$, we define
\begin{align*}
\mu_j=\frac{1}{R}\sum_{i=1}^{R} r_{j,i},
\qquad
\sigma_j=
\left(
\frac{1}{R-1}\sum_{i=1}^{R}(r_{j,i}-\mu_j)^2
\right)^{1/2},
\qquad
z_{j,i}=\frac{r_{j,i}-\mu_j}{\sigma_j}.
\end{align*}

The region contribution score for grid cell $i$ is then
\begin{align*}
c_i(x)=z_{1,i}z_{2,i},
\qquad i=1,\ldots,R .
\end{align*}

The corresponding correlation can be recovered by summing these contributions:
\begin{align*}
\mathrm{corr}(\mathbf{r}_1,\mathbf{r}_2)
=
\frac{1}{R-1}\sum_{i=1}^{R}c_i(x).
\end{align*}

Applying this decomposition to $\rho_{\mathrm{sc}}$ gives a shortcut contribution map $c^{\mathrm{sc}}(x)$. Applying it to $\rho_{\mathrm{task}}$ gives a task contribution map $c^{\mathrm{task}}(x)$. These maps show which grid cells support the corresponding alignment score for that image.

These contribution maps contain both positive and negative terms. We use the positive terms because our goal is to identify regions that support the shortcut or task alignment score. Negative shortcut contribution is not equivalent to task contribution, and negative task contribution is not equivalent to shortcut contribution; negative terms only indicate regions that reduce the corresponding alignment score. We therefore define
\begin{align}
m^{sc}(x)=\big(c^{\mathrm{sc}}(x)\big)_{+},
\qquad
m^{task}(x)=\big(c^{\mathrm{task}}(x)\big)_{+},
\label{eq:positive-contribution-maps}
\end{align}
where $(z)_+=\max(z,0)$ denotes the positive part. Here, $m^{sc}(x)\in\mathbb{R}_{+}^{R}$ and $m^{task}(x)\in\mathbb{R}_{+}^{R}$ denote the shortcut and task contribution maps, respectively.

A positive product includes two same-sign residual cases where either both regional ranks may be higher than their linear predictions, or both may be lower. Accordingly, the positive contribution maps represent complete positive support for shortcut or task alignment and are not restricted to regions jointly assigned high attribution. We decompose these cases and study the corresponding visual inspection and auditing analysis in Appendix Section~\ref{supp:sec-quadrants}.

\subsection{Spatial summaries of contribution maps}
\label{sec:method-patterns}

To utilise the image-level $m^{sc}$ and $m^{task}$ maps for model auditing, we seek to summarise recurring spatial patterns across the test set. We investigate two methods for estimating these recurring spatial patterns: K-means, which assigns each image to one spatial pattern, and NMF, which allows any given image to express several recurring spatial patterns at once.

For both grouping methods, we first concatenate the shortcut and task contribution maps and normalise the complete joint representation:
\begin{align}
v_i
=
\frac{
[m^{sc}(x_i);m^{task}(x_i)]
}{
\lVert m^{sc}(x_i)\rVert_1+
\lVert m^{task}(x_i)\rVert_1
}
\in\mathbb{R}_{+}^{2R}.
\label{eq:joint-contribution-representation}
\end{align}
Thus, $\lVert v_i\rVert_1=1$.

\paragraph{Hard grouping by contribution map similarity.}
For K-means, we apply $L^2$ normalisation,
\begin{align*}
\hat v_i=\frac{v_i}{\lVert v_i\rVert_2},
\end{align*}

and cluster the representations $\{\hat v_i\}_{i=1}^{N}$. Each image is assigned to one of the $K$ shortcut groups. 
Let $\mathcal{I}_k$ denote the images assigned to group $k$, and define the hard membership weight
\begin{align*}
\kappa_{ik}=\mathbbm{1}\{i\in\mathcal{I}_k\}.
\end{align*}

We obtain the K-means group contribution representation by averaging the jointly normalised representations assigned to the group:
\begin{align}
\bar v_k^{\mathrm{KM}}
=
\frac{\sum_i \kappa_{ik}v_i}{\sum_i\kappa_{ik}}
=
\frac{1}{|\mathcal{I}_k|}
\sum_{i\in\mathcal{I}_k}v_i.
\label{eq:kmeans-group-representation}
\end{align}

\paragraph{Soft mixture summaries with NMF.}
We also use nonnegative matrix factorisation (NMF)~\citep{gillis2011nonnegative,lee2000algorithms,lee1999learning} to obtain $K$ prototype contribution representations and soft membership weights. Let $V\in\mathbb{R}_{+}^{N\times 2R}$ be the matrix whose $i$-th row is $v_i$. We factorise
\begin{align*}
V\approx WH,
\qquad
W\in\mathbb{R}_{+}^{N\times K},
\qquad
H\in\mathbb{R}_{+}^{K\times 2R},
\end{align*}

by minimising the generalised Kullback--Leibler (KL) divergence.

We normalise each row of $H$ to unit $L^1$-norm and rescale the corresponding column of $W$ so that the product $WH$ remains unchanged. Let
\begin{align*}
\bar H_{k:}=\frac{H_{k:}}{\lVert H_{k:}\rVert_1},
\qquad
\bar W_{ik}=W_{ik}\lVert H_{k:}\rVert_1.
\end{align*}

We define the membership weights as
\begin{align*}
\pi_{ik}
=
\frac{\bar W_{ik}}
{\sum_{k'}\bar W_{ik'}}.
\end{align*}

The membership-weighted NMF group contribution representation is
\begin{align}
\bar v_k^{\mathrm{NMF}}
=
\frac{\sum_i\pi_{ik}v_i}{\sum_i\pi_{ik}}.
\label{eq:nmf-group-representation}
\end{align}

Throughout the experiments, the shortcut group-level contribution maps or prototypes are membership-weighted averages $\bar v_k^{\mathrm{NMF}}$.

\subsection{Test-time spatial interventions}
\label{sec:method-interventions}

We use test-time input-space and feature-space spatial interventions to evaluate whether shortcut contribution maps and task contribution maps identify regions that can negatively affect the audited task model.

For each image, intervention maps can be formed using three different sources. The image-level intervention is performed by using the image's own $m^{sc}(x_i)$ and $m^{task}(x_i)$ maps. The shortcut group-level intervention is performed using shortcut group contribution maps associated with $x_i$. The dataset-level intervention is performed using the contribution maps obtained by aggregating evidence over the full dataset, following the dataset-level summary used in \cite{achara2025localising}. For a selected source, let $m^{sc}$ and $m^{task}$ denote the shortcut contribution map and task contribution map used for intervention.

\textbf{Input masking.} For the input space interventions, we mask the shortcut and task related regions and compare the model performance to the original (unmasked) performance.

\textbf{Feature space interventions.} For feature space interventions, we propose shortcut residual suppression, which removes the task contribution component from the shortcut contribution map before suppressing shortcut-dominant regions. We first normalise both maps:
\begin{align*}
\hat{m}^{sc}
=
\frac{m^{sc}}{\lVert m^{sc}\rVert_2},
\qquad
\hat{m}^{task}
=
\frac{m^{task}}{\lVert m^{task}\rVert_2}.
\end{align*}
We then compute the shortcut suppression map after projecting out the task contribution direction:
\begin{align*}
\Gamma_{\mathrm{sc}}
=
\left[
\hat{m}^{sc}
-
\left\langle \hat{m}^{sc}, \hat{m}^{task}\right\rangle
\hat{m}^{task}
\right]_{+},
\end{align*}
where $[z]_{+}=\max(z,0)$. The suppression map is rescaled to have maximum value one by dividing all its values by the largest value across the $R$ spatial regions:
\begin{align*}
\tilde{\Gamma}_{\mathrm{sc}}
=
\frac{
\Gamma_{\mathrm{sc}}
}{
\max_j \Gamma_{\mathrm{sc},j}
}.
\end{align*}

The spatial intervention scaling factor at location $i$ is then
\begin{align}
\gamma_i
=
\max\left(
1 - \tilde{\Gamma}_{\mathrm{sc},i},
0
\right),
\end{align}

When shortcut contribution and task contribution strongly overlap, shortcut suppression can also suppress task-relevant regions. Therefore, we also evaluate task amplification where we remove the shortcut contribution direction from the task contribution map and amplify the task relevant regions.

\begin{align*}
\Gamma_{\mathrm{task}}
=
\left[
\hat{m}^{task}
-
\left\langle
\hat{m}^{task},\hat{m}^{sc}
\right\rangle
\hat{m}^{sc}
\right]_{+}.
\end{align*}

Analogously, we rescale the task amplification map to have maximum value one:
\begin{align*}
\tilde{\Gamma}_{{task}}
=
\frac{
\Gamma_{\mathrm{task}}
}{
\max_j \Gamma_{\mathrm{task},j}
}.
\end{align*}

The spatial intervention scaling factor at location $i$ is then
\begin{align}
\gamma_i = 1+ \tilde\Gamma_{\mathrm{task},i}
\end{align}

We also evaluate a combined intervention that applies shortcut suppression and task amplification simultaneously. The corresponding scaling factor is

\begin{align}
\gamma_i = \max\left(1-\tilde{\Gamma}_{\mathrm{sc},i},0\right) \left( 1+\tilde{\Gamma}_{\mathrm{task},i} \right),
\end{align}

Since $\tilde{\Gamma}_{\mathrm{sc},i},\tilde{\Gamma}_{\mathrm{task},i}\in[0,1]$, the combined scale satisfies $\gamma_i\in[0,2]$. Maximum task amplification occurs when $\tilde{\Gamma}_{\mathrm{sc},i}=0$ and $\tilde{\Gamma}_{\mathrm{task},i}=1$, whereas $\tilde{\Gamma}_{\mathrm{sc},i}=1$ suppresses the feature completely regardless of task amplification. Thus, the multiplicative form moderates task amplification in regions where shortcut and task contributions co-occur.

For all the approaches, the feature at location (e.g. patch for ViT) $i$ is scaled as:
\begin{align*}
h_i'
=
\gamma_i h_i.
\end{align*}

%% file: sections/datasets.tex
\section{Datasets and preprocessing}
\label{sec:datasets}

\input{Tables/datasets.tex}

We evaluate on two natural image datasets (\textsc{CelebA}~\citep{liu2015deep} and \textsc{Waterbirds}~\citep{sagawa2019distributionally}), one chest X-ray dataset \citep{irvin2019chexpert}, one histopathology dataset \citep{litjens20181399} and one skin lesion dataset \citep{codella2018skin,tschandl2018ham10000,combalia2019bcn20000}. In all experiments, the prediction task is binary with label $Y\in\{0,1\}$, and the spurious or sensitive attribute is binary with $A\in\{0,1\}$. The meaning of $Y$ and $A$ is dataset specific and is defined below. To evaluate our method, we deliberately introduce spurious correlations into the training data of each of these datasets and the resulting training split compositions are reported in Table~\ref{tab:train_splits}.

\textbf{\textsc{CelebA}.}
\textsc{CelebA}~\citep{liu2015deep} contains face images with annotated attributes. We predict hair colour, with $Y=0$ denoting non-blond and $Y=1$ denoting blond, and use gender as the sensitive attribute, with $A=0$ denoting female and $A=1$ denoting male.

\textbf{\textsc{CheXpert}.}
\textsc{CheXpert}~\citep{irvin2019chexpert} contains chest radiographs. We predict pleural effusion, with $Y=0$ denoting absence and $Y=1$ denoting presence, and use recorded sex as the sensitive attribute, with $A=0$ denoting female and $A=1$ denoting male. We restrict to frontal anterior--posterior views and stratify by age for split construction, grouping subjects younger than $45$ as \emph{young} and the remainder as \emph{old}. All train, validation, and test splits are subject-wise disjoint.

\textbf{\textsc{Waterbirds}.}
\textsc{Waterbirds}~\citep{sagawa2019distributionally} is a synthetic bird classification dataset derived from CUB~\citep{welinder2010caltech} and Places~\citep{zhou2017places}, where bird images are composited onto land or water backgrounds. We retain the original Waterbirds landbird–waterbird categorisation~\cite{sagawa2019distributionally}, with $Y=0$ denoting landbird and $Y=1$ denoting waterbird, and use background as the spurious attribute, with $A=0$ denoting land background and $A=1$ denoting water background. We create a training set of $10000$, validation set of $500$ and a test set of $1000$ images by overlaying the CUB dataset-based bird foregrounds, extracted using their segmentation masks, onto backgrounds from the Places365 dataset. Because the number of distinct waterbird foregrounds is limited, some foregrounds recur across images, including across dataset splits. Similar to~\cite{sagawa2019distributionally}, we use forests for land backgrounds and oceans/lakes for water backgrounds, then stratify by bird type and background.

\textbf{\textsc{Camelyon17}.}
\textsc{Camelyon17}~\citep{litjens20181399} is a histopathology dataset of breast cancer lymph-node tissue patches collected across multiple medical centres. We use the patch-based version, where each original input is a $96\times96$ image patch and the task is binary tumour detection. We predict whether the central region of a patch contains tumour tissue, with $Y=0$ denoting normal tissue and $Y=1$ denoting tumour tissue. We use medical centre as the spurious attribute, binarising the centre metadata into two groups: $A=0$ denotes centres $0$ and $1$, while $A=1$ denotes centres $2$, $3$, and $4$.

\textbf{\textsc{ISIC2019}.}
\textsc{ISIC2019} is a dermoscopic skin lesion dataset aggregated from multiple sources, including \textsc{HAM10000}~\citep{tschandl2018ham10000} and \textsc{BCN\_20000}~\citep{combalia2019bcn20000}. We construct a binary lesion classification task by retaining melanoma (MEL) and Melanocytic Nevus (NV) images, with $Y=0$ denoting benign nevus and $Y=1$ denoting malignant melanoma. We use dataset source as the spurious attribute, with $A=0$ denoting \textsc{HAM10000} ($600\times450$ images) and $A=1$ denoting \textsc{BCN\_20000} ($1024\times1024$ images). 

All images are resized to $224\times224$ and standardised using ImageNet statistics (except for CheXpert where images are z-scored).

%% file: Tables/datasets.tex
\begin{wraptable}{r}{0.5\textwidth}
\centering
\caption{\textbf{Biased training splits with strong label–attribute correlations.}
Counts are reported for the four $(Y,A)$ (label, attribute) groups.}
\label{tab:train_splits}

\resizebox{0.5\textwidth}{!}{%
\begin{tabular}{lcccc}
\toprule
Dataset & $(Y{=}0,A{=}0)$ & $(Y{=}0,A{=}1)$ & $(Y{=}1,A{=}0)$ & $(Y{=}1,A{=}1)$ \\
\midrule
\textsc{CelebA}   & 500 & 4500 & 4500 & 500 \\
\textsc{CheXpert} & 500 & 4500 & 4500 & 500 \\
\textsc{Waterbirds}\footnotemark[2] & 4900 & 100 & 100 & 4900 \\
\textsc{Camelyon17} & 500 & 14500 & 14500 & 500 \\
\textsc{ISIC2019} & 2000 & 500 & 500 & 2000 \\
\bottomrule
\end{tabular}%
}
\end{wraptable}
\footnotetext[2]{We deliberately introduce a higher dataset bias for Waterbirds as the group-level performance disparities were minimal for the model trained using 500 discordant samples.}

%% file: sections/experiments.tex
\section{Experiments}
\label{sec:experiments}

The experiments are organised around three questions:
\begin{itemize}
    \itemsep0em 
    \item Do the learned spatial patterns produce inspectable contribution maps and representative images?
    \item Do these patterns identify subsets with different subgroup composition and error rates which can help auditors efficiently assess model robustness?
    \item Can dataset-level, shortcut group-level, or image-level contribution maps guide spatial interventions in fixed trained models?
\end{itemize}

For evaluation, we use balanced test sets with $N=1000$ images for \textsc{CelebA}, \textsc{CheXpert}, \textsc{Waterbirds}, and \textsc{ISIC2019}, and $N=2000$ images for \textsc{Camelyon17}. This gives $250$ images per $(Y,A)$ subgroup for the first four datasets and $500$ images per subgroup for \textsc{Camelyon17}. For each image, we compute the shortcut contribution map $m^{sc}(x)$, and the task contribution map $m^{task}(x)$ as described in Sections~\ref{sec:method-alignment} and~\ref{sec:method-contribution-maps}.

The audited task model $f_{\mathrm{TS}}$ is trained using the datasets summarised in Table~\ref{tab:train_splits}. The baseline task model $f_{\mathrm{BA}}$ is trained to predict $Y$ on a $(Y,A)$-balanced training set and the sensitive attribute model $f_{\mathrm{SA}}$ is trained to predict the sensitive attribute $A$ on the same balanced training set. For \textsc{CheXpert} and \textsc{Waterbirds}, the balanced training set contains $10000$ images, with $2500$ from each subgroup. The approximately balanced \textsc{CelebA} training set contains $8874$ images, with $2500$ from three subgroups and $1374$ from the blond-Hair/male subgroup. The approximately balanced \textsc{Camelyon17} training set contains $27814$ images, with $7500$ from three subgroups and $5314$ from the tumour/centres 2--4 subgroup. Finally, the approximately balanced \textsc{ISIC2019} training set contains $4742$ images, with $1250$ from three subgroups and $992$ from the malignant/non-BCN subgroup. $f_{\mathrm{TS}}, f_{\mathrm{BA}}$ and $f_{\mathrm{SA}}$ use the same model architectures. Table~\ref{tab:aux_model_perf} shows their performance on the balanced test sets. The baseline task models have a lower largest performance gap (LPG) i.e. the accuracy gap between the best and worst performing subgroups than the audited task models, indicating the presence of shortcut learning in the audited models, and the sensitive attribute models achieve classification accuracies above $0.90$. This preliminary balanced test set performance assessment sets up the further analysis on the attribution maps.

\input{Tables/acc}

In our experiments, we use ResNet50~\citep{he2016deep} and ViT-B/16~\citep{dosovitskiy2020image} models. 
For the main attribution analyses, we use LRP~\citep{binder2016layer} for ResNet50 and AttnLRP~\citep{pmlr-v235-achtibat24a,arras2025close} for ViT-B/16.  To assess whether our findings generalise for other attribution methods, we also report additional results with Grad-CAM~\citep{selvaraju2017grad} for ResNet50, and with ViT-CX~\citep{xie2022vit} and transformer input sampling (TiS)~\citep{englebert2023explaining} for ViT-B/16 (See Figure~\ref{fig:supp_error_strat_heldout} and Appendix Sections~\ref{sec:supp-grid-var},~\ref{sec:supp-contribution-map-figures}).

In the following experiments, unless stated otherwise, shortcut groups are formed from $56\times56$ contribution maps using NMF with $K=8$; LRP is used for ResNet50 and AttnLRP for ViT-B/16. For the quantitative results, we report mean $\pm$ standard deviation over four independently trained models.

\subsection{Spatial contribution patterns are inspectable and stratify failures}
\label{sec:exp-patterns}

We first examine whether the discovered shortcut groups produce inspectable spatial patterns. For each dataset and model, we summarise the identified groups using shortcut contribution maps, task contribution maps, representative examples, subgroup composition, and error rates. The goal is to check whether the shortcut group-level contribution maps show recognisable spatial structure, and whether the subsets of images that belong strongly to specific shortcut groups differ in label, attribute, and/or error rates.

\begin{figure}[t]
\centering
\resizebox{\textwidth}{!}{
\includegraphics[width=\linewidth]{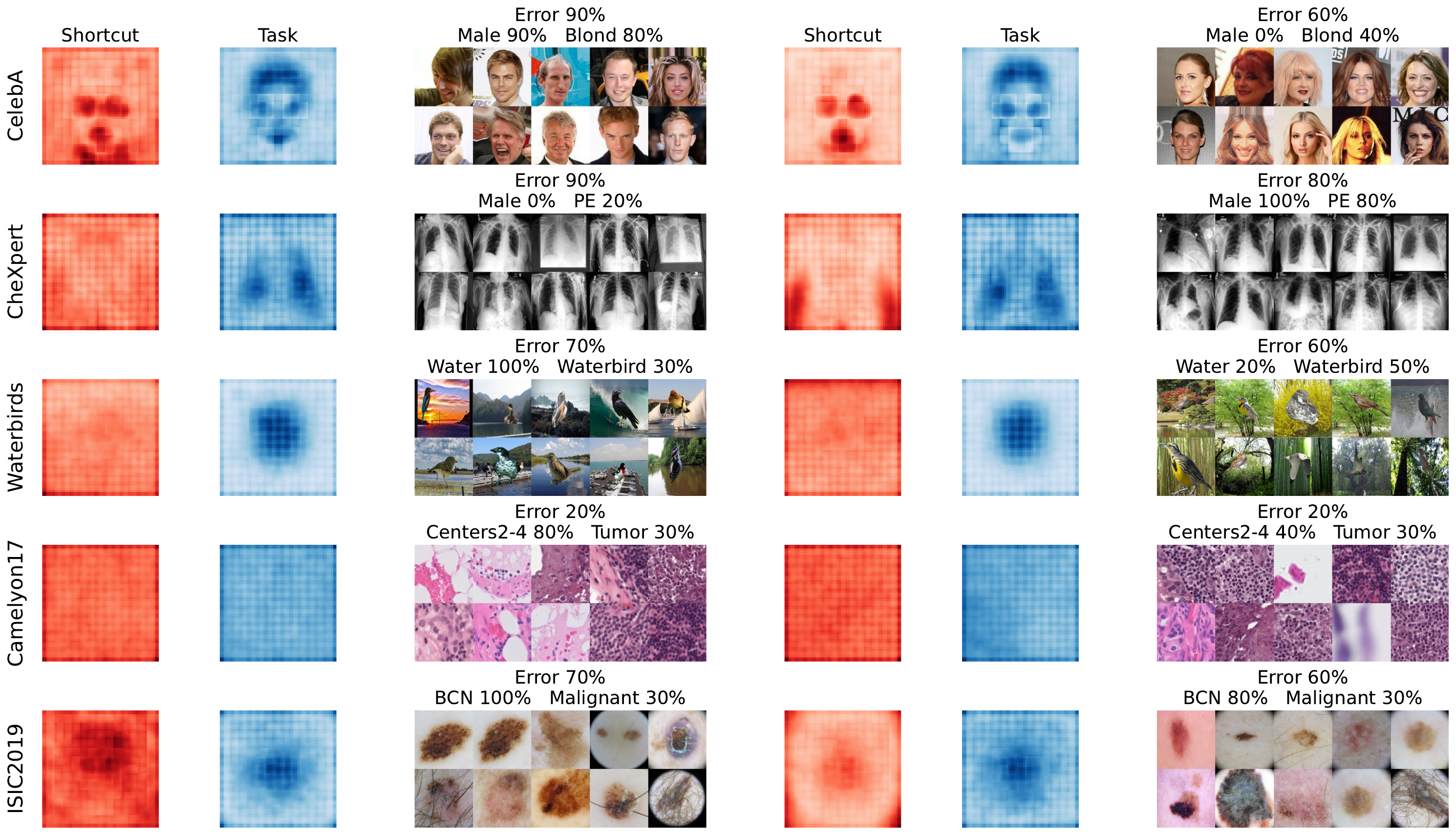}}
\caption{\textbf{ViT shortcut groups separate recurring spatial patterns with distinct subgroup composition and error rates.} Each panel shows a shortcut group prototype and corresponding top $10$ representative images that most strongly express the pattern along with their error rates and subgroup composition.}
\label{fig:regime_exemplars_vit_celeba}
\vspace{-10pt}
\end{figure}

\begin{figure}[t]
\centering
\resizebox{\textwidth}{!}{
\includegraphics[width=\linewidth]{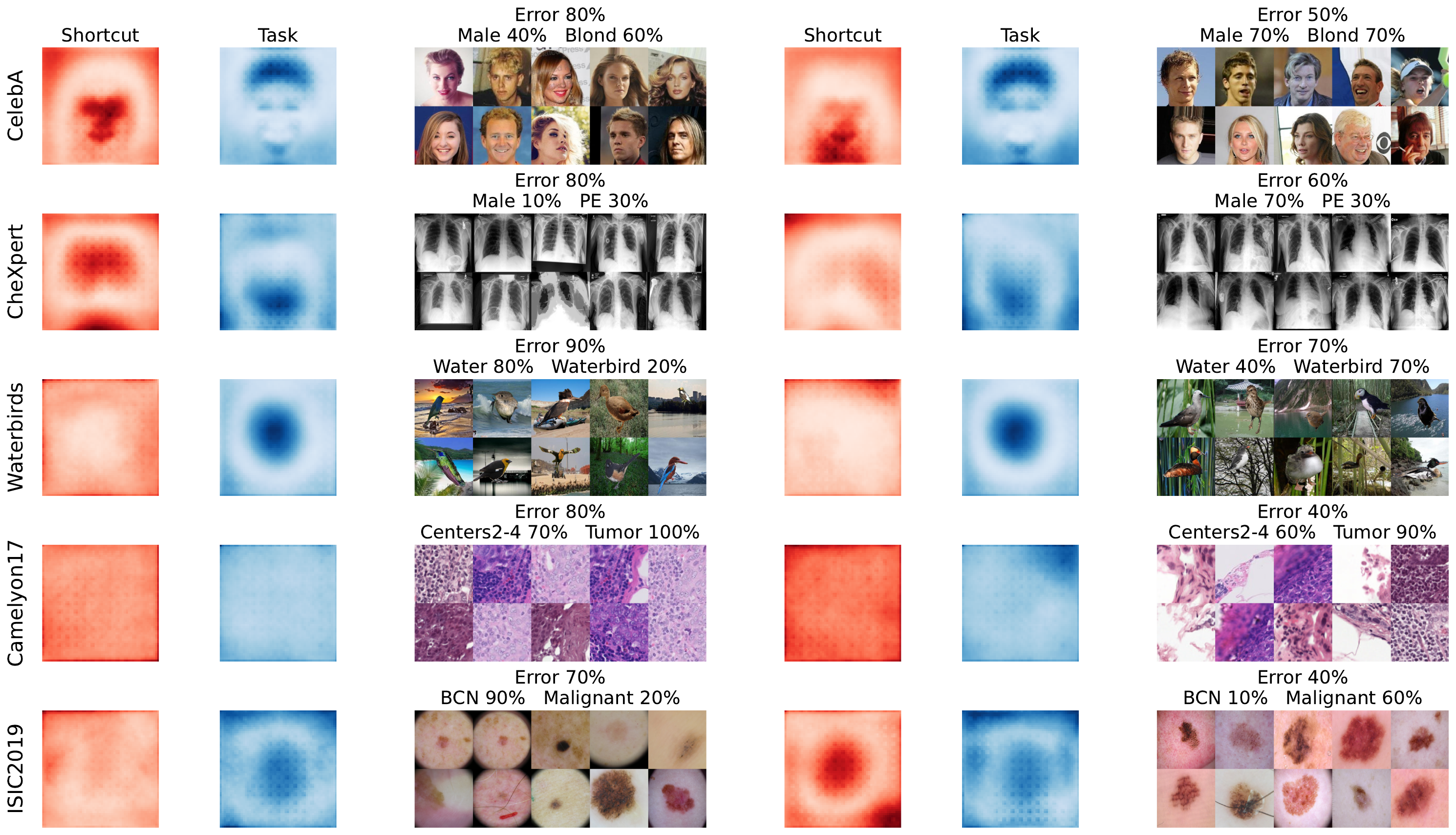}}
\caption{\textbf{ResNet shortcut groups also reveal recurring spatial patterns with distinct subgroup composition and error rates.} Each panel shows a shortcut group prototype and corresponding top $10$ representative images that most strongly express the pattern along with their error rates and subgroup composition.}
\label{fig:chexpert_artifact_exemplars}
\end{figure}

Figures~\ref{fig:regime_exemplars_vit_celeba} and~\ref{fig:chexpert_artifact_exemplars} show the shortcut and task contribution patterns for ViT and ResNet models on \textsc{CelebA}, \textsc{CheXpert}, \textsc{Waterbirds}, \textsc{Camelyon17} and \textsc{ISIC2019}. Each panel contains membership weighted shortcut ($m^{sc}$) and task ($m^{task}$) contribution maps and the corresponding top $10$ images based on the images' NMF membership scores. While we obtain $K=8$ different shortcut groups, we select for visualisation the $2$ groups with the highest top-$10$ error rates. The annotations show the corresponding top $10$ label, sensitive attribute and error percentages.

Below we summarise the key observations from all datasets: 
\begin{itemize}
    \itemsep0em 
    \item Since \textsc{CelebA} faces are approximately aligned, upper cells correspond mostly to hair and forehead regions, central cells to facial features, and lower cells to clothing or background. The primary difference between the task and shortcut contributions is the hair region where the task contribution is concentrated (recall that the task is hair colour classification), and the lower regions where the shortcut contribution is concentrated.
    \item For \textsc{CheXpert}, shortcut prototypes concentrate contribution around the breast shadow or lower non-lung regions, while task prototypes highlight the lung regions (further analysis in Section~\ref{sec:localisation_checks}; Figure~\ref{fig:chexmask}, Table~\ref{tab:chexpert_localisation}).
    \item For \textsc{Waterbirds}, the task prototypes concentrate in the central regions of the image which is where the birds are generally located. The shortcuts are more diffuse as expected for the background shortcut (further analysis in Section~\ref{sec:localisation_checks}; Table~\ref{tab:waterbirds_localisation}).
    \item For \textsc{Camelyon17}, the shortcuts and task contributions are both generally diffuse, although some shortcut contributions may be more localised.
    \item For \textsc{ISIC}, the shortcut contributions are heterogeneous with higher concentrations in some regions closer or overlapping with the task regions. The task contributions however, are generally in the central regions for both ResNet and ViT.
\end{itemize}

Qualitatively, the ViT contribution patterns are often easier to interpret visually than the ResNet patterns, especially on \textsc{CheXpert}. The exemplar images associated with the shortcut groups also highlight subgroup compositions that have higher errors in the top $10$. For example, the first \textsc{CelebA} group has a top-$10$ error rate of 90\%, with subgroup compositions of 90\% male and 80\% blond. Furthermore, the \textsc{CelebA} groups range from 90\% male to 100\% female, while the \textsc{CheXpert} groups range from 100\% female to 100\% male, indicating substantial sensitive attribute separation. Finally, the top-$10$ error rates for the different datasets are substantially higher than the dataset average except for the \textsc{Camelyon17} dataset for the ViT model. Additional shortcut group contribution maps across attribution methods, grouping approaches, and values of $K$ are provided in Appendix Section~\ref{sec:supp-contribution-map-figures}.

We next test whether the shortcut groups can be used for selecting a subset of images that represent a substantial proportion of errors from the overall test set. For each dataset and model, we randomly divide the balanced test set into two equally sized, subgroup-balanced splits (five times). We fit K-means and NMF on one split, estimate a shortcut group risk score from errors in that split, and use the learned patterns to rank examples in the held-out split. Held-out examples are assigned to the nearest K-means centroid or transformed using the fitted NMF components to obtain their membership weights. We then inspect the held-out images with the top $20\%$ risk scores and compare the percentage of total held-out errors in this subset with the random expectation baseline of 20\%.

Let $\mathcal{I}^{\mathrm{fit}}$ denote the indices of images in
$\mathcal{D}_{\mathrm{fit}}$, and let $e_i=\mathbbm{1}\{\hat y_i\neq y_i\}$ indicate whether image $x_i$ is misclassified.

For K-means, let $\mathcal{I}_k^{\mathrm{fit}}$ denote the fitting images assigned to group $k$. Its empirical error rate is
\begin{align*}
\epsilon_k = \frac{1}{|\mathcal{I}_k^{\mathrm{fit}}|}
\sum_{i\in\mathcal{I}_k^{\mathrm{fit}}}e_i.
\end{align*}

For each held-out image $x_j$, we identify the most similar K-means shortcut group based on its joint contribution-map representation $\hat v_j$. Let $k_j^{\mathrm{held}}$ denote the resulting group. Its risk score is
\begin{align*}
s_{\mathrm{risk}}(x_j) = \epsilon_{k_j^{\mathrm{held}}}.
\end{align*}

For NMF, we estimate the error rate of group $k$ as the membership-weighted mean error over the fitting split:
\begin{align*}
\epsilon_k
=
\frac{
\sum_{i\in\mathcal{I}^{\mathrm{fit}}}\pi_{ik}e_i
}{
\sum_{i\in\mathcal{I}^{\mathrm{fit}}}\pi_{ik}
}.
\end{align*}

The held-out NMF risk score should then be:

\begin{align*}
s_{\mathrm{risk}}(x_j)= \sum_{k=1}^{K}\pi_{jk}^{\mathrm{held}}\epsilon_k.
\end{align*}

We rank the held-out images by $s_{\mathrm{risk}}(x_j)$ and evaluate the percentage of all held-out errors in the top 20\%.

\begin{figure}[t!]
\centering
\resizebox{\textwidth}{!}{
\includegraphics[width=\linewidth]{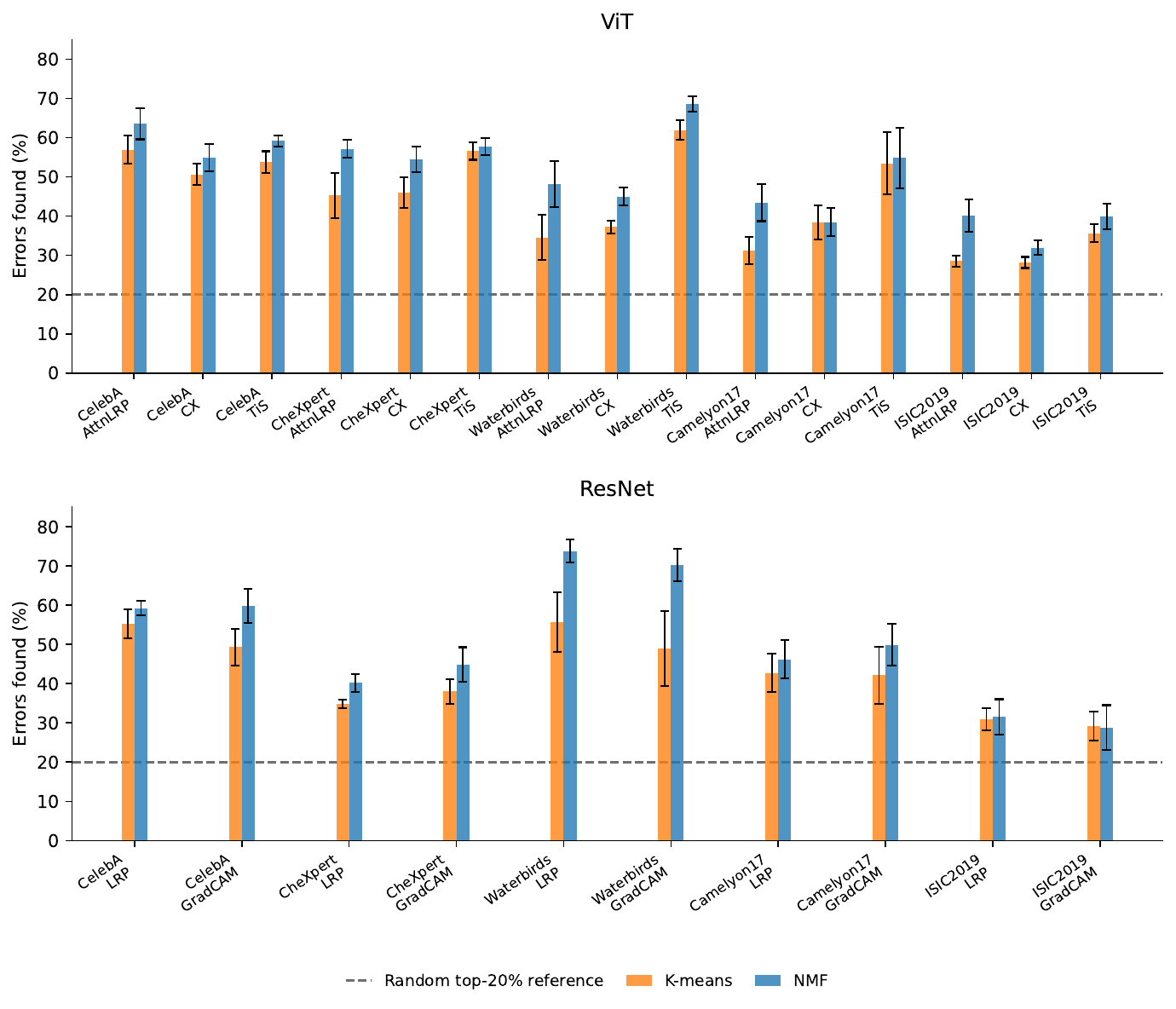}}
\caption{\textbf{Shortcut group risk score derived subsets contain a large fraction of model errors.} The dashed line shows the random baseline, under which $20\%$ of the total errors are expected to be found, while the orange and blue bars show the percentage of total errors found in the selected top $20\%$ of (held-out) images ranked by the shortcut group derived risk score for K-means and NMF, respectively. The error bars indicate the standard deviation across independently trained models (after averaging over 5 random held-out sets for each model).}
\label{fig:supp_error_strat_heldout}
\vspace{-5pt}
\end{figure}

Figure~\ref{fig:supp_error_strat_heldout} shows that the selected subsets contain a substantially larger percentage of errors than the random baseline across datasets, architectures, and attribution methods using a $56\times56$ grid based partition. Additionally, the NMF-based grouping approach is better than K-means in most cases. Therefore, we use NMF for our visual inspection and intervention analyses. We provide further analyses on the number of shortcut groups $K$, the percentage of images used for selection and comparisons against other baselines for computing risk scores in Appendix Sections~\ref{sec:supp-k},~\ref{sec:supp-inspection-budget} and~\ref{sec:supp-joint-representation-baselines}.

\subsection{Intervention tests}
\label{sec:interventions}

We next ask whether the spatial evidence identified by the contribution analysis is relevant to model predictions. If regions with high shortcut contribution are used by the model, then removing or suppressing those regions should change model performance. We therefore use intervention tests to connect the discovered shortcut and task contribution maps to model performance.

We first use input masking which tests whether regions with high shortcut or task contribution contain visual signal that affects the prediction. We then intervene inside the model, where the same spatial evidence is used to modulate internal representations. Finally, we test an iterative intervention approach to understand if shortcut and task contribution maps can be further actionable. In terms of metrics, we report the $\Delta\mathrm{LPG}_{\mathrm{red}}$ = $\mathrm{LPG}_{no\ intervention}$ - $\mathrm{LPG}_{intervention}$, so positive values denote reduced subgroup disparity and $\Delta\mathrm{Acc.} = \mathrm{Acc.}_{intervention} - \mathrm{Acc.}_{no\ intervention}$ which helps us understand if positive $\Delta$ LPG$_{red}$ comes at the cost of overall model performance.

\subsubsection{Input image interventions}

We first perform input interventions by masking the selected image regions obtained using the contribution maps and evaluating the resulting performance. We compare three occlusion types: 
\begin{itemize}
    \itemsep0em 
    \item masking the top 12.5\% ($32$) regions with the highest shortcut contributions ($m^{sc}$), under the constraint that the mask does not coincide with the top 12.5\% regions from $m^{task}$.
    \item masking the top 12.5\% ($32$) regions with the highest task contributions $m^{task}$, under the constraint that the mask does not coincide with the top 12.5\% regions from $m^{sc}$.
    \item masking 12.5\% ($32$) randomly selected regions outside the high $m^{sc}$ cells.
\end{itemize}
If fewer than $32$ non-overlapping cells are available, we fill the remaining mask positions using cells with the lowest value in the opposite evidence type i.e. lowest $m^{task}$ for shortcut masks and lowest $m^{sc}$ for task masks. In this analysis we use a $16\times16$ grid.

\input{Tables/occlusion}

Table~\ref{tab:input_occlusion_summary} shows that the largest degradation in model performance occurs when regions with high task contribution are masked, especially based on LPG, which increases substantially. For both shortcut group-level and image-level contribution maps, task-region masking reduces accuracy more than random masking for every dataset and model except ResNet on \textsc{ISIC2019}, indicating that the task-contribution maps identify regions relevant to the model’s predictions. When regions with high shortcut contribution are masked, the effect is weaker and accuracy is often preserved or improved, although changes in LPG are inconsistent across datasets and models. For \textsc{CelebA}, shortcut masking results in minimal changes in LPG and accuracy for ResNet and slightly larger increases in LPG for ViT, whereas task masking substantially increases LPG for both models. For \textsc{Waterbirds}, shortcut masking largely preserves accuracy but increases LPG. For \textsc{CheXpert}, the largest LPG reduction for ResNet is obtained with image-level masking and with shortcut group-level masking for ViT. In contrast, for \textsc{Camelyon17} and ResNet on \textsc{ISIC2019}, LPG increases following shortcut masking (modest reductions are observed for ViT on \textsc{ISIC2019}), suggesting weaker spatial separability between shortcut and task evidence in input space.

These occlusion tests indicate whether masking using the contribution maps is useful or harmful for the model, but they are not conclusive. Masking image regions can introduce distribution shift, and input occlusion does not reveal where shortcut information is propagated inside the models. We also discuss the results based on a $56 \times 56$ grid where random masking sometimes results in higher performance drops indicating potential distribution shifts (Appendix Section~\ref{sec:supp-occlusion-grid}).

\subsubsection{Internal representation interventions}
\label{sec:experiments-internal-representations}

In this section, we intervene on internal representations and the goal is to test whether the contribution maps can be used to update features inside the trained model to suppress or amplify the shortcut or task contribution regions and understand the corresponding changes in the model predictions.

For the experiments, we evaluate the contribution maps using the feature-space interventions defined in Section~\ref{sec:method-interventions}. We apply shortcut and task suppression to down-weight features identified by the corresponding contribution maps, and task amplification to up-weight task-related features. We also evaluate combined shortcut suppression and task amplification. 

For ViT, we interpolate the $G\times G$ spatial contribution maps to the patch-token grid while keeping the CLS token fixed, and apply the spatial scale ($\gamma$) only to patch tokens. We apply this intervention to the value vectors in all twelve transformer layers. For ResNet, we interpolate the spatial contribution map to the convolutional feature map resolution and apply the spatial scale across channels at the last two residual stages. Additional intervention analyses on target layer selection and transformer block selection are discussed in Appendix Sections~\ref{sec:supp-intervention-sites} and \ref{sec:supp-vit-pathways}, respectively.

In Table~\ref{tab:intervention_source}, we demonstrate the effectiveness of combined shortcut suppression and task amplification across datasets and models for both shortcut group and image-level contribution maps. Additionally, image-level task suppression consistently reduces performance across datasets and models, indicating that the task contribution maps identify regions important for the intended prediction. Image-level shuffled shortcut suppression (random shuffling of shortcut contribution values/regions) is generally neutral or harmful, with only small improvements in some cases, suggesting that the spatial location for suppression or amplification matters. Additionally, we show that shortcut group-level combined interventions are more effective in reducing disparities than dataset and image-level interventions (Appendix Section~\ref{sec:supp-dataset-vs-other-level}).

\input{Tables/intervention_source}

In \textsc{Camelyon17} and \textsc{ISIC2019}, the source attribute can be expressed through distributed acquisition characteristics such as staining, scanner properties, colour calibration, resolution, and texture. Such shortcuts may be difficult to isolate through spatial suppression. Consistent with this premise, shortcut suppression does not reliably reduce LPG on these datasets, showing that suppression alone is less effective in these cases (Section~\ref{sec:supp-alternate-intervention-methods}).

\paragraph{Iterative interventions.}
The preceding analysis considers a single intervention derived from the original contribution maps. To understand whether there is any remaining shortcut alignment after the single interventions, we recompute both shortcut and task contribution maps and use these maps to guide the next intervention. We repeat this process for five iterations and measure the shortcut alignment score $\bar{\rho}_{\mathrm{sc}}$. For this experiment, at iteration~$1$, we apply combined shortcut suppression and task amplification. From iteration~$2$ onward, we apply shortcut suppression using the recomputed maps (alternate intervention strategies are discussed in Appendix Section~\ref{sec:supp-itervariations}). At each step, the newly derived scale is multiplied with the cumulative scale ($\gamma$) from the previous step. So, the scale applied at iteration $5$ is ($\gamma^{(1)}_{combined}\odot\gamma^{(2)}_{sc}\odot\gamma^{(3)}_{sc}\odot\gamma^{(4)}_{sc}\odot\gamma^{(5)}_{sc}$), where the $^(t)$ indicates the iteration at which the scale is applied. The model evaluated at iteration~$t$ therefore reflects all interventions applied from iterations~$1$ through~$t$.

Mean shortcut alignment over the test images is positive before intervention for every dataset and model, and generally decreases over the iterations which is in line with the idea of shortcut suppression. Alongside the first and final iterations, we report the last iteration for which $\bar{\rho}_{\mathrm{sc}}^{(t)}>0$, together with model performance at that point. This shows how performance changes before mean shortcut alignment crosses zero. We also report the final mean shortcut alignment to show whether further intervention moves it below zero. Note that a negative shortcut alignment might still be effective in some cases by suppressing regions positively contributing to the alignment score. However, the effects might be unstable and therefore, we use $\bar{\rho}_{\mathrm{sc}}^{(t)}=0$ as a reference point for assessing the iterative interventions.

\input{Tables/iterative_interventions}

Table~\ref{tab:iterative_value_interventions} shows the first intervention, the last iteration with positive mean shortcut alignment score or closest to zero if mean shortcut alignment is already negative, and the fifth and final iteration. Mean shortcut alignment generally decreases across the interventions. With a combined first step followed by shortcut suppression, these interventions produce positive LPG reductions across all five datasets, with positive accuracy (except for \textsc{CheXpert} where accuracy is near zero). At the final iteration, shortcut group-level maps give LPG reductions of $+9.8$, $+20.3$, $+17.7$, $+6.6$, and $+6.5$ percentage points on \textsc{CelebA}, \textsc{CheXpert}, \textsc{Waterbirds}, \textsc{Camelyon17}, and \textsc{ISIC2019}, respectively.

Results corresponding to ResNet show similar trends and are discussed in Appendix Section~\ref{sec:supp-iterativeresnet}.

%% file: Tables/acc.tex
\begin{wraptable}{r}{0.45\textwidth}
\centering
\caption{\textbf{Balanced training substantially reduces the largest performance gap for every dataset and backbone, while sensitive attributes are highly predictable.} Values are averaged across independently trained models.}
\label{tab:aux_model_perf}
\resizebox{0.45\textwidth}{!}{
\begin{tabular}{llccc}
\toprule
Dataset & Backbone & LPG ($f_{TS}$) $\downarrow$ & LPG ($f_{BA}$) $\downarrow$ & $f_{SA}$ Acc. $\uparrow$ \\
\midrule
\multirow{2}{*}{\textsc{CelebA}}      & ViT    & 0.24 & 0.07 & 0.96 \\
                                      & ResNet & 0.28 & 0.05 & 0.98 \\
\multirow{2}{*}{\textsc{CheXpert}}    & ViT    & 0.37 & 0.10 & 0.94 \\
                                      & ResNet & 0.41 & 0.12 & 0.94 \\
\multirow{2}{*}{\textsc{Waterbirds}}  & ViT    & 0.42 & 0.06 & 0.97 \\
                                      & ResNet & 0.24 & 0.03 & 0.97 \\
\multirow{2}{*}{\textsc{Camelyon17}}  & ViT    & 0.32 & 0.11 & 0.91 \\
                                      & ResNet & 0.40 & 0.28 & 0.91 \\
\multirow{2}{*}{\textsc{ISIC2019}}    & ViT    & 0.38 & 0.14 & 0.99 \\
                                      & ResNet & 0.25 & 0.11 & 0.99 \\
\bottomrule
\end{tabular}}
\end{wraptable}

%% file: Tables/occlusion.tex
\begin{table*}[t]
\centering
\caption{
\textbf{Masking task contribution regions results in severe performance disparities.} Across most datasets and models, performance degradation is highest under task-region masking, intermediate under random masking, and lowest under shortcut-region masking. Random masking values are averaged over 5 random masks for each independently trained model and reported with standard deviation across models. Blue denotes the largest improvement and red the highest performance degradation for each dataset--model pair.
}
\label{tab:input_occlusion_summary}
\resizebox{\linewidth}{!}{%
\begin{tabular}{lllrrrrrrrr}
\toprule
Dataset & Model & Source
& \multicolumn{2}{c}{No intervention}
& \multicolumn{2}{c}{Shortcut}
& \multicolumn{2}{c}{Task}
& \multicolumn{2}{c}{Random} \\
\cmidrule(lr){4-5}
\cmidrule(lr){6-7}
\cmidrule(lr){8-9}
\cmidrule(lr){10-11}
& &
& LPG & Acc.
& $\Delta\mathrm{LPG}_{\mathrm{red}}$ & $\Delta$Acc.
& $\Delta\mathrm{LPG}_{\mathrm{red}}$ & $\Delta$Acc.
& $\Delta\mathrm{LPG}_{\mathrm{red}}$ & $\Delta$Acc. \\
\midrule

\multirow{4}{*}{CelebA}
& \multirow{2}{*}{ResNet} & Shortcut Group
& \multirow{2}{*}{27.6} & \multirow{2}{*}{86.9}
& -1.2 & -.7
& \textcolor{red}{-42.2} & \textcolor{red}{-10.5}
& $-10.9{\pm}5.0$ & $-2.4{\pm}1.5$ \\
& & Image
& &
& \textcolor{blue}{-.3} & \textcolor{blue}{-.2}
& -34.6 & -9.2
& $-10.8{\pm}4.5$ & $-2.2{\pm}1.5$ \\

\cmidrule(lr){2-11}

& \multirow{2}{*}{ViT} & Shortcut Group
& \multirow{2}{*}{23.7} & \multirow{2}{*}{87.2}
& -4.4 & -.8
& \textcolor{red}{-50.7} & \textcolor{red}{-12.5}
& $-8.8{\pm}1.1$ & $-1.8{\pm}.2$ \\
& & Image
& &
& \textcolor{blue}{-3.1} & \textcolor{blue}{-.1}
& -24.7 & -6.2
& $-7.1{\pm}2.4$ & $-1.5{\pm}.4$ \\

\midrule

\multirow{4}{*}{CheXpert}
& \multirow{2}{*}{ResNet} & Shortcut Group
& \multirow{2}{*}{41.4} & \multirow{2}{*}{76.5}
& +5.9 & +.5
& -5.4 & -4.0
& $+5.7{\pm}8.6$ & $-1.5{\pm}1.2$ \\
& & Image
& &
& \textcolor{blue}{+11.8} & \textcolor{blue}{+1.8}
& \textcolor{red}{-6.2} & \textcolor{red}{-4.6}
& $+2.1{\pm}7.9$ & $-1.9{\pm}1.0$ \\

\cmidrule(lr){2-11}

& \multirow{2}{*}{ViT} & Shortcut Group
& \multirow{2}{*}{36.9} & \multirow{2}{*}{79.8}
& \textcolor{blue}{+7.3} & \textcolor{blue}{+1.1}
& -21.4 & -11.3
& $\textcolor{red}{-23.2{\pm}6.0}$ & $-6.5{\pm}1.6$ \\
& & Image
& &
& -3.2 & -2.9
& -18.3 & \textcolor{red}{-12.4}
& $-20.1{\pm}6.0$ & $-5.6{\pm}1.3$ \\

\midrule

\multirow{4}{*}{Waterbirds}
& \multirow{2}{*}{ResNet} & Shortcut Group
& \multirow{2}{*}{24.2} & \multirow{2}{*}{89.1}
& \textcolor{blue}{-1.7} & \textcolor{blue}{+1.3}
& \textcolor{red}{-58.8} & \textcolor{red}{-26.5}
& $-35.1{\pm}5.7$ & $-8.7{\pm}1.8$ \\
& & Image
& &
& -7.4 & -.3
& -55.3 & -20.1
& $-33.6{\pm}5.3$ & $-8.2{\pm}1.6$ \\

\cmidrule(lr){2-11}

& \multirow{2}{*}{ViT} & Shortcut Group
& \multirow{2}{*}{41.6} & \multirow{2}{*}{81.4}
& \textcolor{blue}{-.9} & \textcolor{blue}{+1.1}
& \textcolor{red}{-40.3} & \textcolor{red}{-18.5}
& $-12.3{\pm}2.6$ & $-2.6{\pm}1.0$ \\
& & Image
& &
& -4.1 & 0.0
& -37.5 & -13.4
& $-11.6{\pm}2.6$ & $-2.3{\pm}.9$ \\

\midrule

\multirow{4}{*}{Camelyon17}
& \multirow{2}{*}{ResNet} & Shortcut Group
& \multirow{2}{*}{39.8} & \multirow{2}{*}{84.7}
& -9.7 & -4.1
& -14.6 & -6.2
& $-1.9{\pm}7.8$ & $-5.9{\pm}3.8$ \\
& & Image
& &
& -4.1 & \textcolor{blue}{-3.3}
& \textcolor{red}{-16.3} & \textcolor{red}{-9.8}
& $\textcolor{blue}{-.2{\pm}7.6}$ & $-5.6{\pm}3.7$ \\

\cmidrule(lr){2-11}

& \multirow{2}{*}{ViT} & Shortcut Group
& \multirow{2}{*}{31.6} & \multirow{2}{*}{87.0}
& -7.9 & -3.5
& -11.8 & -5.1
& $-4.3{\pm}1.6$ & $-3.6{\pm}.4$ \\
& & Image
& &
& -8.3 & \textcolor{blue}{-3.3}
& \textcolor{red}{-18.5} & \textcolor{red}{-8.0}
& $\textcolor{blue}{-3.7{\pm}1.0}$ & $-3.6{\pm}.6$ \\

\midrule

\multirow{4}{*}{ISIC2019}
& \multirow{2}{*}{ResNet} & Shortcut Group
& \multirow{2}{*}{25.0} & \multirow{2}{*}{79.7}
& \textcolor{blue}{-2.2} & \textcolor{blue}{-.3}
& -6.7 & -2.5
& $-15.2{\pm}8.9$ & $\textcolor{red}{-3.6{\pm}.8}$ \\
& & Image
& &
& -6.4 & -1.1
& -9.9 & -2.7
& $\textcolor{red}{-15.9{\pm}8.9}$ & $-3.5{\pm}.9$ \\

\cmidrule(lr){2-11}

& \multirow{2}{*}{ViT} & Shortcut Group
& \multirow{2}{*}{38.2} & \multirow{2}{*}{76.3}
& +3.7 & \textcolor{blue}{+.6}
& \textcolor{blue}{+4.0} & \textcolor{red}{-1.3}
& $+1.5{\pm}3.1$ & $-.5{\pm}.7$ \\
& & Image
& &
& +2.6 & -.2
& +2.2 & -.8
& $\textcolor{red}{+1.1{\pm}2.4}$ & $-.5{\pm}.9$ \\
\bottomrule
\end{tabular}%
}
\vspace{-5pt}
\end{table*}

%% file: Tables/intervention_source.tex
\begin{table*}[t]
\centering
\caption{
\textbf{Combined shortcut suppression and task amplification generally reduces subgroup disparities, whereas task suppression degrades performance.} Values show changes post-intervention, reported as mean and standard deviation across independently trained models. Shuffled values are averaged over five random shuffles per model. Blue denotes the largest improvement and red the highest performance degradation for each dataset--model pair.}
\label{tab:intervention_source}
\resizebox{\linewidth}{!}{%
\begin{tabular}{llrrrrrrrr}
\toprule
& & \multicolumn{2}{c}{Shuffled}
& \multicolumn{2}{c}{Task}
& \multicolumn{2}{c}{Shortcut Group}
& \multicolumn{2}{c}{Image} \\
\cmidrule(lr){3-4}
\cmidrule(lr){5-6}
\cmidrule(lr){7-8}
\cmidrule(lr){9-10}
Dataset & Backbone
& $\Delta\mathrm{LPG}_{\mathrm{red}}$ & $\Delta$Acc.
& $\Delta\mathrm{LPG}_{\mathrm{red}}$ & $\Delta$Acc.
& $\Delta\mathrm{LPG}_{\mathrm{red}}$ & $\Delta$Acc.
& $\Delta\mathrm{LPG}_{\mathrm{red}}$ & $\Delta$Acc. \\
\midrule

\multirow{2}{*}{CelebA}
& ResNet
& $+1.4{\pm}2.3$ & $+.1{\pm}.2$
& $\textcolor{red}{-8.1{\pm}5.0}$ & $\textcolor{red}{-3.7{\pm}.6}$
& $\textcolor{blue}{+7.7{\pm}5.0}$ & $\textcolor{blue}{+1.6{\pm}.5}$
& $+5.3{\pm}1.7$ & $+.8{\pm}.4$ \\
& ViT
& $-.3{\pm}.3$ & $-.1{\pm}.1$
& $\textcolor{red}{-11.8{\pm}5.7}$ & $\textcolor{red}{-4.7{\pm}2.3}$
& $\textcolor{blue}{+4.0{\pm}4.4}$ & $\textcolor{blue}{+1.3{\pm}1.1}$
& $+3.1{\pm}2.4$ & $+.8{\pm}.6$ \\

\multirow{2}{*}{CheXpert}
& ResNet
& $-1.7{\pm}1.3$ & $-.8{\pm}.5$
& $\textcolor{red}{-10.5{\pm}5.5}$ & $\textcolor{red}{-3.6{\pm}1.7}$
& $+.1{\pm}4.5$ & $+.3{\pm}1.3$
& $\textcolor{blue}{+5.2{\pm}3.9}$ & $\textcolor{blue}{+1.4{\pm}1.3}$ \\
& ViT
& $0.0{\pm}2.2$ & $-1.0{\pm}.3$
& $\textcolor{red}{-23.0{\pm}8.0}$ & $\textcolor{red}{-9.7{\pm}1.5}$
& $\textcolor{blue}{+12.8{\pm}7.6}$ & $\textcolor{blue}{+3.2{\pm}.5}$
& $+6.5{\pm}1.6$ & $+.8{\pm}.1$ \\

\multirow{2}{*}{Waterbirds}
& ResNet
& $-.9{\pm}1.0$ & $-1.0{\pm}.3$
& $\textcolor{red}{-39.6{\pm}4.7}$ & $\textcolor{red}{-20.3{\pm}1.4}$
& $\textcolor{blue}{+5.4{\pm}1.5}$ & $\textcolor{blue}{+3.7{\pm}.7}$
& $+.6{\pm}.4$ & $+.8{\pm}.2$ \\
& ViT
& $-7.1{\pm}2.0$ & $-1.5{\pm}.2$
& $\textcolor{red}{-38.3{\pm}3.2}$ & $\textcolor{red}{-14.5{\pm}1.9}$
& $\textcolor{blue}{+9.7{\pm}.4}$ & $\textcolor{blue}{+3.7{\pm}.4}$
& $+3.4{\pm}2.0$ & $+1.0{\pm}.9$ \\

\multirow{2}{*}{Camelyon17}
& ResNet
& $-1.5{\pm}.9$ & $-.8{\pm}.2$
& $\textcolor{red}{-15.2{\pm}3.3}$ & $\textcolor{red}{-6.6{\pm}1.4}$
& $\textcolor{blue}{+3.5{\pm}1.6}$ & $+1.0{\pm}.3$
& $+3.2{\pm}.4$ & $\textcolor{blue}{+1.1{\pm}1.0}$ \\
& ViT
& $+.1{\pm}1.1$ & $-.1{\pm}.3$
& $\textcolor{red}{-8.8{\pm}.3}$ & $\textcolor{red}{-3.4{\pm}.4}$
& $\textcolor{blue}{+3.1{\pm}.9}$ & $\textcolor{blue}{+.7{\pm}.3}$
& $+1.8{\pm}.7$ & $+.7{\pm}.4$ \\

\multirow{2}{*}{ISIC2019}
& ResNet
& $-2.5{\pm}2.9$ & $-.7{\pm}.6$
& $\textcolor{red}{-9.1{\pm}6.4}$ & $\textcolor{red}{-3.1{\pm}1.1}$
& $\textcolor{blue}{+1.0{\pm}2.7}$ & $\textcolor{blue}{+.3{\pm}.7}$
& $-.5{\pm}1.5$ & $-.1{\pm}.8$ \\
& ViT
& $-3.3{\pm}1.9$ & $+.1{\pm}.6$
& $\textcolor{red}{-8.3{\pm}4.7}$ & $\textcolor{red}{-.9{\pm}1.2}$
& $\textcolor{blue}{+7.4{\pm}5.8}$ & $+.3{\pm}.8$
& $\textcolor{blue}{+7.4{\pm}3.2}$ & $\textcolor{blue}{+1.0{\pm}.2}$ \\

\bottomrule
\end{tabular}
}
\vspace{-5pt}
\end{table*}

%% file: Tables/iterative_interventions.tex
\begin{table*}[t]
\centering
\caption{
\textbf{Iterative interventions reduce shortcut alignment and can decrease subgroup disparities beyond a single intervention.} Values are mean changes relative to no intervention across independently trained models. The $\bar{\rho}_{\mathrm{sc}_0}$ stage is the last iteration with a positive mean shortcut partial correlation across test images (or closest to $0$ if mean across all iterations is negative). For iteration $1$ and $5$, we also show the mean shortcut partial correlation $\bar{\rho}_{\mathrm{sc}}$. Blue denotes the largest improvement per dataset.
}
\label{tab:iterative_value_interventions}
\resizebox{\textwidth}{!}{%
\begin{tabular}{llrrrrrrrr}
\toprule
Dataset & Source
& \multicolumn{3}{c}{Iteration $1$}
& \multicolumn{2}{c}{$\bar{\rho}_{\mathrm{sc}_0}$}
& \multicolumn{3}{c}{Iteration $5$} \\
\cmidrule(lr){3-5}
\cmidrule(lr){6-7}
\cmidrule(lr){8-10}
&
& $\bar{\rho}_{\mathrm{sc}}$ & $\Delta\mathrm{LPG}_{\mathrm{red}}$ & $\Delta$Acc.
& $\Delta\mathrm{LPG}_{\mathrm{red}}$ & $\Delta$Acc.
& $\bar{\rho}_{\mathrm{sc}}$ & $\Delta\mathrm{LPG}_{\mathrm{red}}$ & $\Delta$Acc. \\
\midrule

\multirow{2}{*}{CelebA}
& Shortcut Group
& +.110
& +4.0
& \textcolor{blue}{+1.3}
& +7.8
& +1.2
& -.025
& \textcolor{blue}{+9.8}
& +.1 \\
& Image
& +.112
& +3.1
& +.8
& +2.7
& +.7
& -.024
& +1.1
& +.7 \\

\midrule

\multirow{2}{*}{CheXpert}
& Shortcut Group
& +.062
& +12.8
& \textcolor{blue}{+3.2}
& \textcolor{blue}{+20.3}
& -.4
& +.003
& \textcolor{blue}{+20.3}
& -.4 \\
& Image
& +.035
& +6.5
& +.8
& +6.5
& +.8
& -.087
& +12.6
& +1.9 \\

\midrule

\multirow{2}{*}{Waterbirds}
& Shortcut Group
& +.036
& +9.7
& +3.7
& +13.9
& \textcolor{blue}{+5.2}
& -.019
& \textcolor{blue}{+17.7}
& +4.5 \\
& Image
& +.022
& +3.4
& +1.0
& +3.4
& +1.0
& -.074
& +2.6
& +.6 \\

\midrule

\multirow{2}{*}{Camelyon17}
& Shortcut Group
& +.058
& +3.1
& +.7
& \textcolor{blue}{+6.6}
& \textcolor{blue}{+1.1}
& +.051
& \textcolor{blue}{+6.6}
& \textcolor{blue}{+1.1} \\
& Image
& +.002
& +1.8
& +.7
& +1.8
& +.7
& -.096
& +1.7
& +.6 \\

\midrule

\multirow{2}{*}{ISIC2019}
& Shortcut Group
& +.079
& \textcolor{blue}{+7.4}
& +.3
& +6.5
& +1.5
& +.078
& +6.5
& +1.5 \\
& Image
& +.011
& \textcolor{blue}{+7.4}
& +1.0
& \textcolor{blue}{+7.4}
& +1.0
& -.072
& +4.1
& \textcolor{blue}{+1.6} \\

\bottomrule
\end{tabular}}
\vspace{-10pt}
\end{table*}

%% file: sections/discussion.tex
\section{Discussion}
\label{sec:discussion}

Shortcut groups provide an inspectable summary of spatial shortcut evidence. The visual prototypes presented in Section~\ref{sec:exp-patterns} show that grouping image-level shortcut and task contribution maps reveals recurring spatial patterns that are not visible from a single dataset-level map. The representative examples make these patterns easier to inspect because each shortcut group is shown together with its error rate, label composition, and attribute composition. This is especially useful in medical datasets, where a high-error group with shortcut contribution in peripheral or non-anatomical regions can direct the audit toward acquisition, workflow, or source-related cues. Beyond visualisation, the same prototypes provide a compact way to collect candidate shortcut types across a dataset (Appendix Section~\ref{sec:supp-contribution-map-figures}). For medical datasets such as \textsc{CheXpert}, \textsc{Camelyon17}, and \textsc{ISIC2019}, this helps find multiple attribute-specific shortcut and task patterns.

Shortcut groups also support targeted failure inspection. 
The held-out risk-ranking experiment (Section~\ref{sec:exp-patterns}) shows that shortcut group risk scores recover a larger fraction of total errors than random inspection across datasets, architectures, attribution methods, and grid resolutions. This demonstrates that the discovered groups are not only visually interpretable, but also useful for selecting smaller subsets for manual review.

Input-level interventions highlight the distinction between task and shortcut contribution maps. Masking high task-contribution regions usually causes large drops in accuracy and increases in LPG, which is consistent with these regions containing evidence needed for the intended prediction. Masking high shortcut-contribution regions has weaker and more dataset-dependent effects as the performance often remains closer to the no-intervention performance or improves sometimes. This difference indicates that shortcut contribution maps capture evidence that can affect predictions, but whose removal depends on whether the shortcut evidence is separable from task evidence, redundant, or already mixed into internal features.

Internal interventions show that shortcut-group-level combined interventions are generally more effective than dataset-level interventions, while image-level interventions are strongest in some cases. Combined shortcut suppression and task amplification is effective in reducing performance disparities across all datasets. Iterative interventions further test the actionability of the contribution maps and the shortcut alignment score. When shortcut suppression reduces the mean shortcut alignment score while preserving or improving performance, it indicates that shortcut-aligned evidence can be reduced without heavily attenuating task-aligned evidence.

Overall, spatial shortcut group discovery enables visual inspection through shortcut group prototypes and exemplars, targeted auditing through shortcut group risk scores, and intervention-based characterisation through input and feature-space tests.

\subsection{Limitations}
\label{sec:limitations}

In our experiments, the shortcut patterns are linked to a single sensitive attribute. Real world scenarios may involve several sensitive attributes, including demographic, acquisition, site, scanner, source, and context variables. We envisage that, in a practical audit workflow, slice discovery methods \citep{eyuboglu2022domino} can first suggest candidate attributes, metadata variables, or failure subsets to investigate, after which our framework can be applied to these candidates one by one to obtain attribute-specific spatial shortcut group prototypes. Comparing the resulting prototype sets can then help identify whether different candidate attributes are associated with shared or distinct shortcut patterns.

The grid-based partitioning used for contribution maps can be affected by spatial alignment across the images, especially when the contribution maps are aggregated across the dataset or when forming a smaller number of shortcut groups. This might be suitable for approximately aligned datasets such as \textsc{CelebA} and \textsc{CheXpert}, but may affect datasets with larger spatial variability such as \textsc{Waterbirds}, \textsc{Camelyon17}, and \textsc{ISIC2019}. While our analyses provide meaningful shortcut group visual patterns, risk-based subset identification and actionable interventions, we do not conduct analyses on how much the image alignment across the dataset can affect the dataset-level inspection and leave this as future work.

The positive contribution maps contain all regions that support positive conditional rank alignment. This includes both regions that the two relevant models rank more highly than predicted from the conditioning model and regions that both rank less highly than predicted. Consequently, positive contribution should not be interpreted as absolute model importance or as ground-truth causal shortcut or task evidence. Our quadrant analysis (Appendix Section~\ref{supp:sec-quadrants}) shows that these cases can exhibit different spatial structure, but determining which components provide the most meaningful evidence for different auditing objectives is left as future work.

%% file: sections/conclusion.tex
\section{Conclusion}
\label{sec:conclusion}

We have introduced \emph{spatial shortcut group discovery}, a framework for auditing shortcut reliance in vision models. The framework obtains image-level shortcut contribution maps and task contribution maps, then summarises recurring joint shortcut and task contribution patterns into shortcut group-level contribution maps.

Across \textsc{CelebA}, \textsc{CheXpert}, \textsc{Waterbirds}, \textsc{Camelyon17}, and \textsc{ISIC2019}, and across ResNet and ViT models, the discovered shortcut groups provide compact and inspectable summaries of shortcut contribution and task contribution. The resulting prototypes and representative examples show that shortcut contribution can be spatially shared, shortcut group specific, image specific, or entangled with task contribution depending on the dataset and model. The shortcut groups can also be used to identify small subsets containing a large proportion of total test errors, making them useful for auditing. The intervention experiments further show that the shortcut and task contribution maps contain actionable spatial patterns, and that shortcut group-level maps can continue to guide interventions.

Overall, spatial shortcut pattern discovery connects visual inspection, auditing subset identification, and test-time intervention.

%% file: sections/acknowledgements.tex
\section{Acknowledgements}
This research was supported by the UK Engineering and Physical Sciences Research Council (EPSRC) [Grant reference number EP/Y035216/1] Centre for Doctoral Training in Data-Driven Health (DRIVE-Health) at King’s College London. This work was also supported by the SkincAIr project, funded by the Global Health EDCTP3 Joint Undertaking under the Horizon Europe research and innovation programme (Grant Agreement No. 101190743). Views and opinions expressed are however those of the author(s) only and do not necessarily reflect those of the European Union or EDCTP3. Neither the European Union nor the granting authority can be held responsible for them.

%% file: sections/appendix.tex
\section{Implementation details}
\label{sec:supp-impl}

\paragraph{Dataset construction.}
We construct fixed dataset splits and use them throughout all experiments. Training sets are sampled to match the subgroup compositions reported in Table~\ref{tab:train_splits}. If a subgroup contains fewer examples than the specified sample size, all available examples of that subgroup are used.

For \textsc{Camelyon17}, we retain patches with valid binary labels and complete patient, node, centre, and coordinate metadata. Patches with very little image detail are excluded using a Laplacian-variance threshold of $7.5\times10^{-5}$. The dataset splits are patient-disjoint, with patches sampled across patients. For \textsc{ISIC2019}, if lesion-level metadata are available, all images of the same lesion are assigned to a single split; otherwise, splits are disjoint at the image level.

\paragraph{Model training.}
For each dataset, the audited task model, bias-reduced baseline model, and sensitive attribute model share the same architecture i.e. all three are either ResNet or ViT models. We use ImageNet-pretrained ResNet-50 models from \texttt{timm}~\citep{rw2019timm} and ImageNet-pretrained~\citep{deng2009imagenet} ViT-B/16 models from \texttt{torchvision}~\citep{torchvision2016}. We resize inputs to $224\times224$ and use three-channel images for all datasets. We apply random augmentations such as horizontal flip, vertical flip, random rotation, colour jitter and random grayscaling for \textsc{Camelyon17} and apply horizontal flip, vertical flip and random rotation for \textsc{ISIC2019}.

We train all models with cross-entropy loss and AdamW~\citep{loshchilov2017decoupled}. We also use early stopping on validation F1-score.

\paragraph{Attribution maps.}
We resize all attribution maps to $224\times224$ and aggregate them over regular image grids, including $8\times8$, $16\times16$, $32\times32$, and $56\times56$. For each model, the attribution map is computed with respect to its predicted class and therefore represents the evidence contributing to that prediction.

Code is available at \url{https://github.com/acharaakshit/shortcut-groups}.

\section{Robustness of shortcut groups}
\label{sec:supp-robustness}

In this section, we test whether the discovered shortcut groups are stable across independently trained models and whether they depend strongly on auxiliary model choice.

\subsection{Stability of shortcut groups across independently trained models}
\label{sec:supp-stability}

We evaluate whether the shortcut groups are stable by assessing the shortcut group derived risk score (see Section~\ref{sec:exp-patterns}) based subsets across independently trained models ($f_{BA}, f_{TS}, f_{SA}$). We first create fitting and subgroup-balanced held-out datasets as in Section~\ref{sec:exp-patterns} and then for each model, we inspect the top-$20\%$ based on the risk score to check if high error subsets are consistently selected. Additionally, for K-means, images in the same shortcut group have identical group-derived risk scores. If the inspection budget (top-20\%) includes only part of a group, images closest to the group centroid are selected.

\input{Tables/seed_stability}

\Cref{tab:supp_seed_stability_and_overlap_k8} shows that shortcut groups consistently select higher-error subsets across all five datasets and models. Top-$20\%$ subsets based on both K-means and NMF derived risk scores find substantial \% of total held-out errors, but NMF based subsets contain the highest \% of total held-out errors in every dataset and model. For ViT, the NMF subset contains about $63.5\%$ errors on \textsc{CelebA}, $57.1\%$ on \textsc{CheXpert}, $48.1\%$ on \textsc{Waterbirds}, $43.4\%$ on \textsc{Camelyon17}, and $40.2\%$ on \textsc{ISIC2019}. For ResNet, the NMF subset contains about $59.3\%$ errors on \textsc{CelebA}, $40.2\%$ on \textsc{CheXpert}, $73.8\%$ on \textsc{Waterbirds}, $46.2\%$ on \textsc{Camelyon17}, and $31.5\%$ on \textsc{ISIC2019}.

\Cref{tab:supp_seed_stability_and_overlap_k8} also shows how similar the selected subsets are across independently trained models. For two independently trained models $a$ and $b$, let $S_a$ and $S_b$ be their selected sets. \emph{Intersection} is defined as
\[
\frac{|S_a\cap S_b|}{|S_a|},
\]
which measures what fraction of one independently trained model's selected images are also selected by using the other model.

The selected examples are not identical across models, which is expected because independently trained models can rely on slightly different spatial evidence. However, NMF based subsets are more stable than K-means on every dataset, with the exception of \textsc{ISIC2019} for ResNet where K-means has a slightly better overlap. Overall, for NMF based subsets, intersection values range from $.259$ on \textsc{ISIC2019} with ResNet to $.526$ on \textsc{CheXpert} with ViT, meaning that approximately $26\%$ to $53\%$ of selected examples are shared across pairs of independently trained models. This shows that exact image-level rankings vary across independently trained models, but shortcut group structure still provides a stable signal for identifying higher-risk subsets.

\subsection{Stability of shortcut groups across independently trained auxiliary-models}
\label{sec:supp-aux}

We know that the shortcut and task contribution maps depend on the audited task model $f_{\mathrm{TS}}$, sensitive attribute model $f_{\mathrm{SA}}$, and bias-reduced baseline/reference model $f_{\mathrm{BA}}$. In this section, we study how stable the discovered shortcut group prototypes and risk scores are to retraining either of the auxiliary models $f_{\mathrm{SA}}$ and $f_{\mathrm{BA}}$.

Specifically, we hold the audited task model $f_{\mathrm{TS}}$ fixed and retrain one auxiliary model at a time across independently trained models. The \emph{Changed model} column in \Cref{tab:supp_aux_prototype_sensitivity} indicates whether $f_{\mathrm{BA}}$ or $f_{\mathrm{SA}}$ is replaced with a different independently trained model, while the other auxiliary model is held fixed. For each retraining, we recompute the shortcut and task contribution maps, form the corresponding joint shortcut and task contribution representation, and obtain NMF shortcut group prototypes.

\Cref{tab:supp_aux_prototype_sensitivity} shows that, across datasets and models, the results under auxiliary model retraining are generally similar to the Non-Aux (\%). This shows that retrained auxiliary models still identify high-error subsets using the shortcut group risk scores. We also compare the learned shortcut group prototypes across independently trained auxiliary models. Since shortcut groups are unordered, we match them across independently trained models using the similarity of their shortcut and task prototype maps. We then report \emph{shortcut similarity} and \emph{task similarity} using cosine similarity between the matched prototype maps. The shortcut and task similarities are consistently high across datasets and models, indicating that the recurring spatial shortcut and task patterns are stable under independently trained auxiliary-models.

\input{Tables/auxiliary}

\section{Design choices and ablation studies}
\label{sec:supp-design}

In this section, we study the main design choices behind shortcut group discovery and evaluate whether they affect the conclusions of our results presented in Section~\ref{sec:experiments}.

\subsection{Decomposing shortcut and task alignment}
\label{supp:sec-quadrants}

The region contribution $c_i(x)=z_{1,i}z_{2,i}$ can be understood by separating the signs of the two standardised residuals already defined in Section~\ref{sec:method-contribution-maps}. For shortcut alignment, $z_{1,i}$ and $z_{2,i}$ are obtained from $\mathbf{r}_{u\mid w}$ and $\mathbf{r}_{v\mid w}$, respectively. For task alignment, they are obtained from $\mathbf{r}_{u\mid v}$ and $\mathbf{r}_{w\mid v}$. In either case, $c_i(x)>0$ when the residuals have the same sign and $c_i(x)<0$ when they have opposite signs. The same-sign quadrants contribute positively to residual-rank agreement, whereas the off-diagonal quadrants contribute negatively.

\begin{figure*}[t]
\centering
\includegraphics[width=\textwidth]{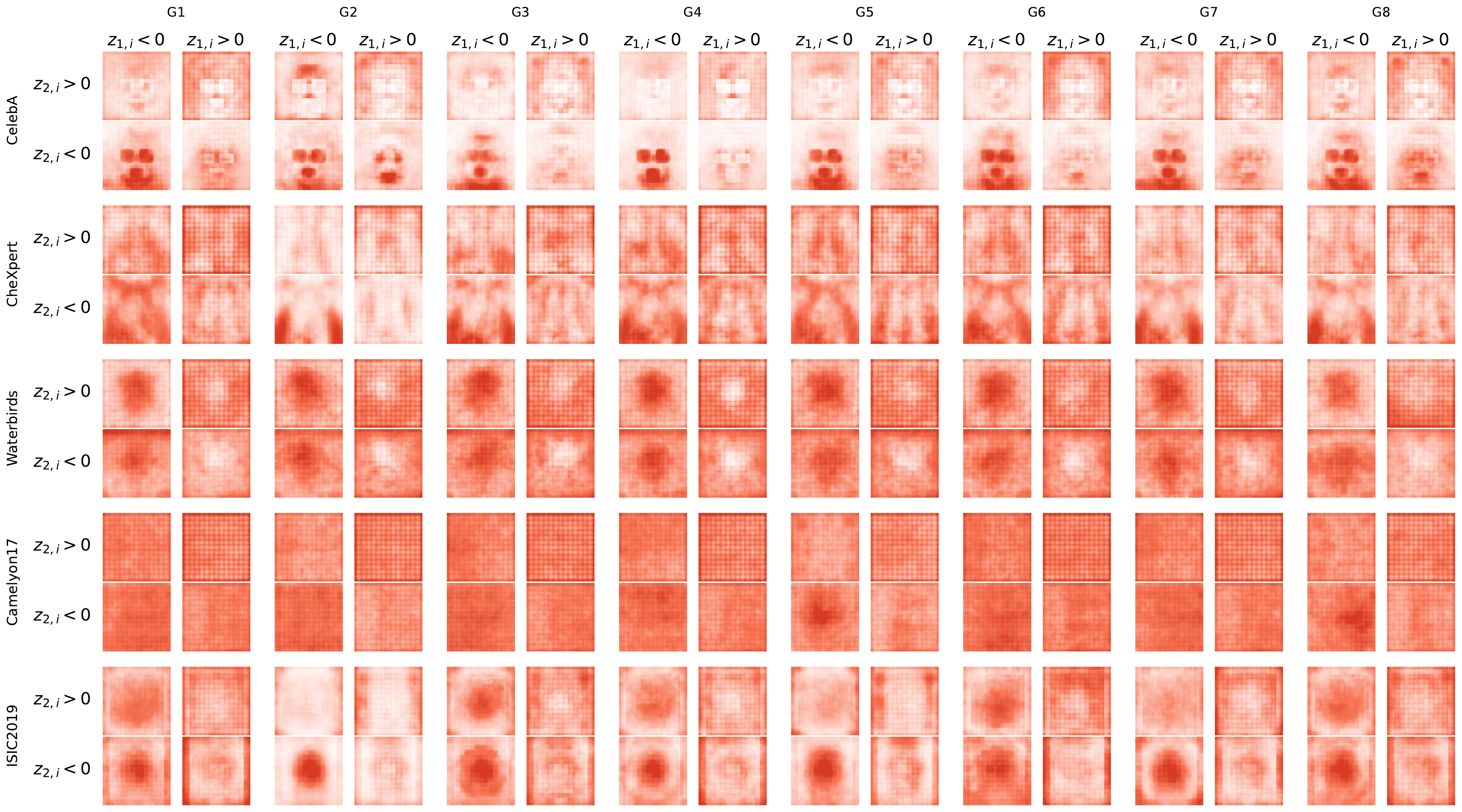}
\caption{\textbf{Quadrant decomposition of shortcut alignment score using ViT based shortcut groups.} Orange shows $|z_{1,i}z_{2,i}|$ within the indicated quadrant.}
\label{fig:supp-quadrant-decomposition-vit}
\vspace{-5pt}
\end{figure*}

\begin{figure*}[t]
\centering
\includegraphics[width=\textwidth]{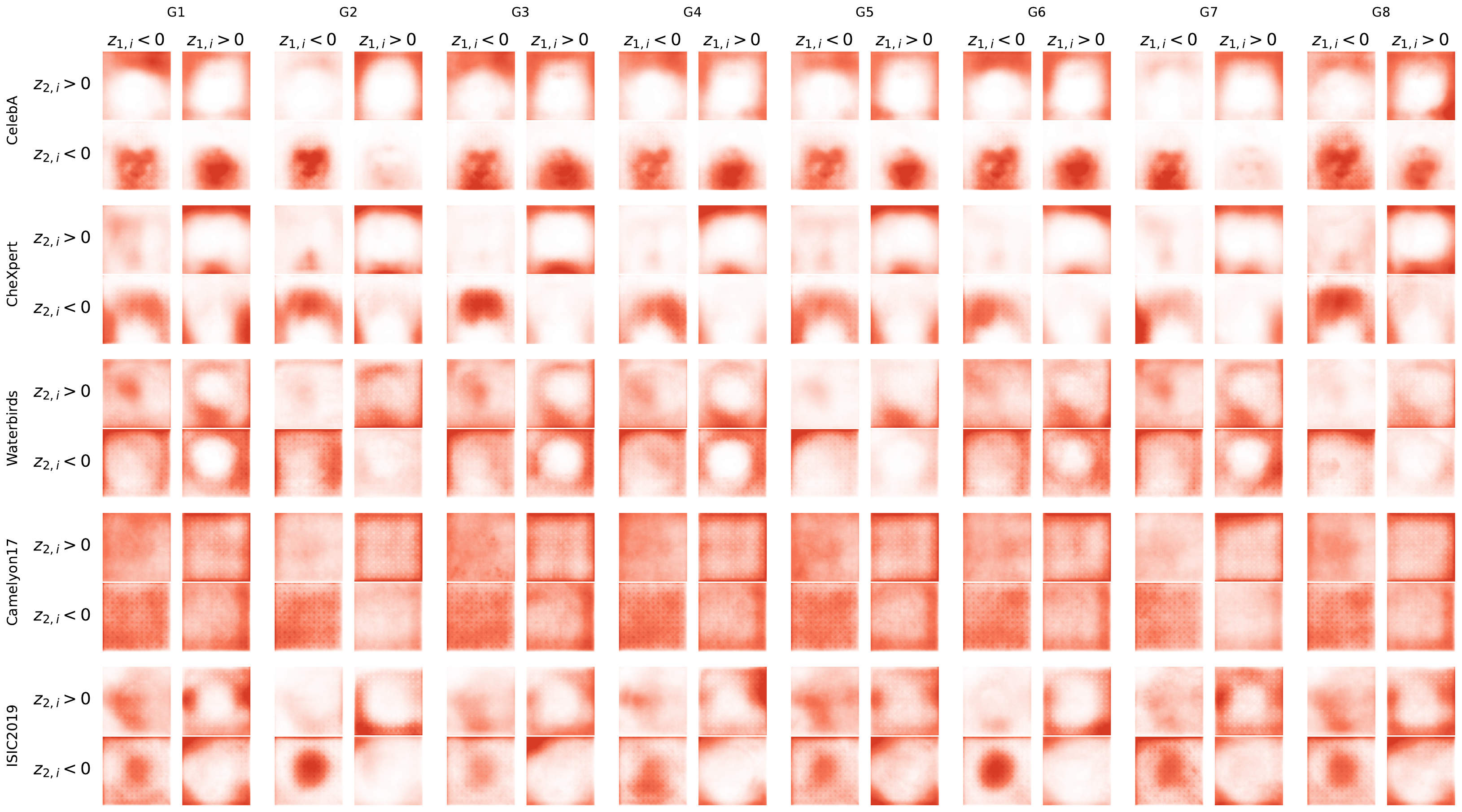}
\caption{\textbf{Quadrant decomposition of shortcut alignment score using ResNet based shortcut groups.} Orange shows $|z_{1,i}z_{2,i}|$ within the indicated quadrant.}
\label{fig:supp-quadrant-decomposition-resnet}
\vspace{-5pt}
\end{figure*}

\begin{figure*}[t]
\centering
\includegraphics[width=\textwidth]{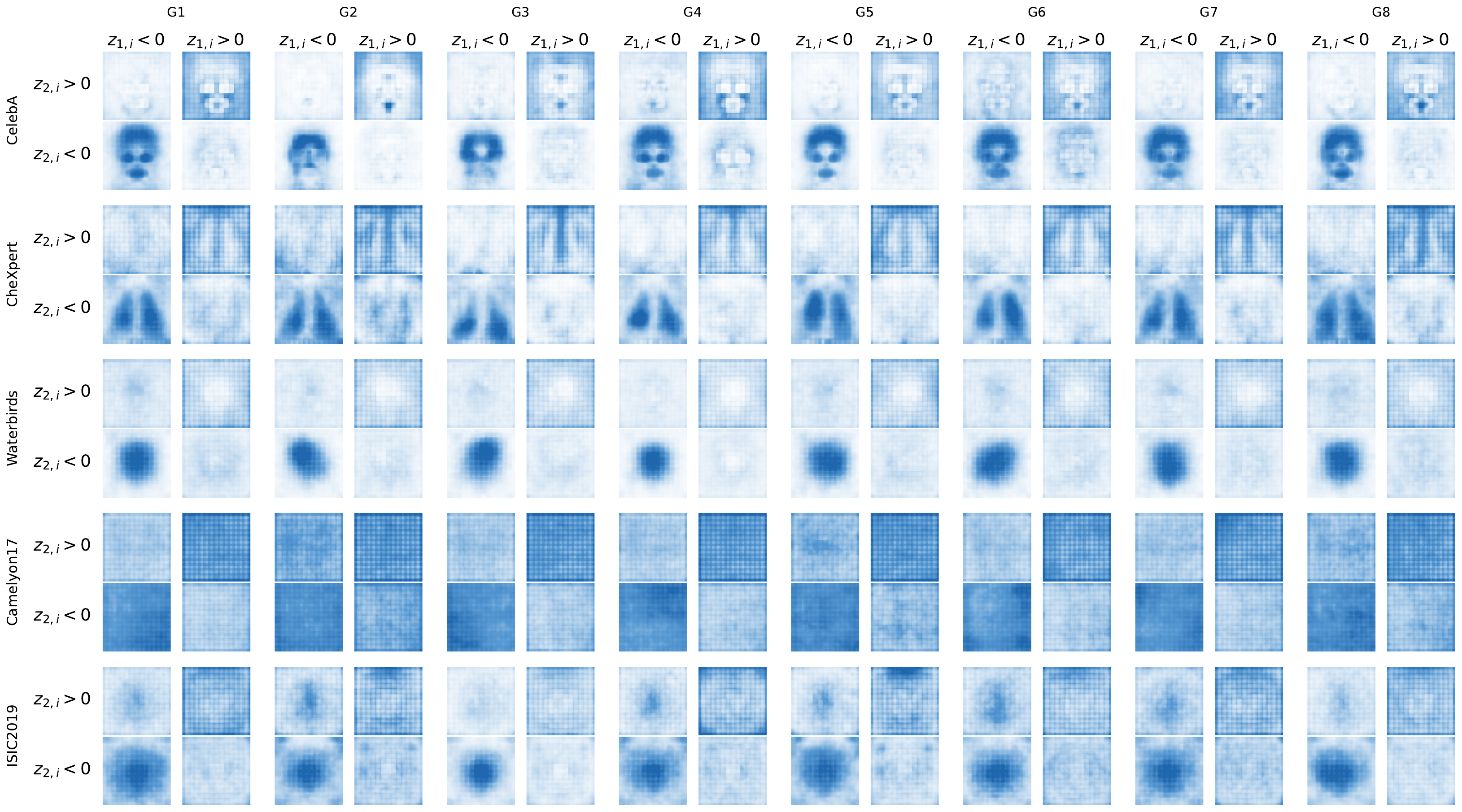}
\caption{\textbf{Quadrant decomposition of task alignment score using ViT based shortcut groups.} Blue shows $|z_{1,i}z_{2,i}|$ within the indicated quadrant.}
\label{fig:supp-quadrant-decomposition-task-vit}
\vspace{-5pt}
\end{figure*}

\begin{figure*}[t]
\centering
\includegraphics[width=\textwidth]{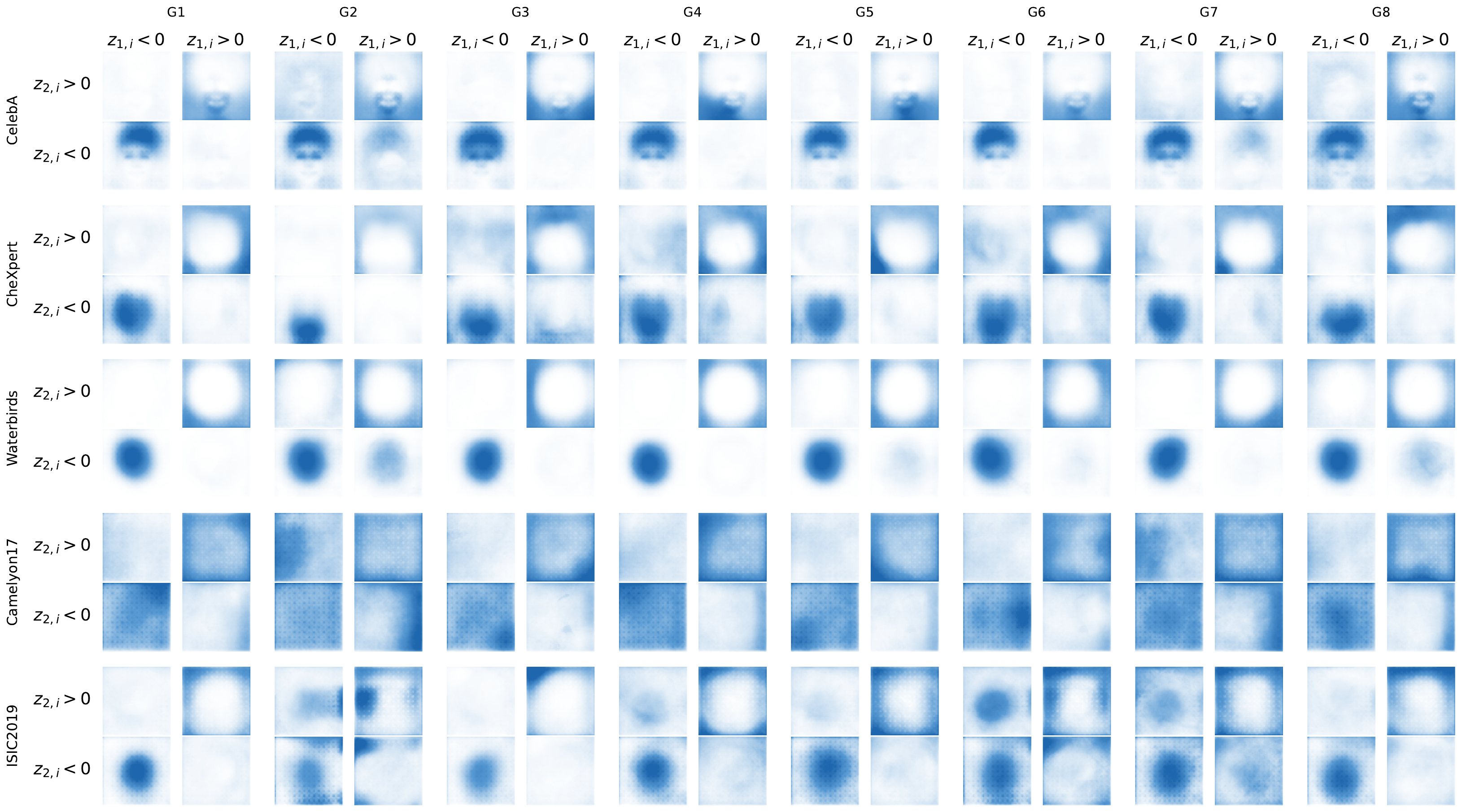}
\caption{\textbf{Quadrant decomposition of task alignment score using ResNet based shortcut groups.} Blue shows $|z_{1,i}z_{2,i}|$ within the indicated quadrant.}
\label{fig:supp-quadrant-decomposition-task-resnet}
\vspace{-5pt}
\end{figure*}

To visualise the contributions associated with each quadrant, we use the NMF based shortcut group memberships ($K=8$) obtained from the joint shortcut and task contribution representation and decompose the shortcut-alignment contribution of every existing group into all four quadrants. Figures~\ref{fig:supp-quadrant-decomposition-vit},~\ref{fig:supp-quadrant-decomposition-resnet},~\ref{fig:supp-quadrant-decomposition-task-vit},~\ref{fig:supp-quadrant-decomposition-task-resnet} show the shortcut and task contributions for each quadrant for all five datasets and both models (ViT and ResNet). In these figures, each $2\times2$ block corresponds to one fixed shortcut group whose columns indicate the sign of $z_{1,i}$ and its rows indicate the sign of $z_{2,i}$, forming all the four quadrants. The jointly higher-ranked case $z_{1,i}<0,\ z_{2,i}<0$ therefore appears in the lower-left cell. Across datasets and models, this quadrant contains the principal interpretable visual structure, recovering recognisable face, non-lung (e.g. breast shadow) and background from the shortcut contribution maps and hair, lung, bird from the task contribution maps. The $z_{1,i}>0,\ z_{2,i}>0$ maps are generally more diffuse or concentrated around borders and grid structure, while the off-diagonal maps show where the two conditional rankings disagree. This indicates that the main visual evidence in the shortcut groups comes from regions that both models rank more highly than conditionally expected. However, restricting the contribution definition to this quadrant would change the quantity from all positive support for shortcut or task alignment to shared high-priority agreement.

\input{Tables/quadrant_error}

\Cref{tab:supp_quadrant_errors_found_k8} shows a comparison of positive $(m_{sc}(z_1<0, z_2<0) + m_{sc}(z_1>0, z_2>0), m_{task}(z_1<0, z_2<0) + m_{task}(z_1>0, z_2>0))$, negative $(m_{sc}(z_1<0, z_2>0) + m_{sc}(z_1>0, z_2<0), m_{task}(z_1<0, z_2>0) + m_{task}(z_1>0, z_2<0))$ , and combined $(m_{sc}(z_1<0, z_2<0) + m_{sc}(z_1>0, z_2>0), m_{sc}(z_1<0, z_2>0) + m_{sc}(z_1>0, z_2<0), m_{task}(z_1<0, z_2<0) + m_{task}(z_1>0, z_2>0), m_{task}(z_1<0, z_2>0) + m_{task}(z_1>0, z_2<0))$ and individual quadrant-based representations. Shortcut groups are obtained for each of the representation by applying NMF ($K=8$) to compute associated (held-out) risk scores. It shows that no individual quadrant is uniformly best for error capture and the risk score derived from groups based on the combined representation result in subsets that contain the highest errors across all datasets and models. Additionally, the negative and positive representation based subsets contain similar percentages of errors across the datasets (except for \textsc{Waterbirds} subsets for ResNet where negative representations contain higher errors). As discussed in Section~\ref{sec:method-contribution-maps}, this does not imply that the contributions from the off-diagonal quadrants are anti-shortcut or anti-task evidence. Rather, these spatial patterns measure conditional-rank disagreement, which can be predictive of model error even though it does not support the corresponding alignment score.

Overall, the $(-,-)$ quadrant provides the clearest qualitative spatial structure, whereas combining all four quadrants is more effective for prioritising failures. However, since the shortcut and task contribution maps are intended to localise positive regional contributions to the corresponding alignment score, we use the representation based on the two same-sign quadrants (joint shortcut and task representation) in the primary analyses.

\subsection{Varying the resolution of grid partitions}
\label{sec:supp-grid-var}

We next test whether the held-out auditing results depend on the spatial resolution used to form the regional contribution vectors. Similar to the Section~\ref{sec:exp-patterns}, we rank the held-out examples by the shortcut group risk score, then measure the percentage of total held-out set errors contained in the top 20\% selected examples.

\begin{figure}[t]
\centering
\includegraphics[width=\linewidth]{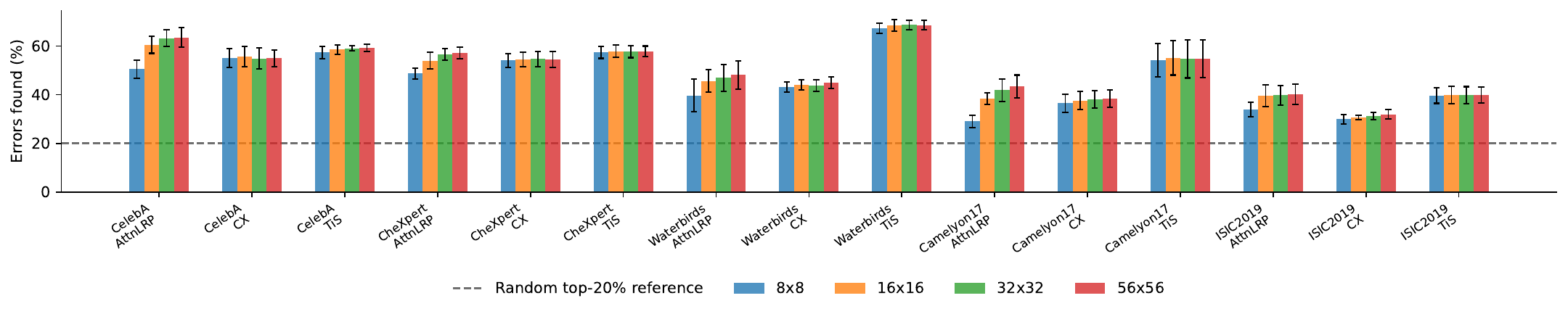}
\caption{\textbf{Risk-score-based error identification is robust across grid-partition resolutions.} Coloured bars represent the percentage of total errors captured within the top 20\% of samples ranked by risk score; the dashed line indicates the 20\% random baseline. Error bars denote the standard deviation across independently trained ViT models averaged over five (held-out) splits.
}
\label{fig:varying_grid_risk}
\end{figure}

Figure~\ref{fig:varying_grid_risk} shows that shortcut group risk score based subsets contain substantially more held-out errors than the random baseline across datasets, attribution methods, and grid resolutions. Under the random baseline, $20\%$ of images are expected to contain $20\%$ of the errors. The improvement is already visible at coarse resolutions such as $8\times 8$, and performance is generally stable from $16\times 16$ to $56\times 56$. Finer grids sometimes improve error recovery for AttnLRP, but no single grid resolution uniformly dominates across all datasets for the other attribution methods. This suggests that the shortcut group auditing signal is not an artefact of a particular grid partition.

\subsection{Choice of number of shortcut groups}
\label{sec:supp-k}

While we use $K=8$ shortcut groups in our main experiments, we do not interpret $K$ as the true number of shortcut mechanisms in the data. Instead, $K$ defines the number of spatial patterns that are shown, inspected, and compared. Therefore, $K$ can be based on the specific task requirements. In this section, following the main risk score experiment in Section~\ref{sec:exp-patterns}, we conduct a sensitivity analysis on the \% of errors found based on NMF based shortcut group derived risk score by varying the number of shortcut groups $K$.

\begin{figure}[t]
\centering
\includegraphics[width=0.8\linewidth]{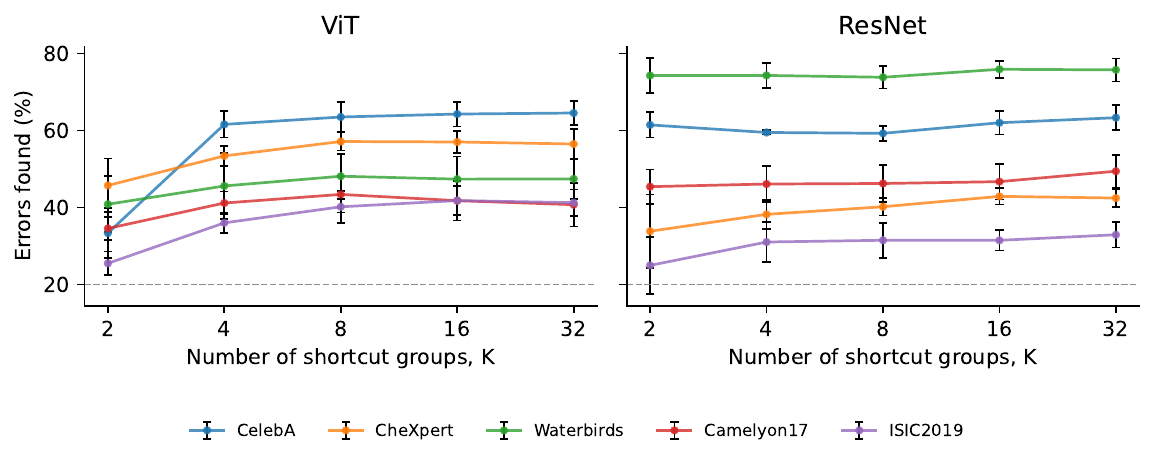}
\caption{\textbf{Increasing the number of shortcut groups beyond $K=8$ yields comparable error-subset identification.} The dashed line marks random selection at the same 20\% inspection budget. Values are first averaged across five different held-out sets and the error bars indicate the deviations across independently trained models.}
\label{fig:varying_k_risk}
\end{figure}

Figure~\ref{fig:varying_k_risk} shows that modest values of $K=8$ can help identify larger error subsets and at the same time, provide a compact set of shortcut groups. Therefore, choosing a higher $K$ does not automatically improve the audit and a modest value such as $K=8$ is a reasonable choice.

\subsection{Varying number of images for inspection}
\label{sec:supp-inspection-budget}

In this section, we study if the risk based subset selection consistently contains a substantial portion of total (held-out) errors across varying sizes of the selected subsets. We follow the risk experiment in Section~\ref{sec:exp-patterns} to create (five) fitting and held-out sets and vary the number of images selected for the inspection to measure how many errors are captured as the selected subset grows.

\begin{figure}[t]
\centering
\includegraphics[width=0.8\linewidth]{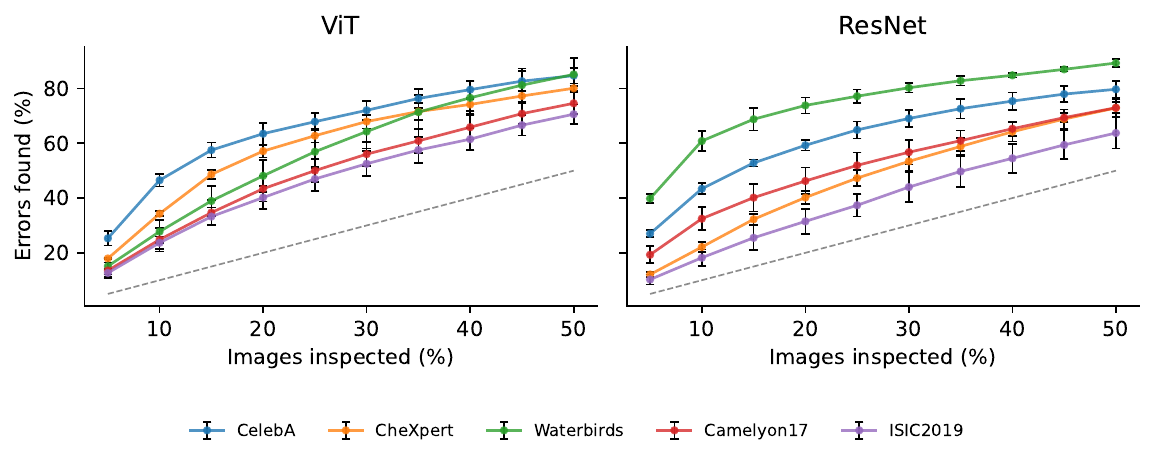}
\caption{\textbf{Risk-score-ranked subsets consistently capture higher error proportions across varying inspection budgets.} Curves show the percentage of total errors identified as the subset size increases; the dashed line represents the random selection baseline. Error bars denote standard deviations across independently trained models averaged over five evaluation splits.}
\label{fig:varying_n_risk}
\end{figure}

Figure~\ref{fig:varying_n_risk} shows the shortcut group risk score based selected subsets consistently contain substantially more errors than random selection for the same inspection budget averaged across independently trained models.

\subsection{Comparison of shortcut and task representation against alternative baselines for risk score based subset identification}
\label{sec:supp-joint-representation-baselines}

In this section, we additionally compare the risk score based subset identification based on the joint shortcut and task contribution representation with different baselines. We compare rankings derived from the joint shortcut and task representation $[m^{sc}(x);m^{task}(x)]$ with rankings based only on the shortcut alignment score $\rho_{sc}$, negative task alignment score $-\rho_{task}$ (lower task alignment is indicative of higher risk), shortcut contribution $m^{sc}(x)$, task contribution $m^{task}(x)$, or the raw audited task model attribution rankings ($\mathbf{s}(f_{TS},x,\hat{y}_{TS})$, risk score based on NMF $K=8$) over the held-out splits. For each ranking, we measure the percentage of total errors contained in the top $20\%$ of examples.

\input{Tables/risk_ablation}

\Cref{tab:error_capture_score_map_baselines} shows that the risk score derived from the negative task alignment score $-\rho_{task}$ results in subsets containing the highest \% of held-out set errors across all datasets and models except for \textsc{Waterbirds} on ViT model. The task score based subsets contain $64.8\%/63.5\%$ errors on \textsc{CelebA}, $60.8\%/43.0\%$ on \textsc{CheXpert}, $46.2\%/79.5\%$ on \textsc{Waterbirds}, $55.3\%/53.6\%$ on \textsc{Camelyon17}, and $44.2\%/34.5\%$ on \textsc{ISIC2019} for ViT/ResNet respectively. Therefore, high-risk examples are better identified by the task alignment score as compared to the joint shortcut and task contribution, individual shortcut/task contribution maps, shortcut alignment score, or raw audited-model attribution alone. However, in general, the risk score based on combined four quadrant representation based contribution maps results in subsets containing the highest errors as compared to all other tested risk-score based subsets (see Table~\ref{tab:supp_quadrant_errors_found_k8}).

\section{Additional intervention analyses}
\label{sec:supp-interventions}

This section provides additional analyses for the input and internal interventions.

\subsection{Input interventions for finer grid resolutions}
\label{sec:supp-occlusion-grid}

We evaluate input intervention for the $56\times56$ grid partition to test whether the effect of masking regions is the same as the main analysis in Table~\ref{tab:input_occlusion_summary}, where we use a $16\times16$ grid partition. 

\input{Tables/occlusion_appendix}

\Cref{tab:input_occlusion_56x56} shows that task masking results in substantial performance drops for all the datasets, models and contribution map sources. However, random masking at this finer resolution can also sometimes result in a severe performance drop compared to the $16\times16$ grid resolution masking, indicating that there could be a stronger distribution shift in the finer grid masking. Overall, shortcut masking performance is higher than task and random masking, and task masking results in a drop in performance. Therefore, masking task contribution regions consistently degrades performance, indicating that these regions contain evidence required for the intended prediction. In contrast, masking shortcut contribution regions is substantially less harmful and often preserves or improves performance, supporting the interpretation that shortcut contribution is concentrated in cues that are not essential for the task and may reflect non-generalisable evidence learned from the biased training distribution.

\subsection{Comparison of contribution map sources for interventions}
\label{sec:supp-dataset-vs-other-level}

Dataset-level contribution maps provide an aggregation of shortcut and task evidence, but this aggregation can smooth over spatial patterns that differ across images. To examine whether this matters during intervention, we apply the combined shortcut suppression and task amplification using dataset, shortcut group, and image-level maps. Table~\ref{tab:supp_combined_source_comparison} shows that dataset-level interventions are less effective than shortcut group and image-level interventions. Shortcut group-level interventions are the most effective across most datasets and architectures, while image-level interventions provide the strongest results in some cases. Overall, the comparison shows that averaging contribution maps across the entire dataset can reduce their effectiveness for intervention.

\input{Tables/source_comparison_intervention}

\subsection{Alternate internal intervention methods}
\label{sec:supp-alternate-intervention-methods}

In this section, we perform an intervention analysis across independently trained models and evaluate shortcut suppression and task amplification separately.

\input{Tables/intervention_seeds}

Table~\ref{tab:supp_intervention_seed_source_stability} shows that shortcut suppression is an effective intervention strategy for \textsc{CelebA}, \textsc{CheXpert} and \textsc{Waterbirds} as it generally reduces the LPG while preserving or improving the overall accuracy of the model. However, this suppression is not effective for the \textsc{Camelyon17} and \textsc{ISIC2019} where the LPG increases or is similar. As compared to shortcut suppression, task amplification has varied effects  on performance for \textsc{CelebA}, \textsc{CheXpert} and \textsc{Waterbirds} where it increases the LPG when image-level interventions are applied on ViT for \textsc{CelebA}, for \textsc{CheXpert} when both shortcut group and image-level interventions are applied on ViT. However, in general, task amplification improves LPG more than shortcut suppression for \textsc{Camelyon17} and \textsc{ISIC2019} datasets.

Overall, the combined shortcut suppression and task amplification is more effective across the datasets and models as compared to the shortcut suppression or task amplification alone. Therefore, we use the combined approach in our intervention experiments.

\subsection{Intervention target layer analysis}
\label{sec:supp-intervention-sites}

In this section, we study where internal intervention (combined shortcut suppression and task amplification) should be applied inside ResNet and ViT. In ResNets, we apply the contribution maps to convolutional feature maps after the final residual stage, the last two residual stages, last three residual stages or all residual stages. In ViTs, the same spatial contribution map is mapped to patch tokens and used to intervene on value vectors in the last layer, the last three layers, the last six layers, or all layers.

\input{Tables/intervention_layers}

\Cref{tab:intervention_site_appendix} shows that combined shortcut suppression and task amplification over the last two residual stages results in the highest average LPG reduction for ResNets across shortcut group-level and image-level contribution maps, with $\Delta\mathrm{LPG}_{\mathrm{red}}=+3.15\%$ and $\Delta$Acc. $=+1.1\%$. Although intervention over the last three residual stages gives the highest improvements for image-level maps, intervention over the last two stages gives better overall improvements across both sources. For ViTs, intervention across all value layers gives the highest average LPG reduction, with $\Delta\mathrm{LPG}_{\mathrm{red}}=+5.9\%$ and $\Delta$Acc. $=+1.35\%$. We therefore use the last two residual stages for ResNets and all value layers for ViTs in our intervention experiments.

\subsection{ViT intervention target block analysis}
\label{sec:supp-vit-pathways}

We next compare query (Q), key (K), value (V), and MLP interventions for ViT.

\input{Tables/vit_pathways}

Across contribution map sources, combined shortcut residual suppression and task residual amplification over value vectors result in the best intervention performance in \Cref{tab:vit_pathways_56x56}. Value $+$ MLP interventions generally give smaller improvements, while interventions involving $Q$, $K$ and MLP increase LPG and reduce accuracy.

Overall, internal interventions based on shortcut and task contribution maps alter model predictions. The source comparison indicates variation in shortcut structure across datasets, while the target block comparisons show that intervention over all value vectors is the most effective intervention. 

The differences between intervention targets can be interpreted from the self-attention computation
\[
O=\mathrm{softmax}\left(\frac{QK^\top}{\sqrt{d}}\right)V.
\]
Applying the spatial intervention to $Q + K$ modifies the attention weights, whereas applying it to $V$ modifies the information aggregated under those weights without directly changing the attention matrix. For the MLP, the intervention instead scales the feed-forward output of the patch tokens. Together with the target-block analysis (Section~\ref{sec:supp-intervention-sites}), these results show that applying the intervention to value vectors across all transformer blocks is the most effective configuration in our experiments. A more detailed analysis of block specific intervention effects is left to future work.

\subsection{Iterative interventions on ResNet}
\label{sec:supp-iterativeresnet}

In this section, we evaluate iterative interventions on ResNets, and summarise the results in Table~\ref{tab:supp_resnet_iterative_diagnostics}. The initial combined shortcut suppression and task amplification reduces LPG for every dataset and both contribution-map sources (except ISIC2019 image-level), with accuracy also improving in general. Subsequent shortcut suppression shows particularly consistent improvements on \textsc{CelebA}, and \textsc{Waterbirds}. The effects are less stable on \textsc{CheXpert}, \textsc{Camelyon17} and \textsc{ISIC2019},
where continued intervention reduces some of the initial performance improvements. At the $\rho_{\mathrm{sc}_0}$ iteration, shortcut group-level interventions show a positive LPG reduction in general, whereas the intervention becomes slightly detrimental on shortcut group source \textsc{CheXpert} and for both sources for \textsc{ISIC2019}. Additionally, similar to the iterative analysis on ViT (Table~\ref{tab:iterative_value_interventions}), the $\rho_{sc}$ at iteration 5 is lower than iteration 1. Overall, the ResNet results also show that iterative interventions can provide further reduction in performance disparities depending on the dataset and model.

\input{Tables/resnet_iterative_interventions}

\subsection{How do iterative interventions change the contribution maps?}
\label{sec:supp-itervariations}

In this section, we study how the shortcut and task contributions change over iterative internal interventions. Specifically, we visualise the contribution maps after iteratively applying per-image shortcut suppression, task amplification, combined shortcut suppression and task amplification and finally initial combined intervention followed by shortcut suppression. We compare the NMF based ($K=8$) shortcut group-level contribution maps (prototypes) at every iteration. 

\begin{figure}[t]
\centering
\includegraphics[width=\linewidth]{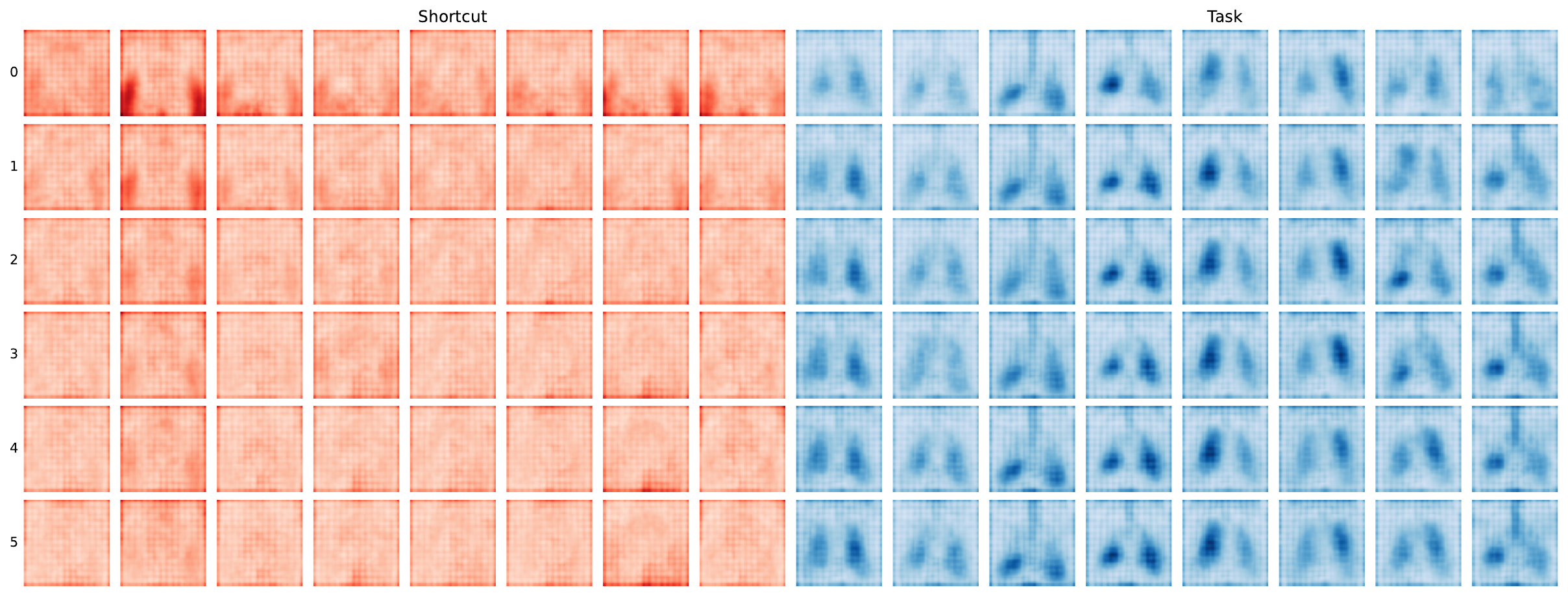}
\caption{\textbf{Shortcut contributions are suppressed over iterations whereas task contributions are comparatively stable.} Iterative contribution maps under shortcut suppression on \textsc{CheXpert} using ViT.}
\label{fig:celeba_iter_residual_suppression}
\end{figure}

\begin{figure}[t]
\centering
\includegraphics[width=\linewidth]{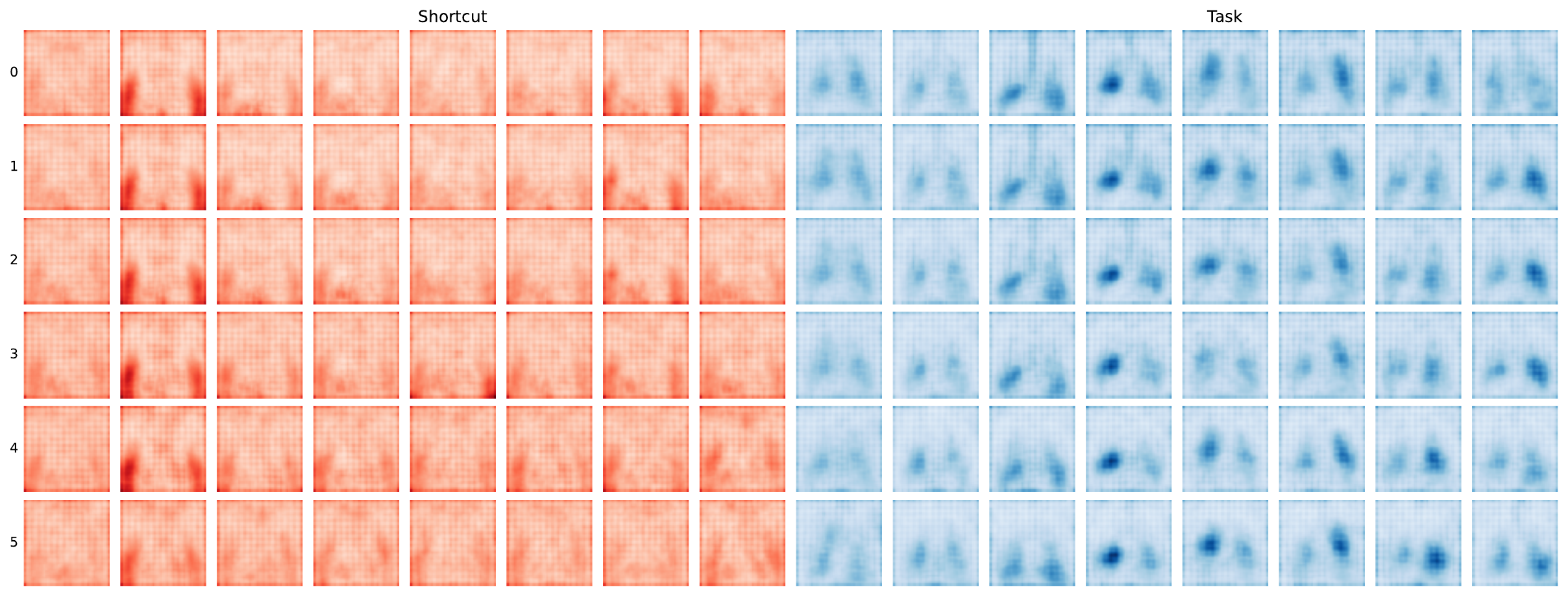}
\caption{\textbf{Shortcut contributions are comparatively stable whereas task contributions often vary across iterations.} Iterative contribution maps under task amplification on \textsc{CheXpert} using ViT.}
\label{fig:celeba_iter_suppression}
\end{figure}

\begin{figure}[t]
\centering
\includegraphics[width=\linewidth]{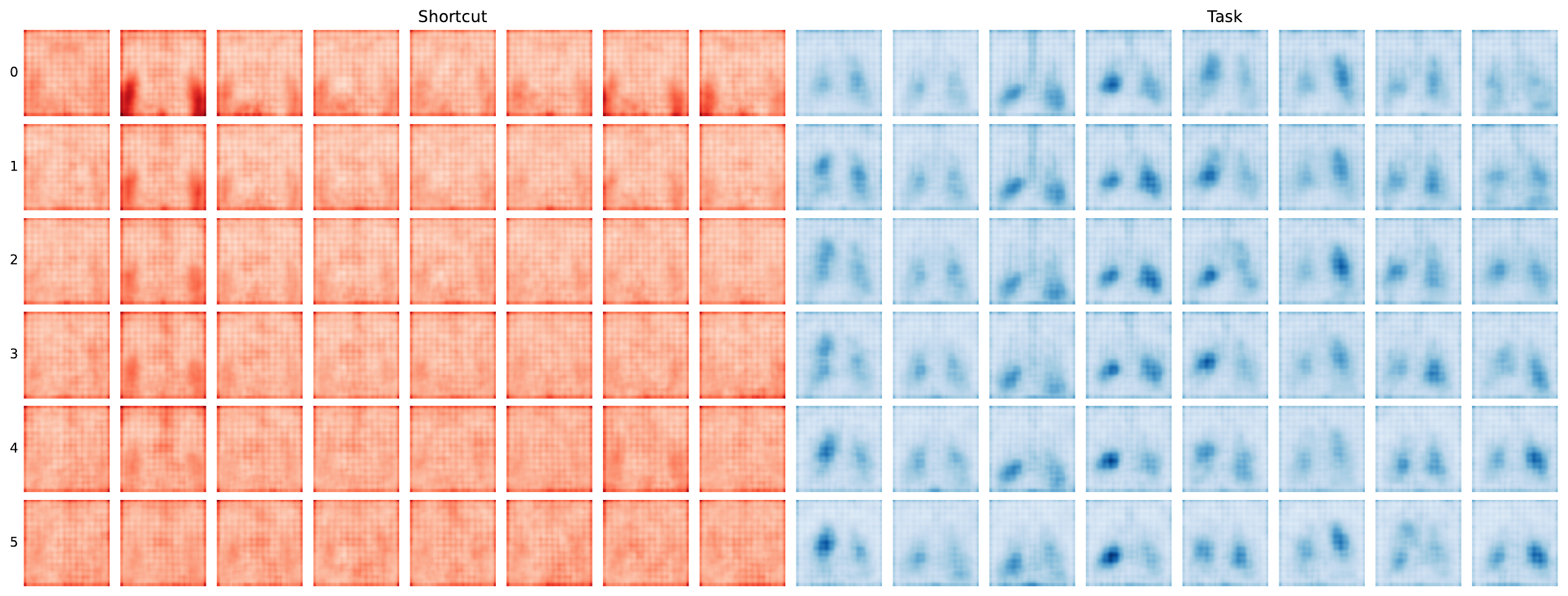}
\caption{\textbf{Shortcut contributions are suppressed over iterations whereas task contributions often vary.} Iterative contribution maps under combined shortcut suppression and task amplification on \textsc{CheXpert} using ViT.}
\label{fig:chexpert_iter_residual_suppression}
\end{figure}

\begin{figure}[t]
\centering
\includegraphics[width=\linewidth]{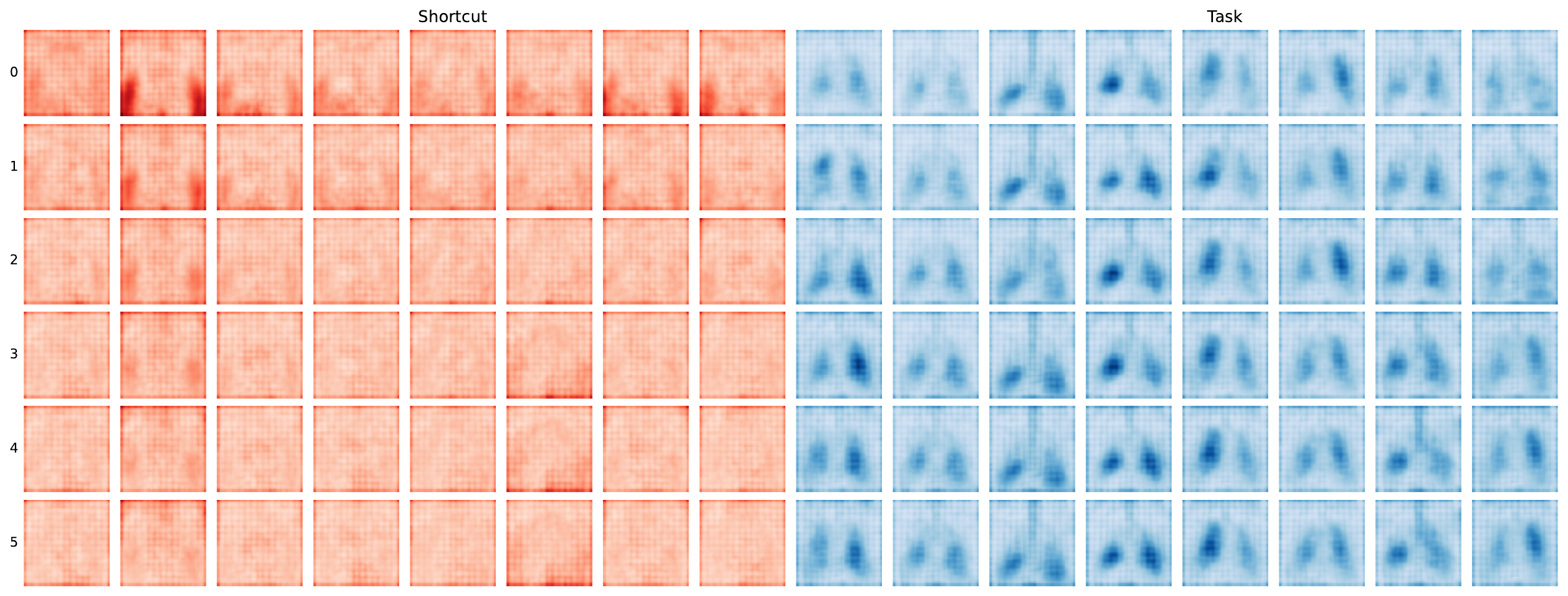}
\caption{\textbf{Shortcut contributions are suppressed from the second iteration onwards and the task contributions are comparatively stable.} Iterative contribution maps under one initial combined intervention followed by only shortcut suppression on \textsc{CheXpert} using ViT.}
\label{fig:chexpert_iter_suppression}
\end{figure}

\Cref{fig:celeba_iter_residual_suppression,fig:celeba_iter_suppression,fig:chexpert_iter_residual_suppression,fig:chexpert_iter_suppression} show the shortcut group-level contribution maps (prototypes) for a) image-level shortcut suppression, b) task amplification c) combined shortcut suppression and task amplification and d) initial combined intervention followed by shortcut suppression, respectively, on \textsc{CheXpert} over five iterative interventions. In these figures, the rows correspond to intervention iterations, with iteration $0$ denoting the non-intervened model. Columns correspond to the $K=8$ NMF shortcut groups. Orange coloured maps show shortcut contribution and blue maps show task contribution after each iteration. These figures show that a) the shortcut suppression based contribution maps iteratively reduce the concentration in the shortcut regions while keeping the task regions similar over the five iterations, b) iterative task amplification does not affect the shortcut regions and deteriorates or changes the task regions heavily, c) combined iterative shortcut suppression and task amplification  results in suppression of shortcut regions but at the same time, damages the task regions and, d) initial combined intervention followed by just shortcut suppression suppresses the shortcut while preserving the task regions. Therefore, we use d) as the main iterative intervention approach in our experiments.

\section{Annotation based contribution localisation}
\label{sec:localisation_checks}

In Section~\ref{sec:exp-patterns}, we visualise shortcut group-level contribution maps that highlight the shortcut and task regions. However, the analysis was still based on visual analysis on whether the task and shortcut regions are plausible. Here, we use segmentation masks to analyse which image regions are associated with shortcut and task contribution. For \textsc{Waterbirds}, we use the CUB~\citep{welinder2010caltech} bird masks to separate the bird foreground from the background. For \textsc{CheXpert}, we use \textsc{CheXmask} annotations~\citep{PhysioNet-chexmask-cxr-segmentation-data-1.0.0,gaggion2024chexmask,pollard2026physionet}\footnotemark[3] to identify lung regions from non-lung regions. For each shortcut group, we measure the overlap between the top-10\% shortcut or task contribution regions and the corresponding membership-weighted mask.

\footnotetext[3]{We use $987$ of the $1000$ test images for which \textsc{CheXmask} annotations are available.}

Across our test images, the bird foreground mask covers only $13.4\%$ of the image (we resize the masks to a $56\times56$ grid for this analysis). A uniform contribution map would therefore roughly overlap with $13.4\%$ of the bird region. Table~\ref{tab:waterbirds_localisation} shows that, a) on average and across shortcut groups, task--bird overlap is consistently higher than shortcut--bird overlap and b), across shortcut groups, the contributions vary but have a consistently lower shortcut--bird overlap than the dataset-level contribution map~\citep{achara2025localising}. These differences indicate that task contribution is concentrated on the bird, whereas shortcut contribution is concentrated in the background.

Figure~\ref{fig:chexmask} and Table~\ref{tab:chexpert_localisation} provide a dataset and shortcut group-level anatomical analysis for \textsc{CheXpert} based on the annotations from \textsc{CheXmask}. Figure~\ref{fig:chexmask} shows the top-10\% regions of  NMF ($K=8)$ membership-weighted contribution maps for all shortcut groups. It can be seen that the top task regions are mostly anatomical with contribution concentrated around left and right lung regions, and the top shortcut regions are mostly concentrated outside of the lungs. As the \textsc{CheXmask} annotations are averaged at the shortcut group level, there will not be exact alignment between the segmented anatomical regions and the actual image regions and therefore, the visual interpretation is approximate.

In the test image annotations, the lung region covers $24.1\%$ of the $56\times56$ grid (we resize the masks to $56\times56$), so a uniform contribution map would roughly overlap with $24.1\%$ of the lung region. Similar to the \textsc{Waterbirds} analysis, Table~\ref{tab:chexpert_localisation} shows that a) on average and across shortcut groups, task--lung overlap is consistently higher than shortcut--lung overlap and, b) across shortcut groups, the contributions vary but have a consistently lower shortcut--lung overlap than the dataset-level contribution map. 

Overall, this shows that contributions are heterogeneous, so shortcut group-level contribution maps are more informative than a single dataset-level map.

\input{Tables/annotated_localisation}

\section{Additional shortcut group contribution maps}
\label{sec:supp-contribution-map-figures}

\begin{figure}[p]
\centering
\includegraphics[width=\linewidth]{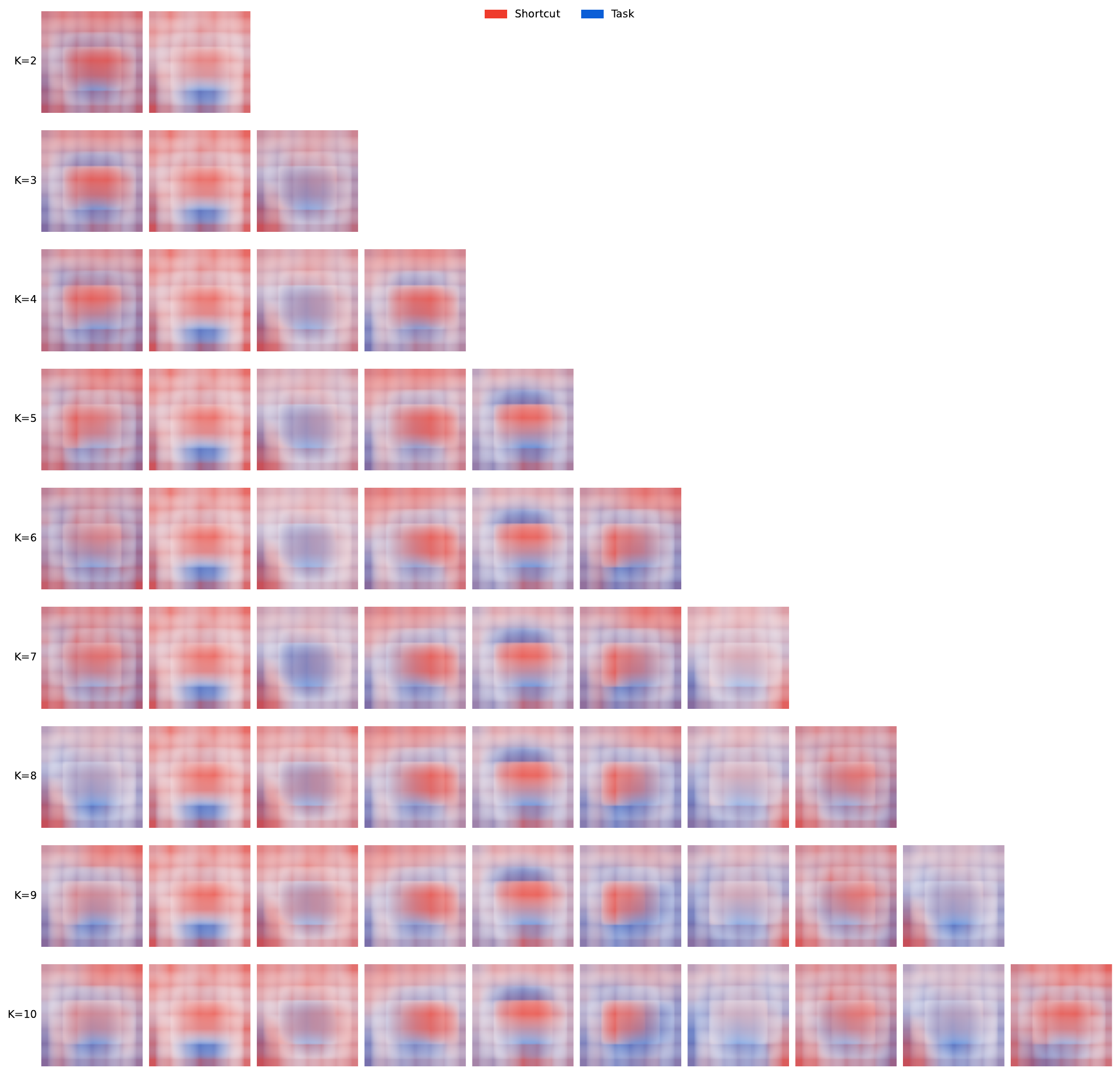}
\caption{\textbf{\textsc{CheXpert}, ResNet, Grad-CAM: NMF shortcut group contribution maps.}}
\label{fig:supp_resnet_chexpert_gradcam_nmf}
\end{figure}

\begin{figure}[p]
\centering
\includegraphics[width=\linewidth]{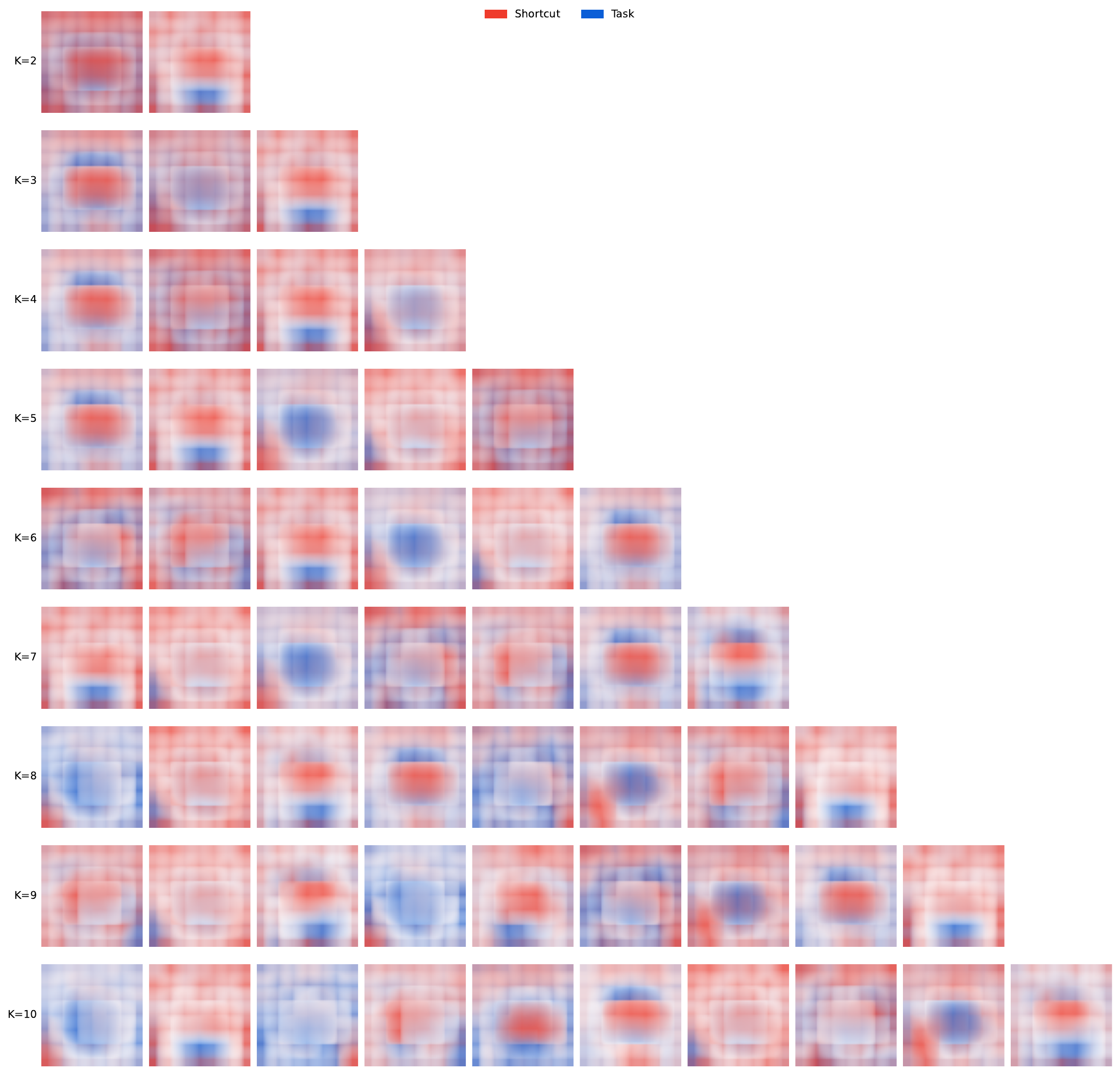}
\caption{\textbf{\textsc{CheXpert}, ResNet, Grad-CAM: K-means shortcut group contribution maps.}}
\label{fig:supp_resnet_chexpert_gradcam_kmeans}
\end{figure}

\begin{figure}[p]
\centering
\includegraphics[width=\linewidth]{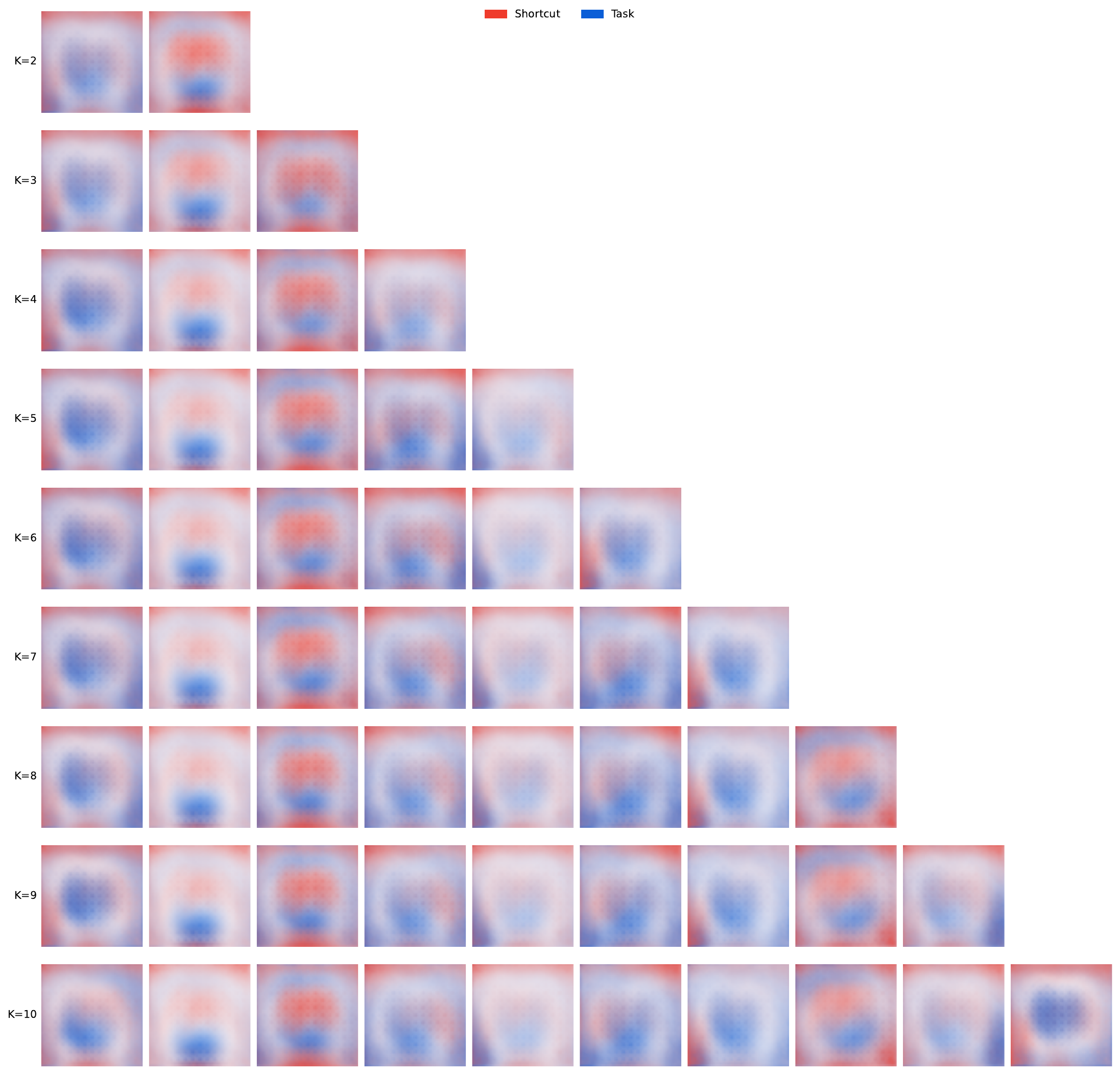}
\caption{\textbf{\textsc{CheXpert}, ResNet, LRP: NMF shortcut group contribution maps.}}
\label{fig:supp_resnet_chexpert_lrp_nmf}
\end{figure}

\begin{figure}[p]
\centering
\includegraphics[width=\linewidth]{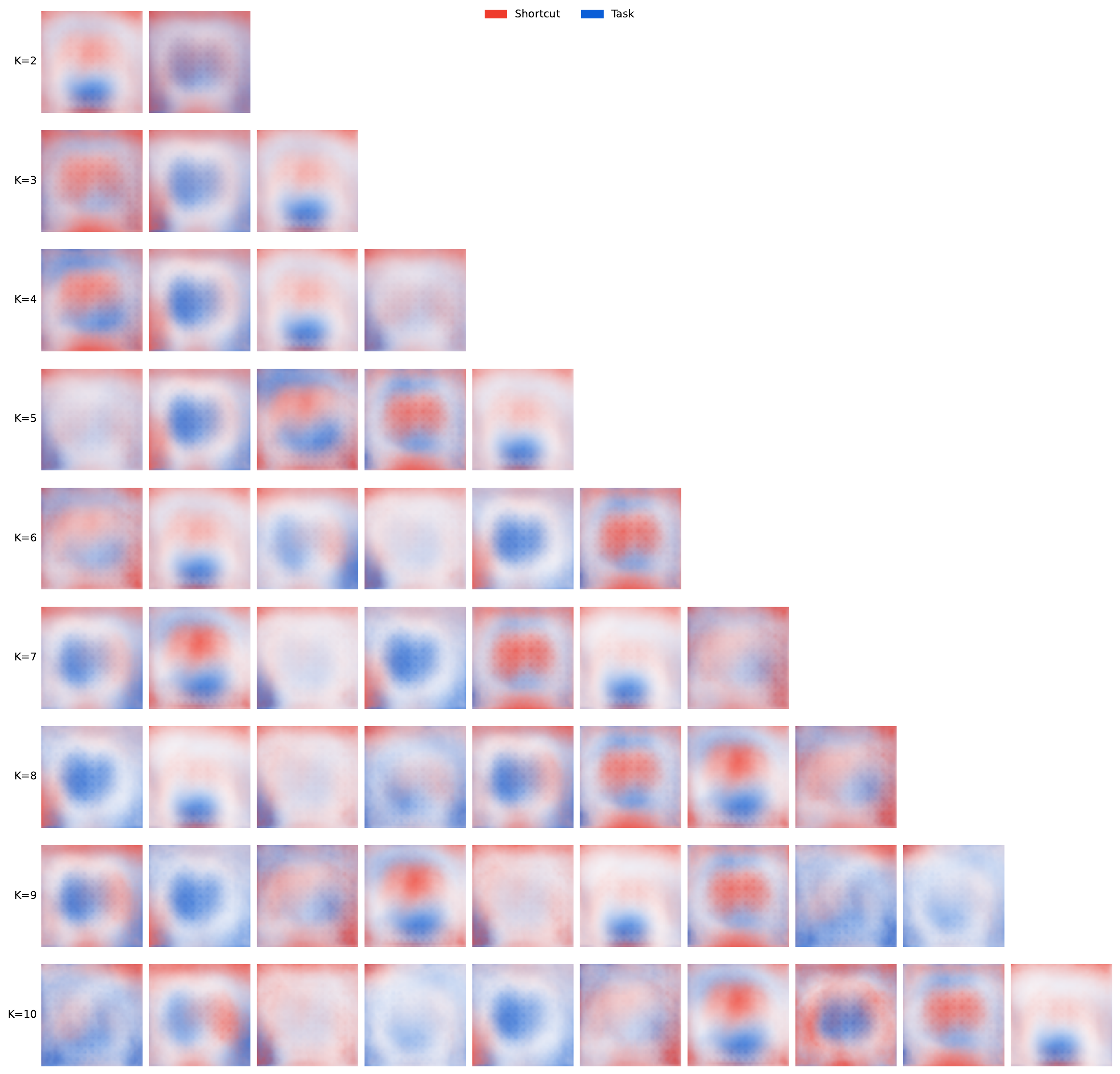}
\caption{\textbf{\textsc{CheXpert}, ResNet, LRP: K-means shortcut group contribution maps.}}
\label{fig:supp_resnet_chexpert_lrp_kmeans}
\end{figure}

\begin{figure}[p]
\centering
\includegraphics[width=\linewidth]{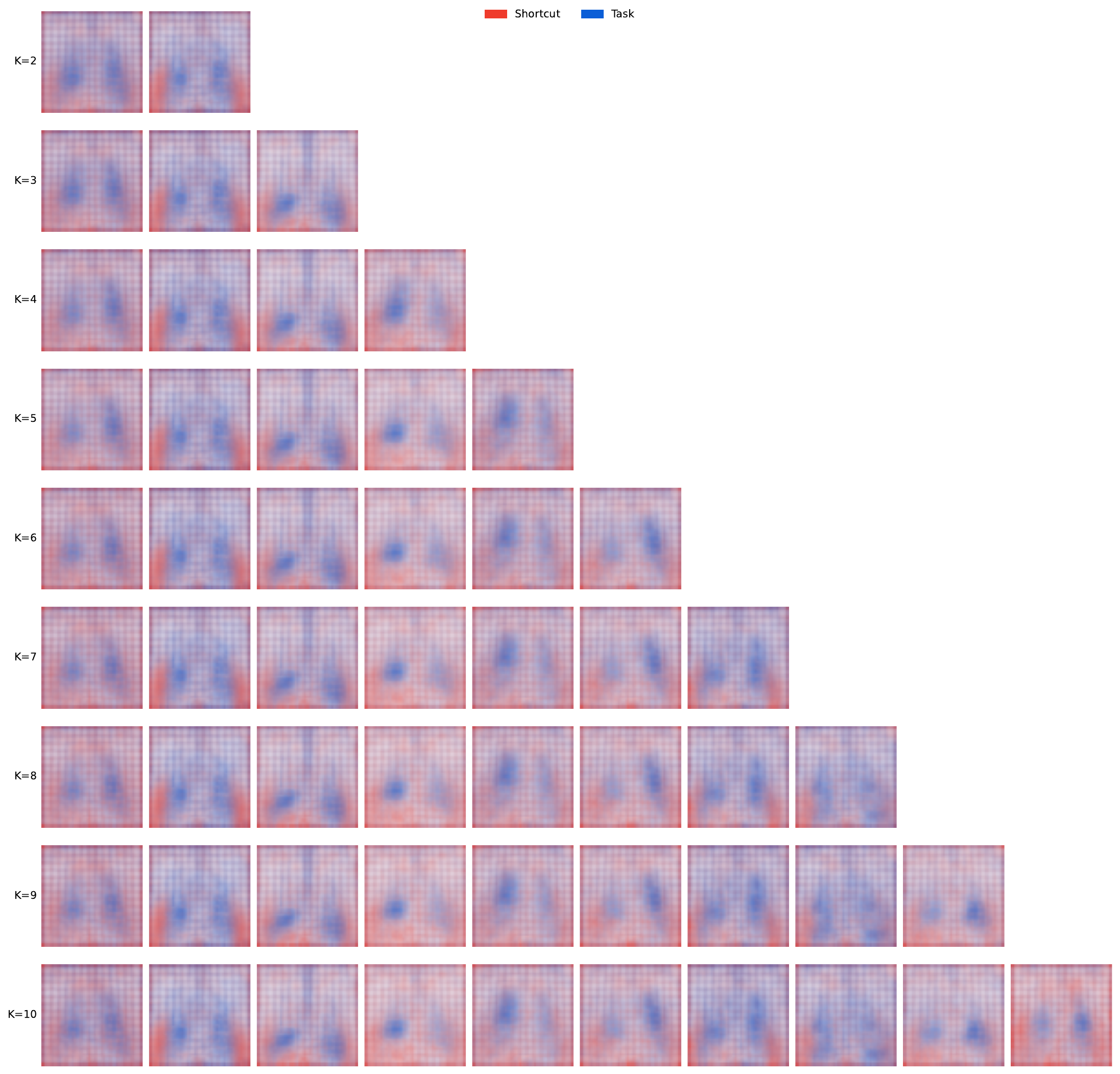}
\caption{\textbf{\textsc{CheXpert}, ViT, AttnLRP: NMF shortcut group contribution maps.}}
\label{fig:supp_vit_chexpert_lrp_nmf}
\end{figure}

\begin{figure}[p]
\centering
\includegraphics[width=\linewidth]{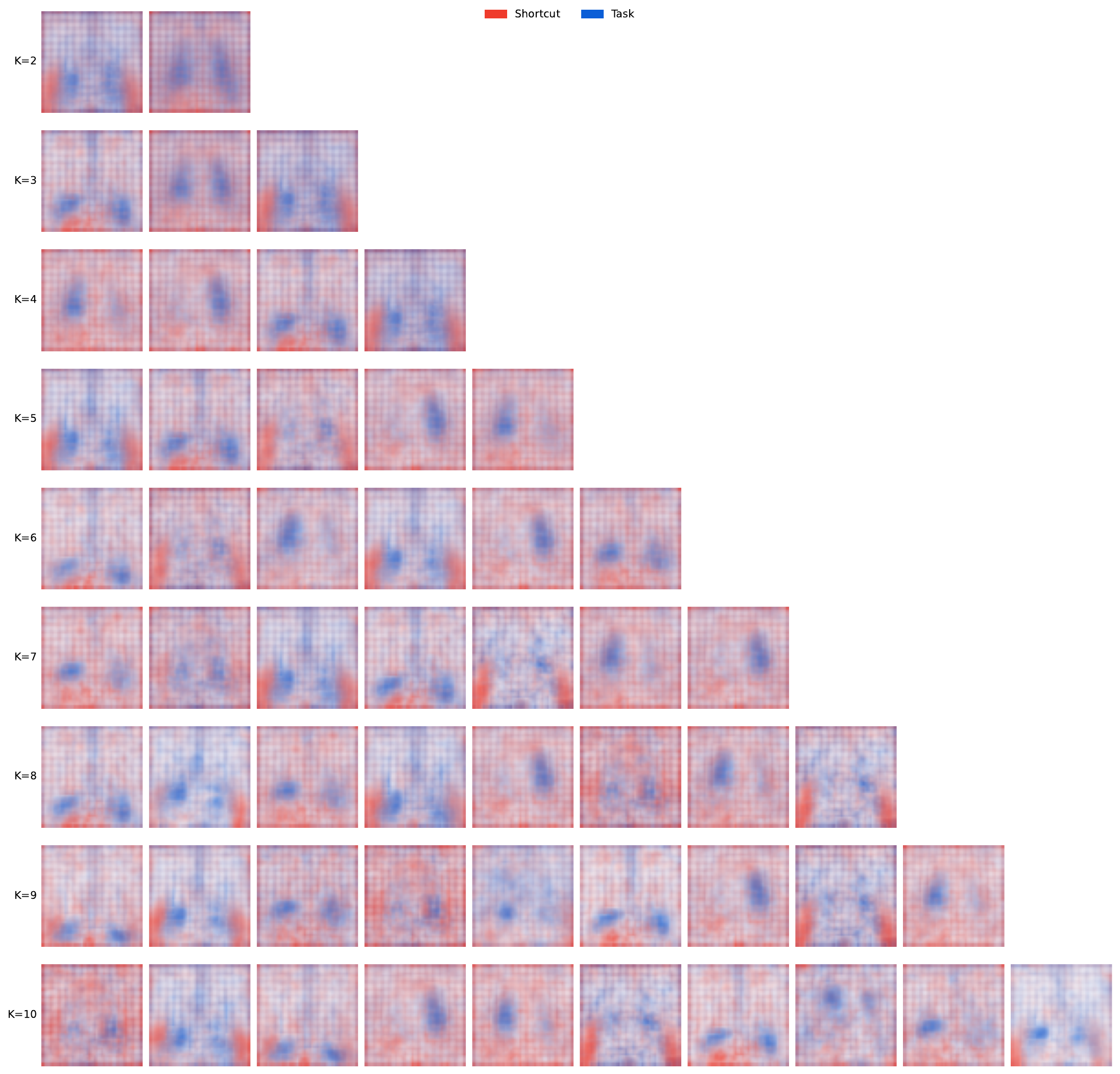}
\caption{\textbf{\textsc{CheXpert}, ViT, AttnLRP: K-means shortcut group contribution maps.}}
\label{fig:supp_vit_chexpert_lrp_kmeans}
\end{figure}

\begin{figure}[p]
\centering
\includegraphics[width=\linewidth]{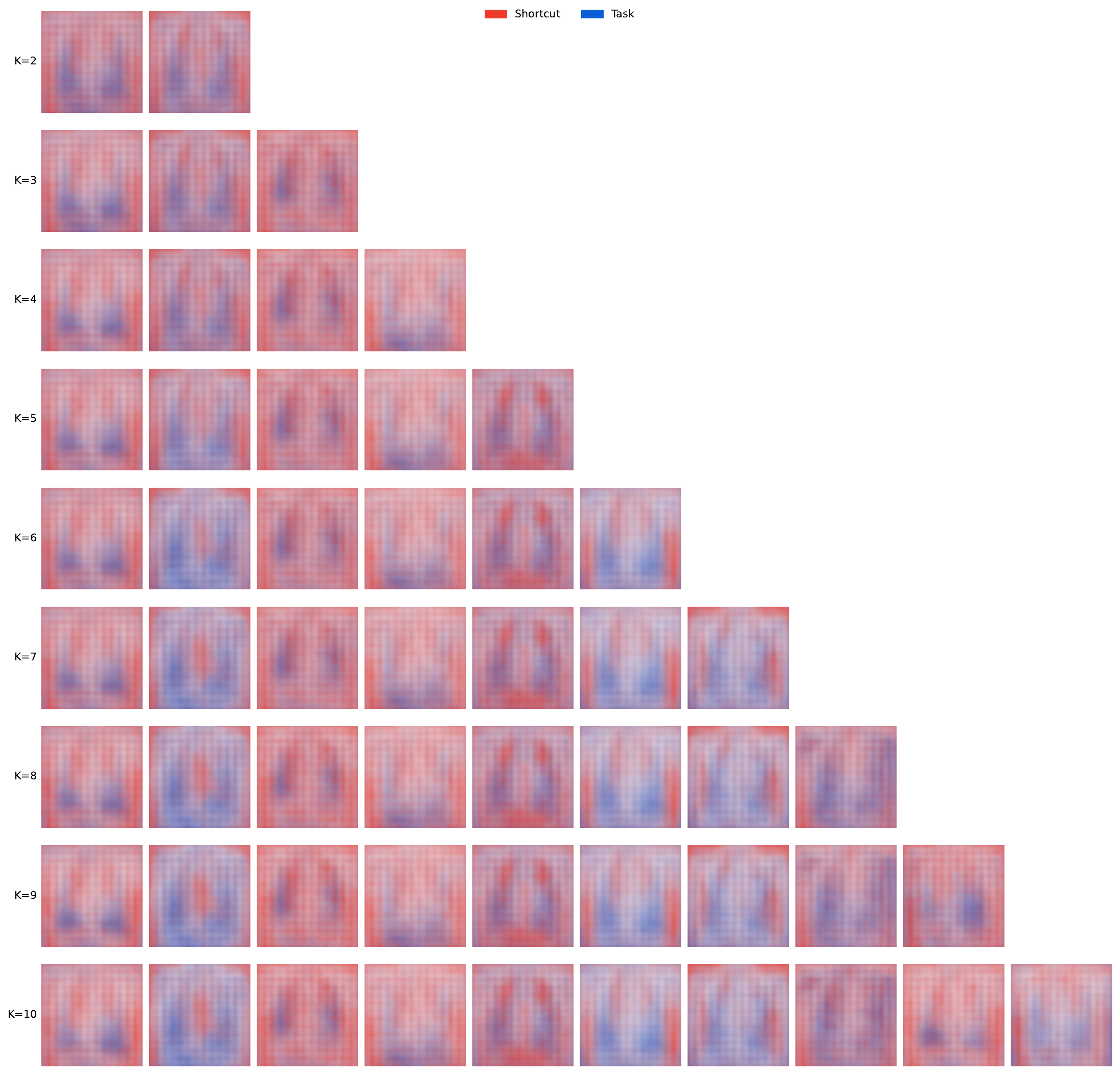}
\caption{\textbf{\textsc{CheXpert}, ViT, TiS: NMF shortcut group contribution maps.}}
\label{fig:supp_vit_chexpert_tis_nmf}
\end{figure}

\begin{figure}[p]
\centering
\includegraphics[width=\linewidth]{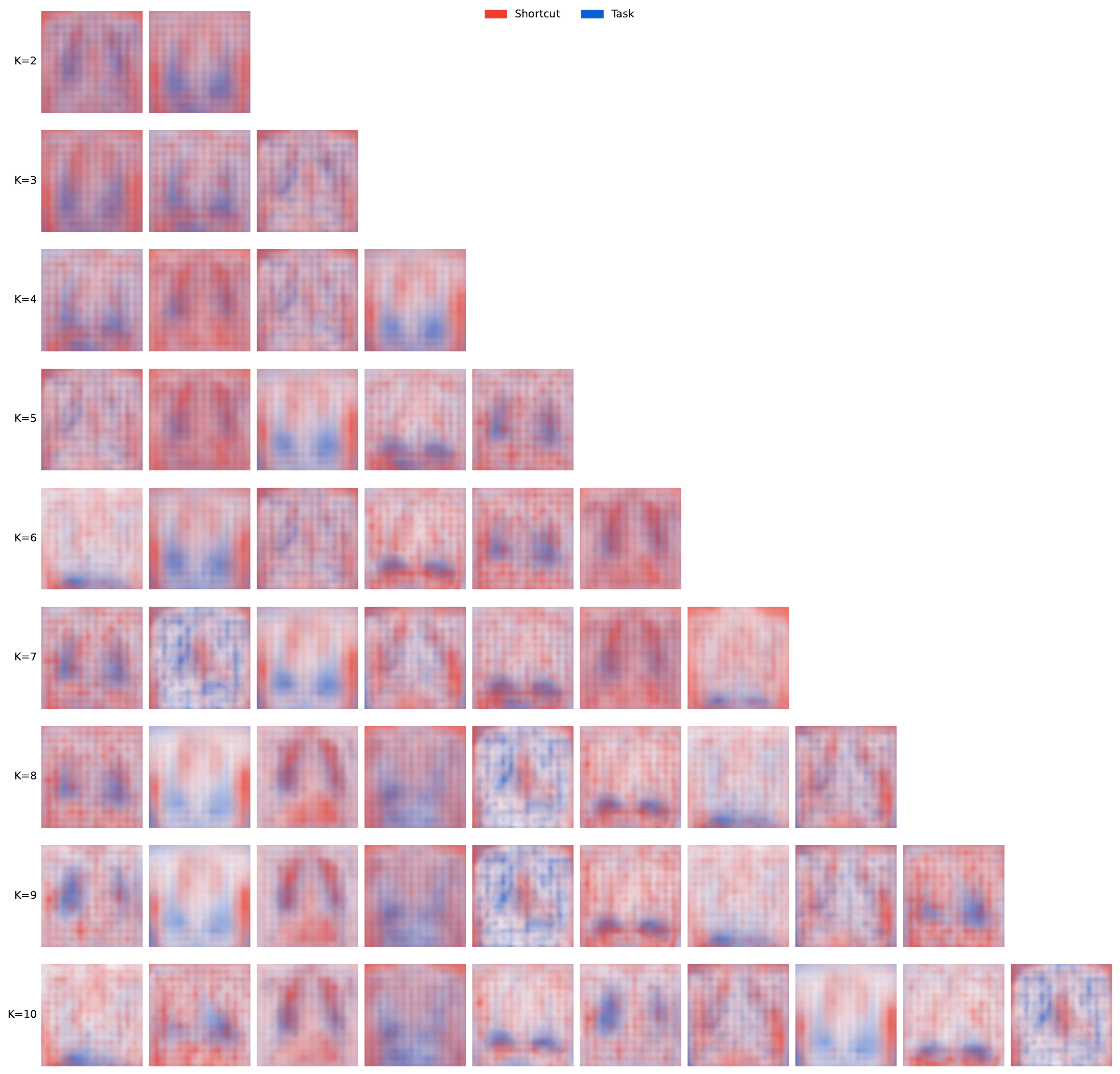}
\caption{\textbf{\textsc{CheXpert}, ViT, TiS: K-means shortcut group contribution maps.}}
\label{fig:supp_vit_chexpert_tis_kmeans}
\end{figure}

\begin{figure}[p]
\centering
\includegraphics[width=\linewidth]{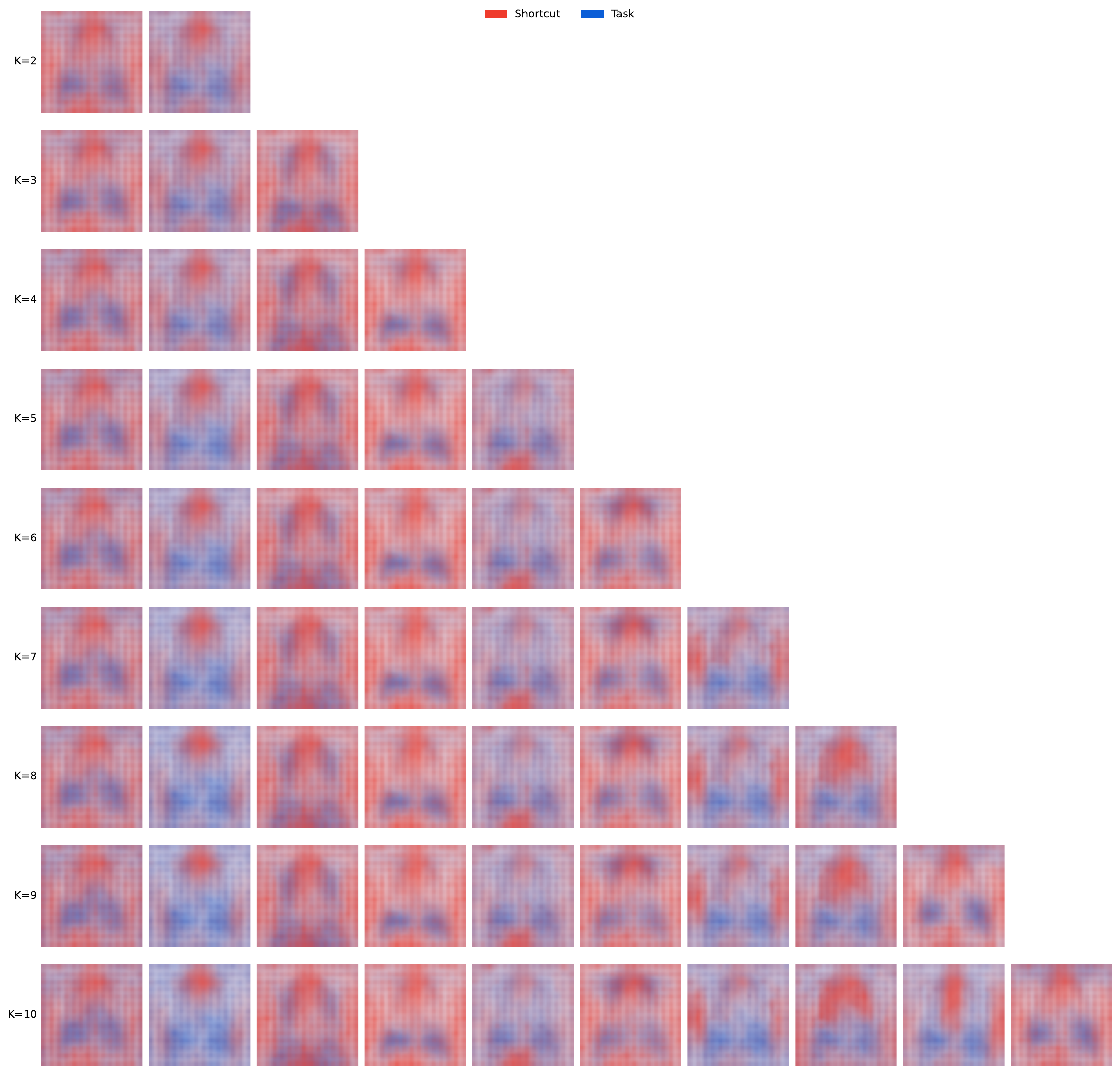}
\caption{\textbf{\textsc{CheXpert}, ViT, CX: NMF shortcut group contribution maps.}}
\label{fig:supp_vit_chexpert_cx_nmf}
\end{figure}

\begin{figure}[p]
\centering
\includegraphics[width=\linewidth]{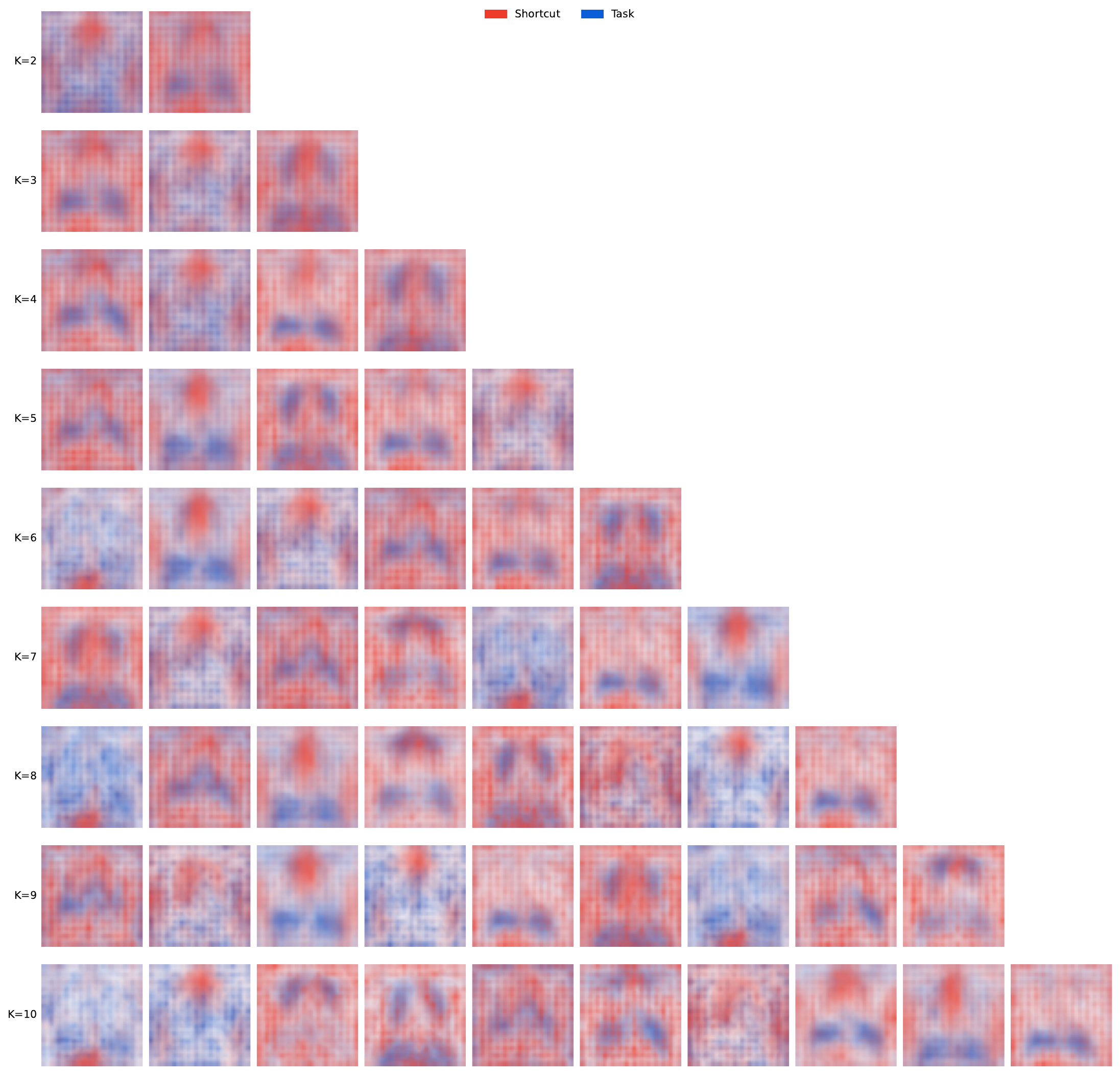}
\caption{\textbf{\textsc{CheXpert}, ViT, CX: K-means shortcut group contribution maps.}}
\label{fig:supp_vit_chexpert_cx_kmeans}
\end{figure}

\begin{figure}[p]
\centering
\includegraphics[width=\linewidth]{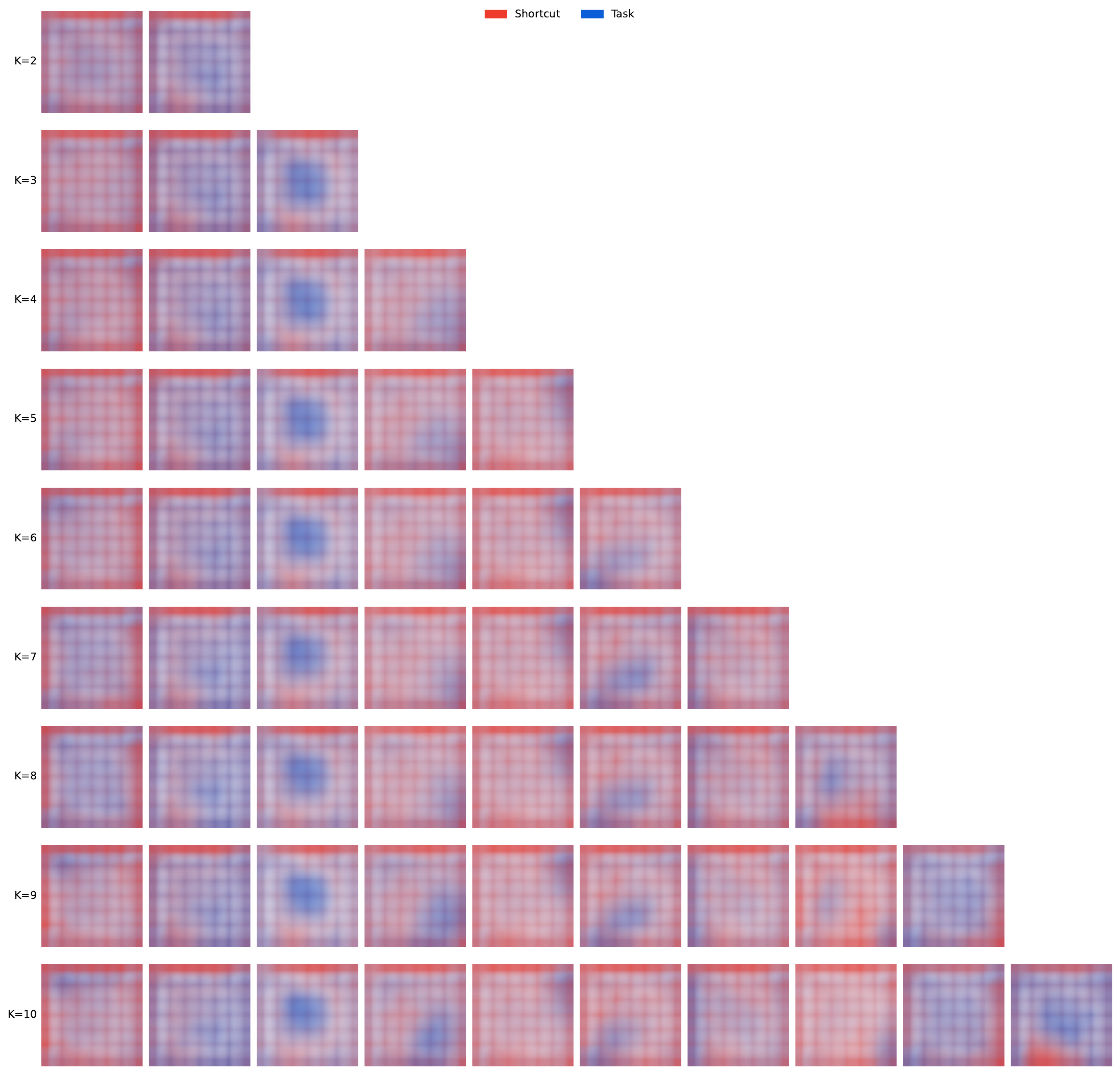}
\caption{\textbf{\textsc{Camelyon17}, ResNet, Grad-CAM: NMF shortcut group contribution maps.}}
\label{fig:supp_resnet_camelyon_gradcam_nmf}
\end{figure}

\begin{figure}[p]
\centering
\includegraphics[width=\linewidth]{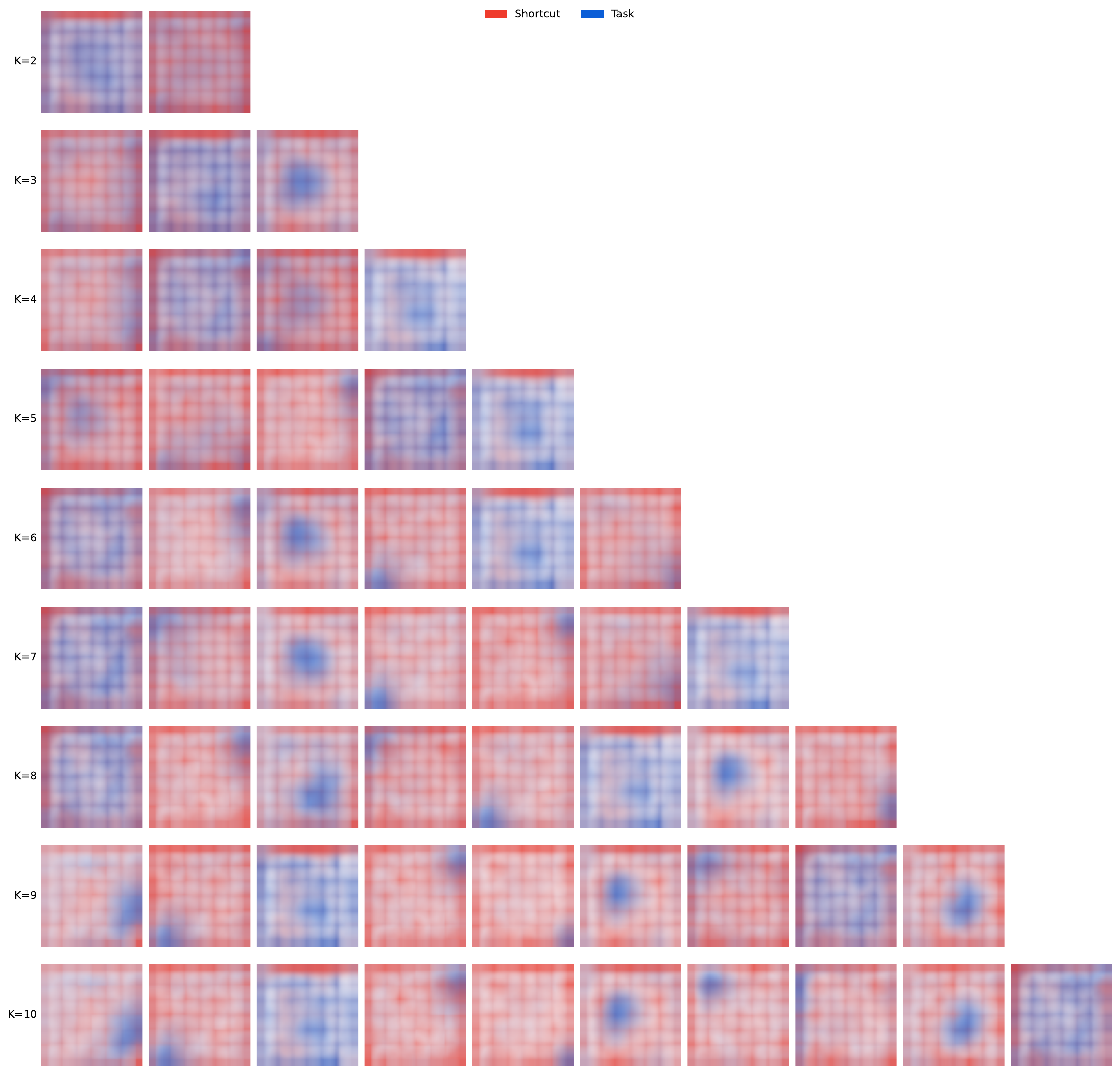}
\caption{\textbf{\textsc{Camelyon17}, ResNet, Grad-CAM: K-means shortcut group contribution maps.}}
\label{fig:supp_resnet_camelyon_gradcam_kmeans}
\end{figure}

\begin{figure}[p]
\centering
\includegraphics[width=\linewidth]{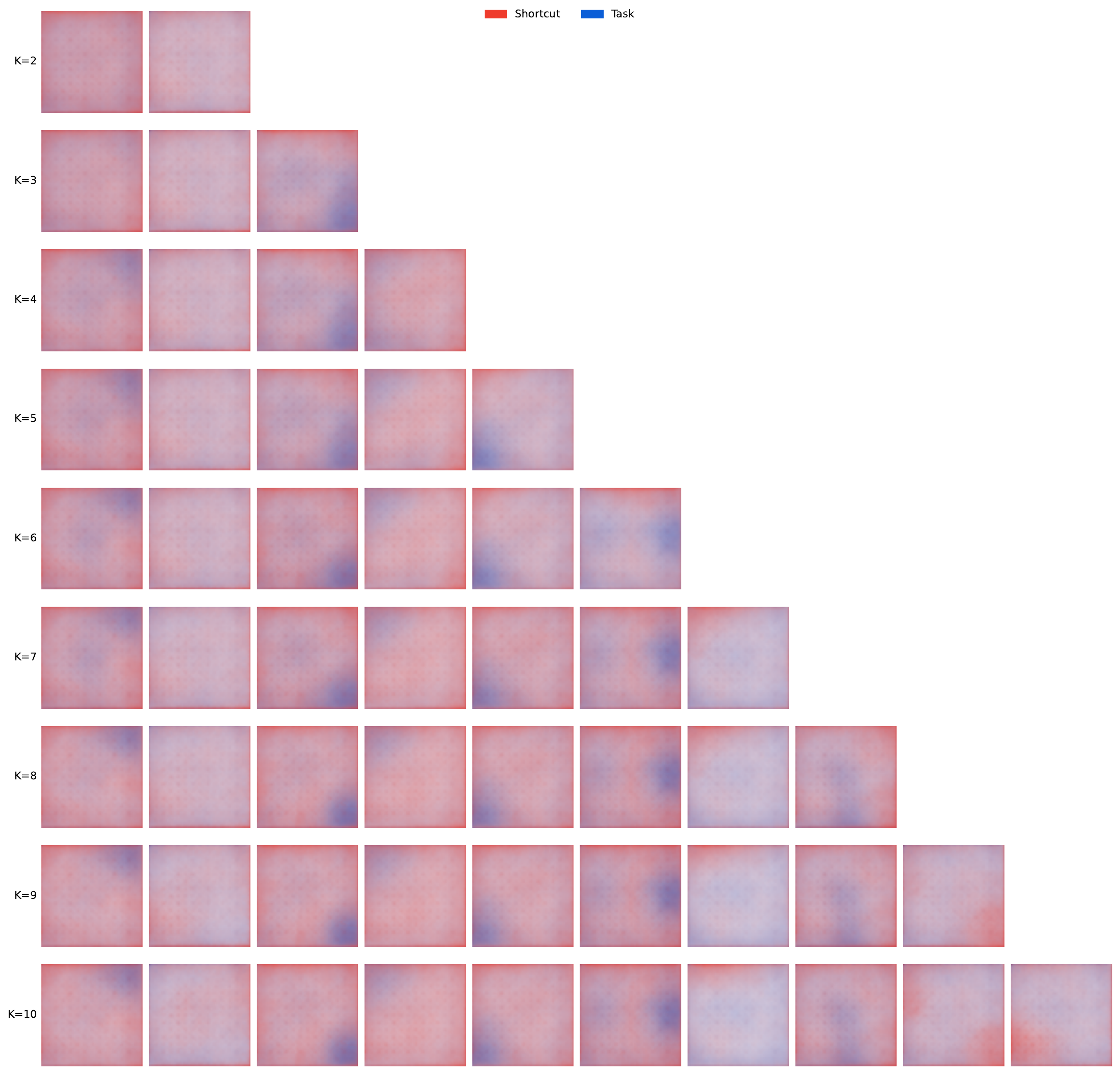}
\caption{\textbf{\textsc{Camelyon17}, ResNet, LRP: membership-weighted NMF shortcut group contribution maps.}}
\label{fig:supp_resnet_camelyon_lrp_nmf}
\end{figure}

\begin{figure}[p]
\centering
\includegraphics[width=\linewidth]{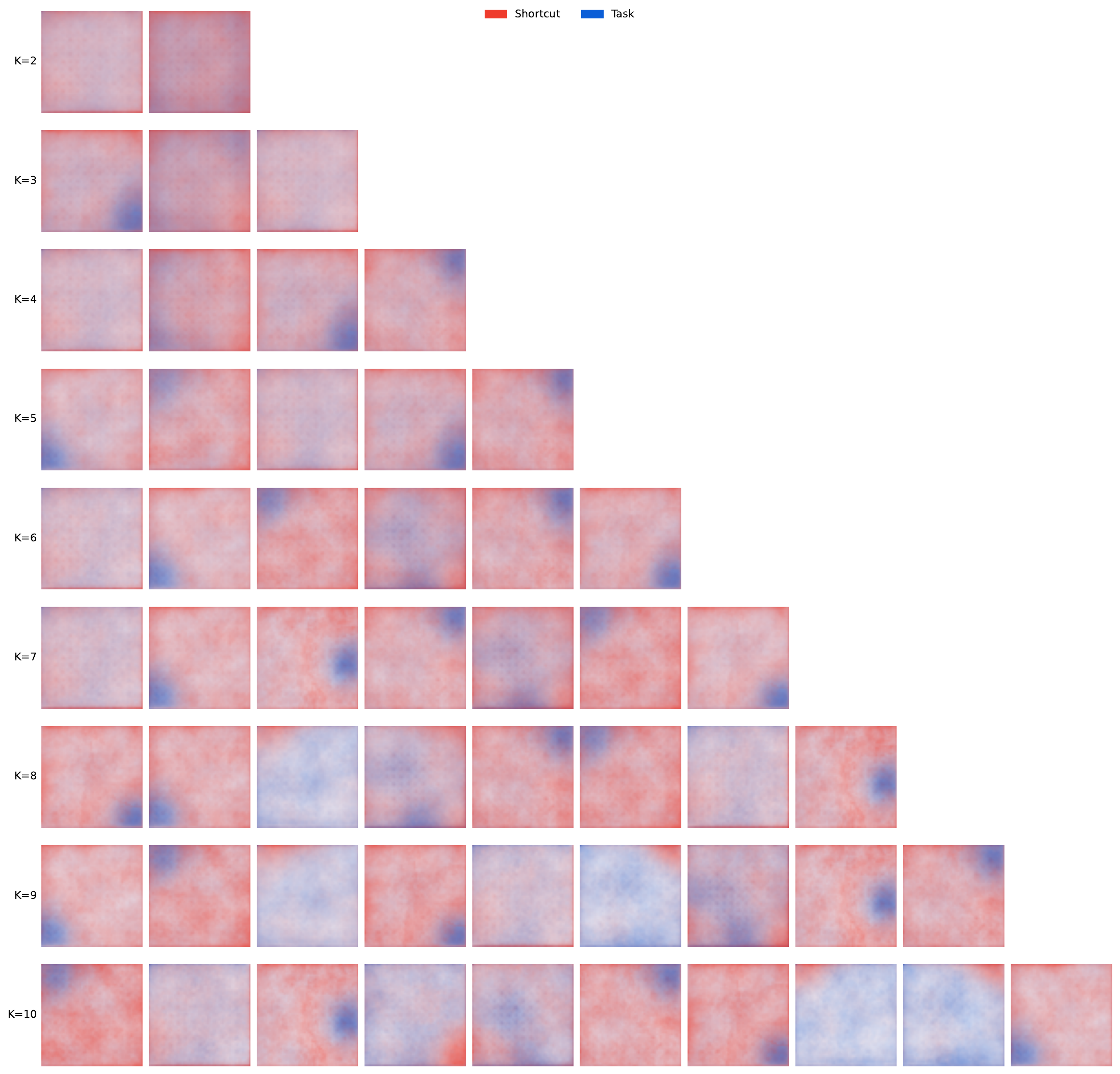}
\caption{\textbf{\textsc{Camelyon17}, ResNet, LRP: K-means shortcut group contribution maps.}}
\label{fig:supp_resnet_camelyon_lrp_kmeans}
\end{figure}

\begin{figure}[p]
\centering
\includegraphics[width=\linewidth]{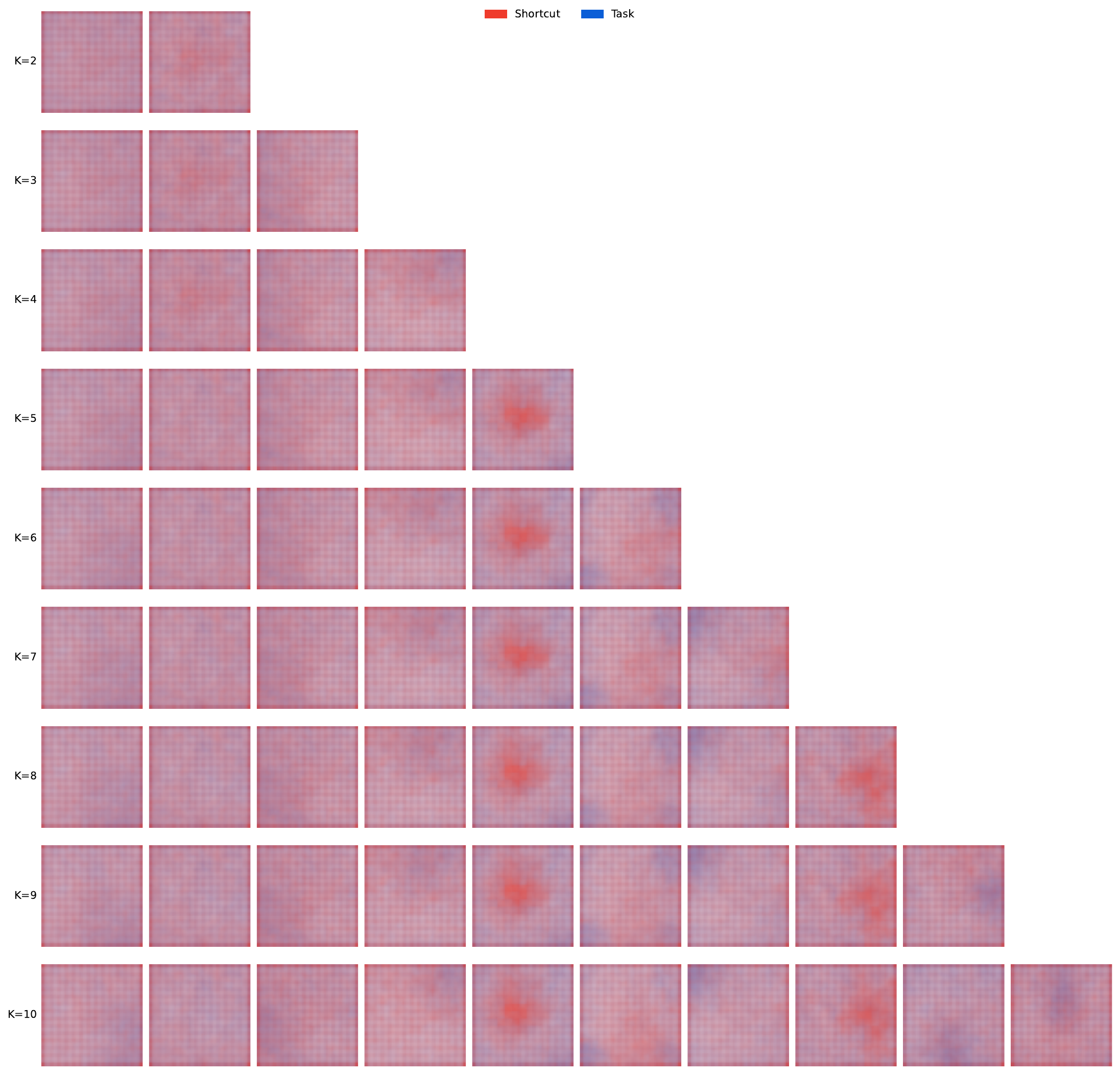}
\caption{\textbf{\textsc{Camelyon17}, ViT, AttnLRP: NMF shortcut group contribution maps.}}
\label{fig:supp_vit_camelyon_lrp_nmf}
\end{figure}

\begin{figure}[p]
\centering
\includegraphics[width=\linewidth]{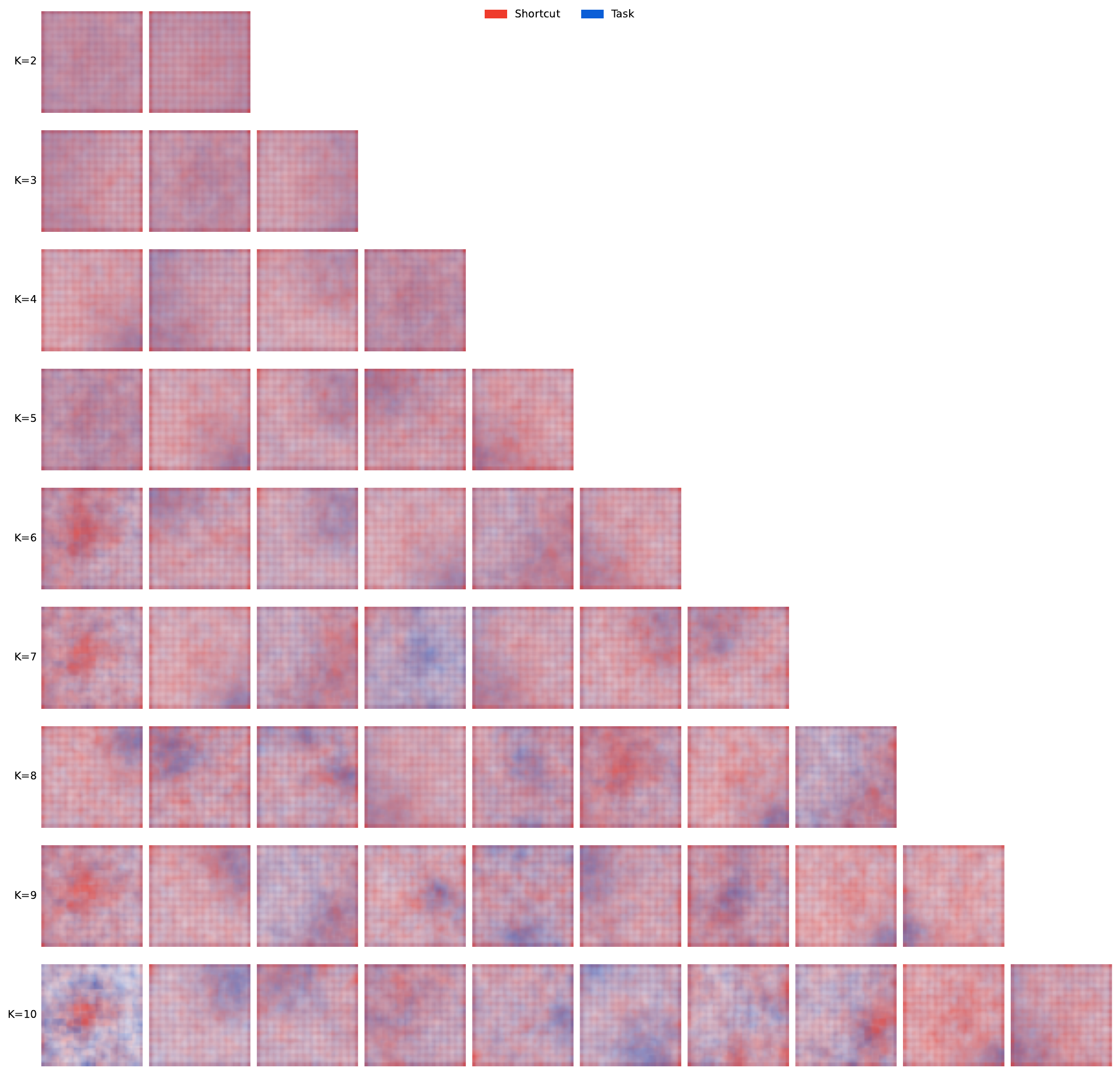}
\caption{\textbf{\textsc{Camelyon17}, ViT, AttnLRP: K-means shortcut group contribution maps.}}
\label{fig:supp_vit_camelyon_lrp_kmeans}
\end{figure}

\begin{figure}[p]
\centering
\includegraphics[width=\linewidth]{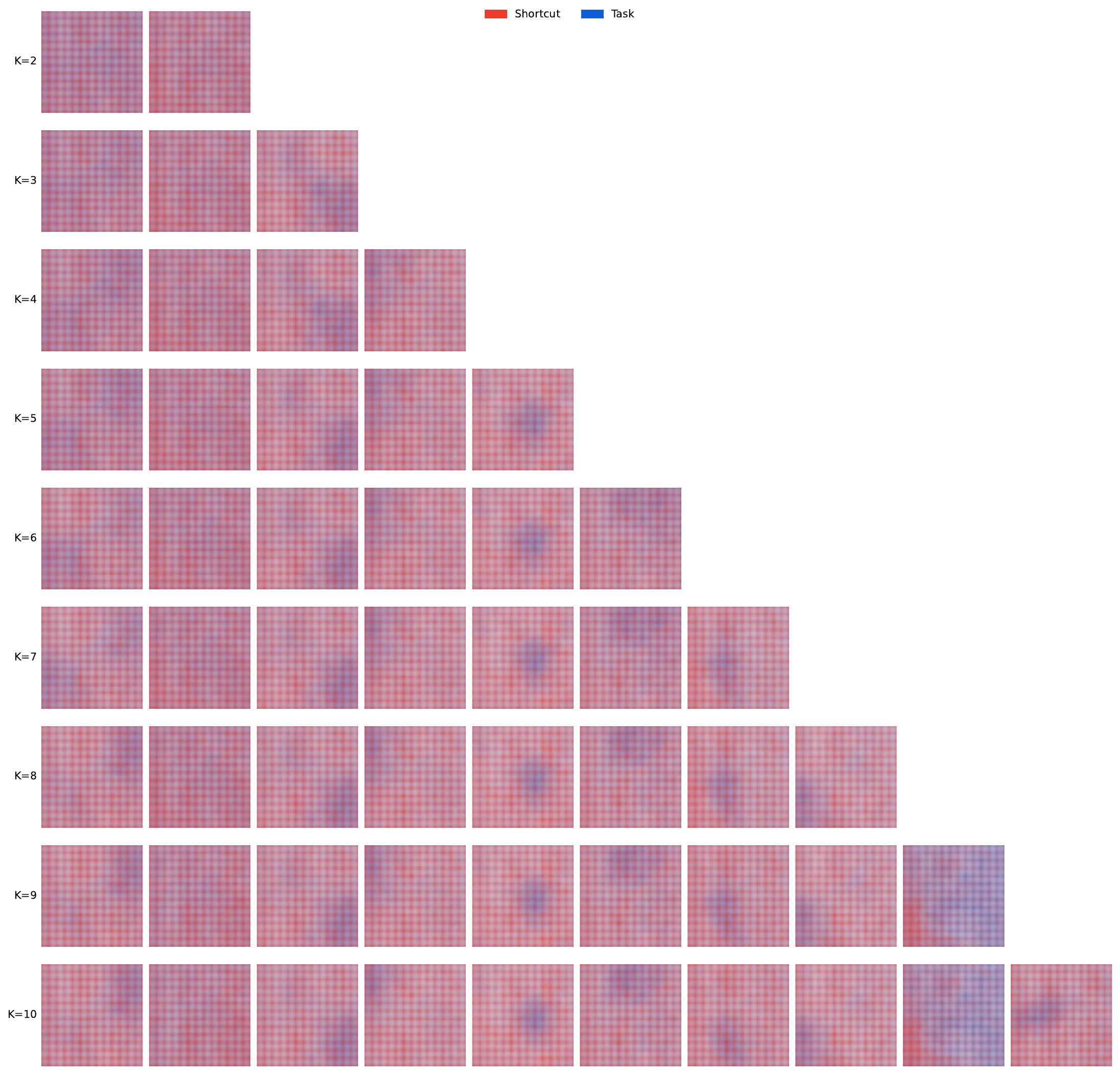}
\caption{\textbf{\textsc{Camelyon17}, ViT, TiS: NMF shortcut group contribution maps.}}
\label{fig:supp_vit_camelyon_tis_nmf}
\end{figure}

\begin{figure}[p]
\centering
\includegraphics[width=\linewidth]{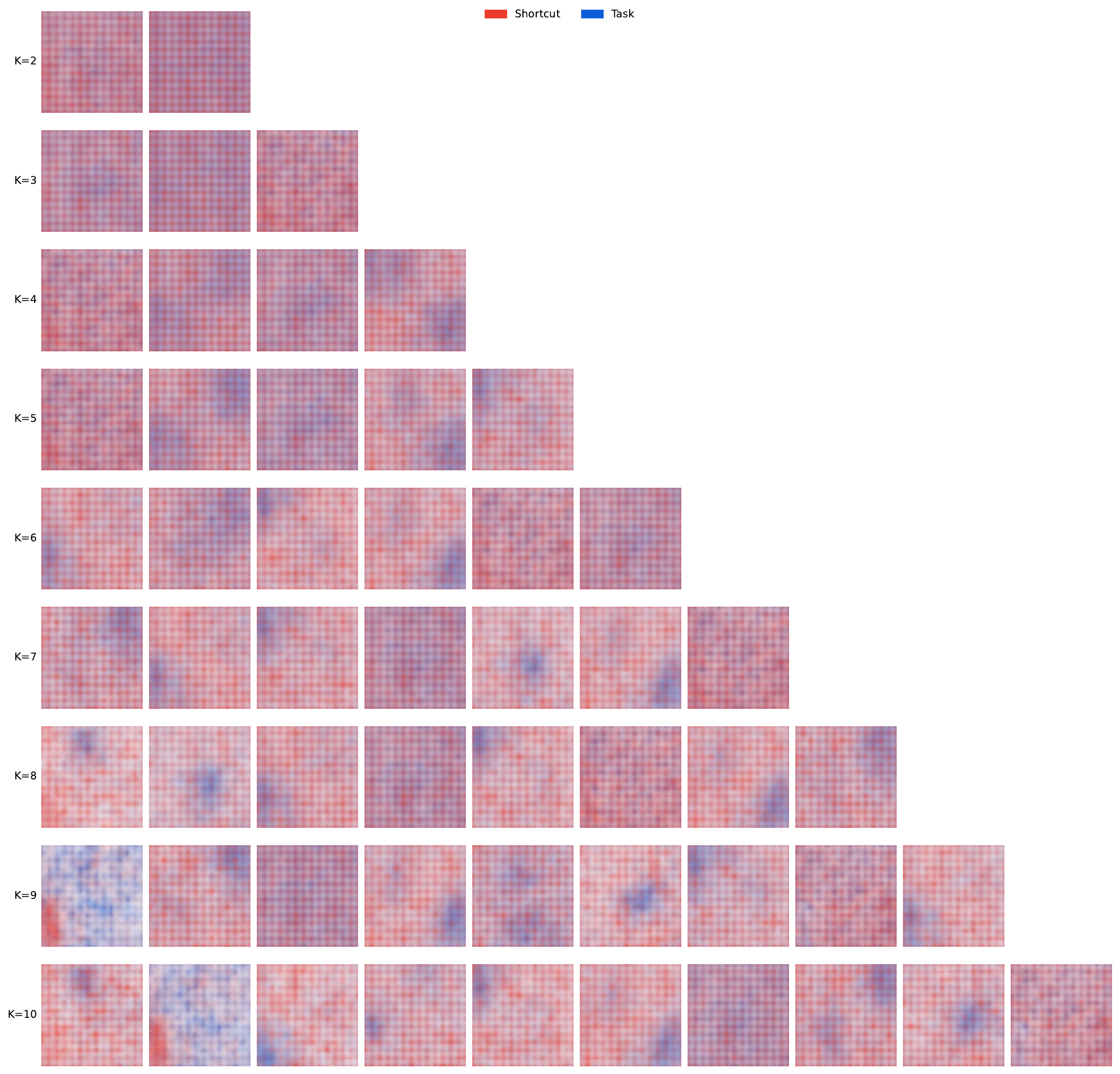}
\caption{\textbf{\textsc{Camelyon17}, ViT, TiS: K-means shortcut group contribution maps.}}
\label{fig:supp_vit_camelyon_tis_kmeans}
\end{figure}

\begin{figure}[p]
\centering
\includegraphics[width=\linewidth]{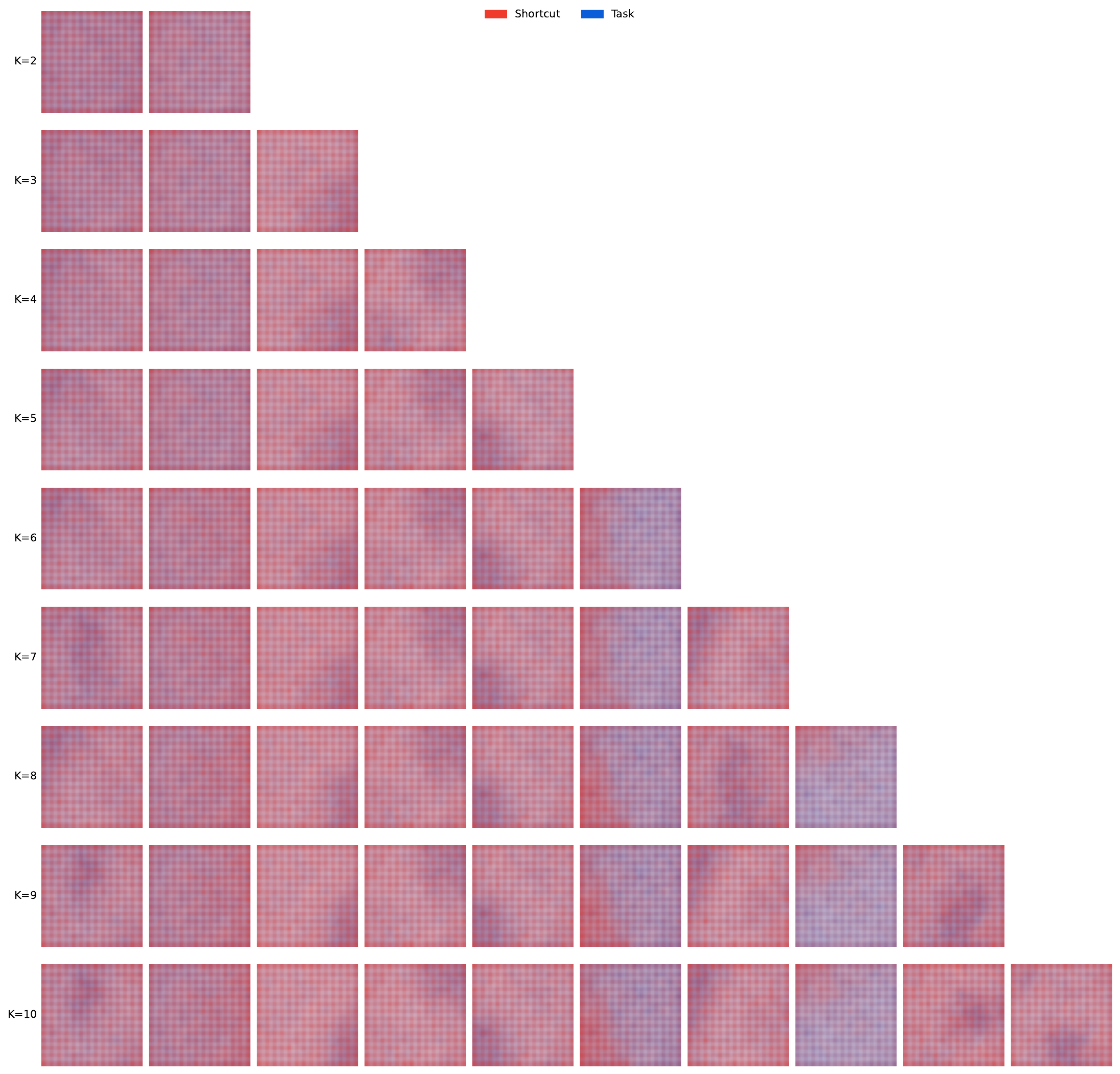}
\caption{\textbf{\textsc{Camelyon17}, ViT, CX: NMF shortcut group contribution maps.}}
\label{fig:supp_vit_camelyon_cx_nmf}
\end{figure}

\begin{figure}[p]
\centering
\includegraphics[width=\linewidth]{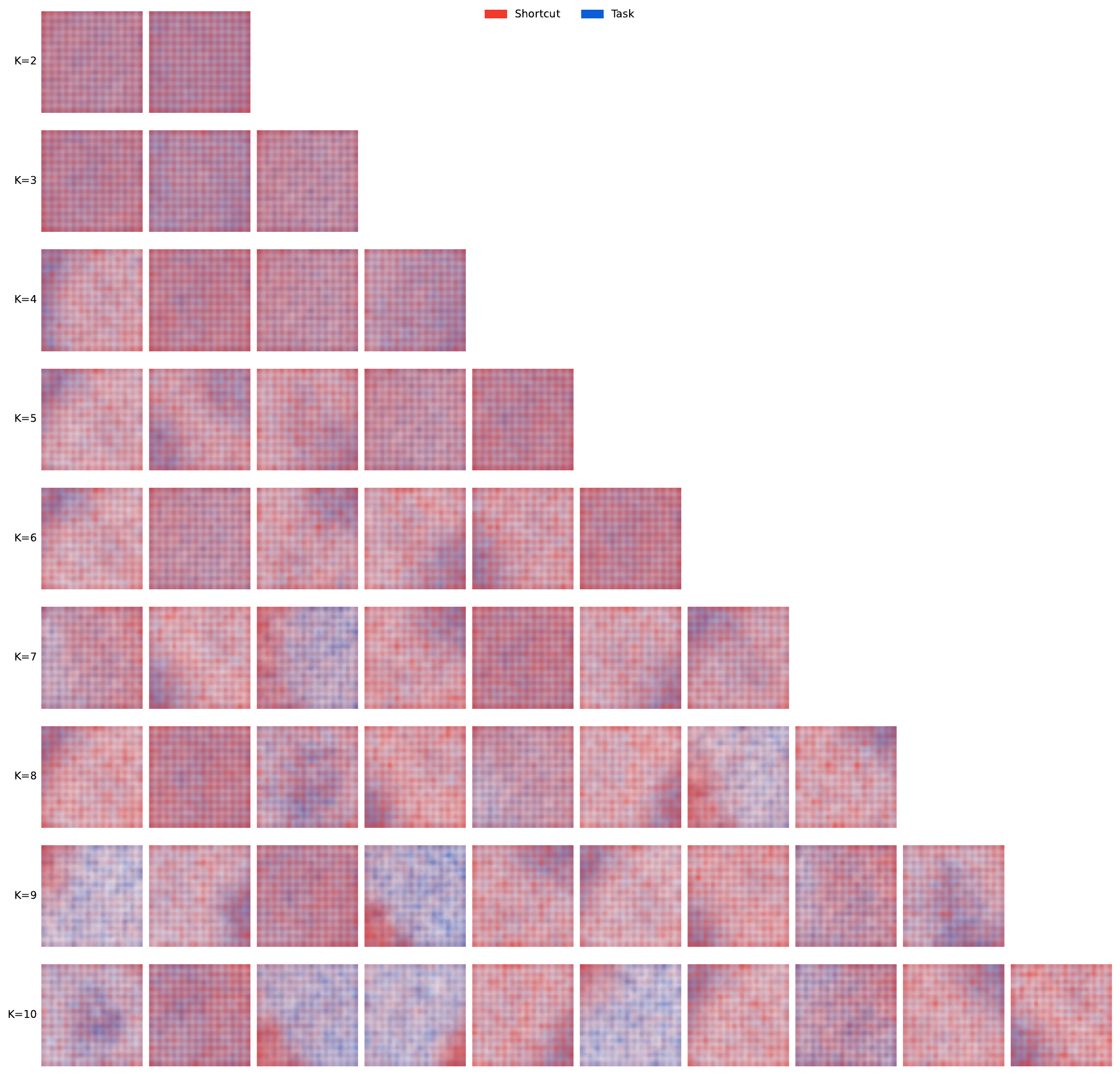}
\caption{\textbf{\textsc{Camelyon17}, ViT, CX: K-means shortcut group contribution maps.}}
\label{fig:supp_vit_camelyon_cx_kmeans}
\end{figure}

\begin{figure}[p]
\centering
\includegraphics[width=\linewidth]{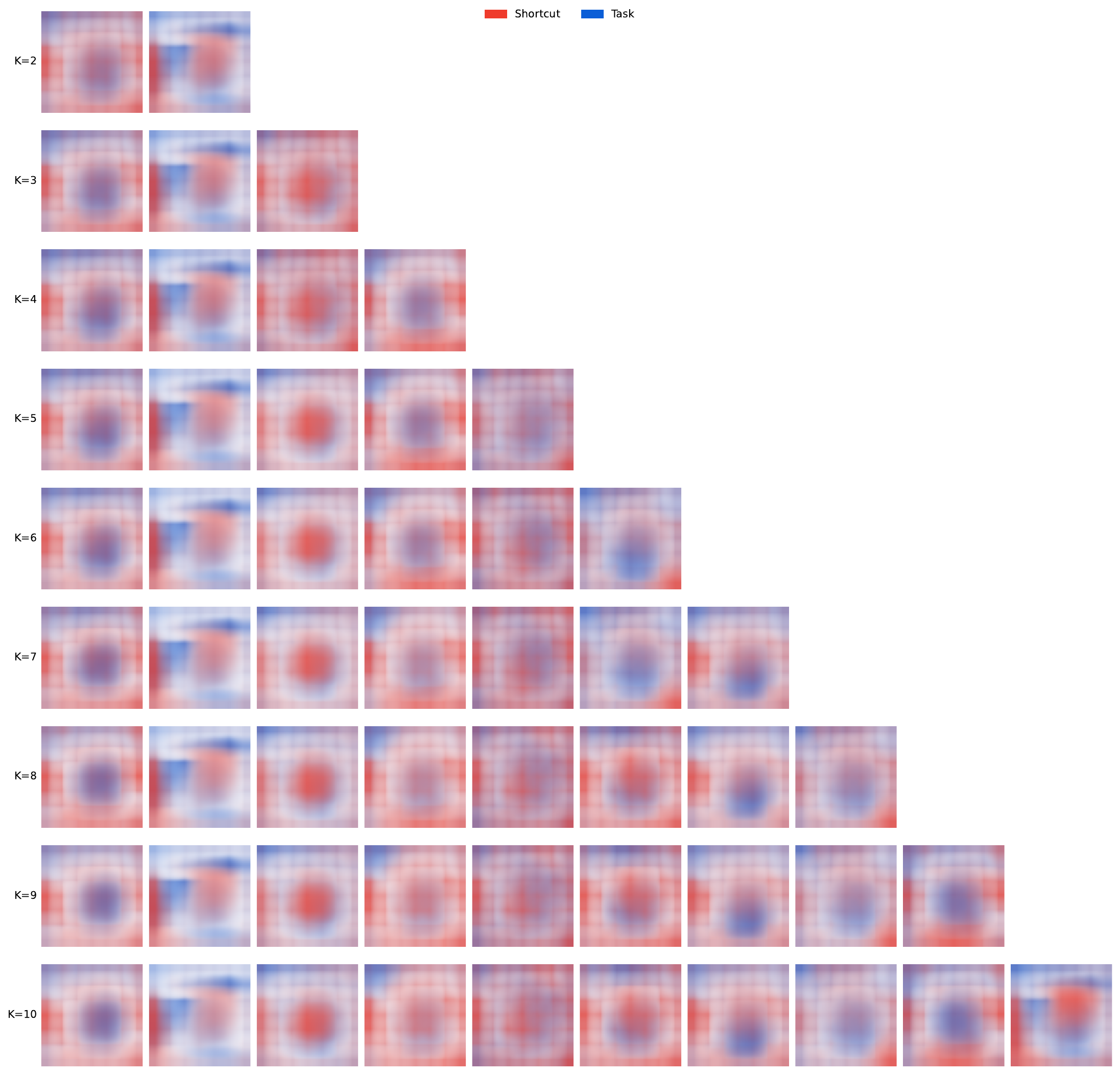}
\caption{\textbf{\textsc{ISIC2019}, ResNet, Grad-CAM: NMF shortcut group contribution maps.}}
\label{fig:supp_resnet_isic_gradcam_nmf}
\end{figure}

\begin{figure}[p]
\centering
\includegraphics[width=\linewidth]{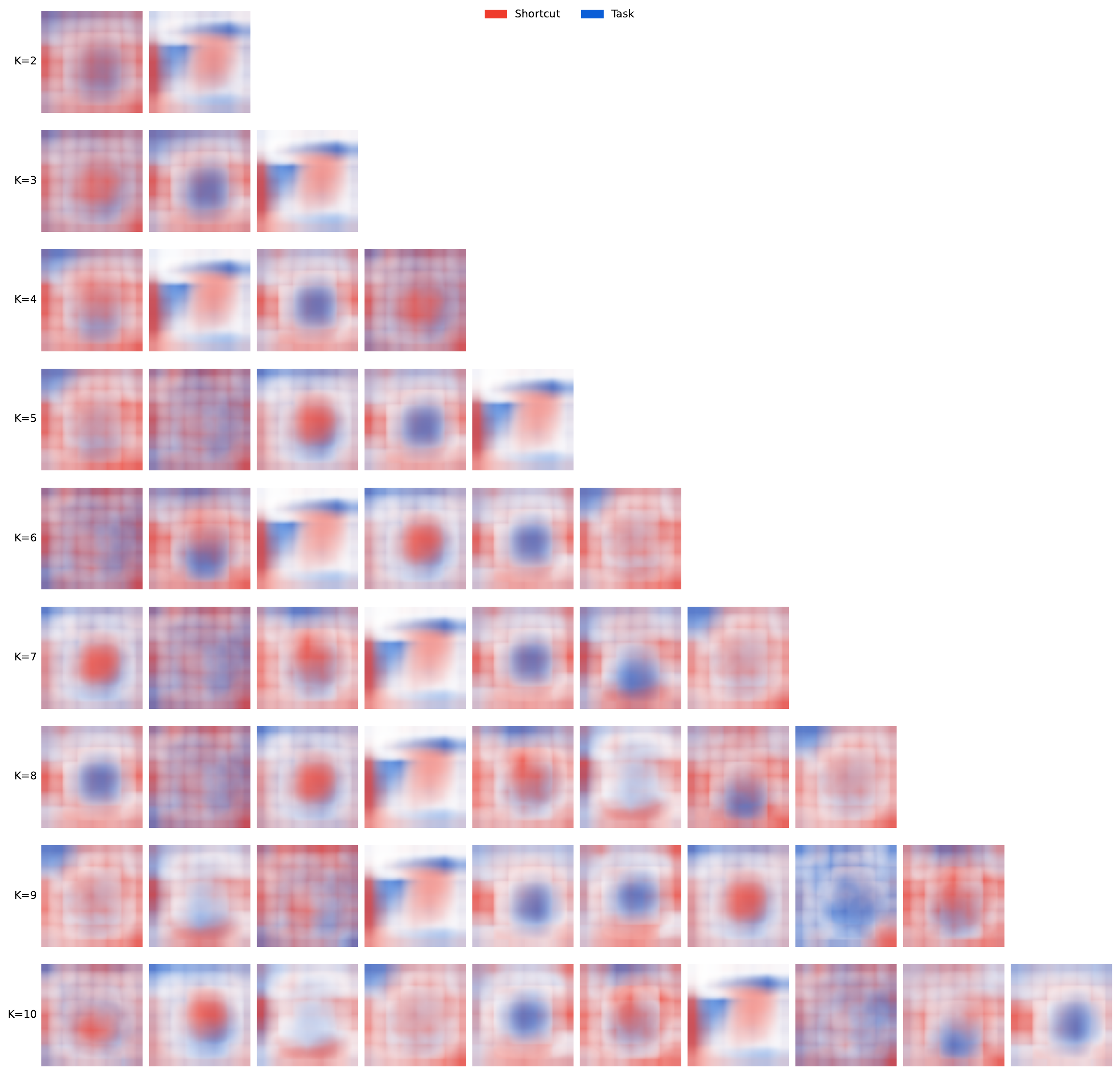}
\caption{\textbf{\textsc{ISIC2019}, ResNet, Grad-CAM: K-means shortcut group contribution maps.}}
\label{fig:supp_resnet_isic_gradcam_kmeans}
\end{figure}

\begin{figure}[p]
\centering
\includegraphics[width=\linewidth]{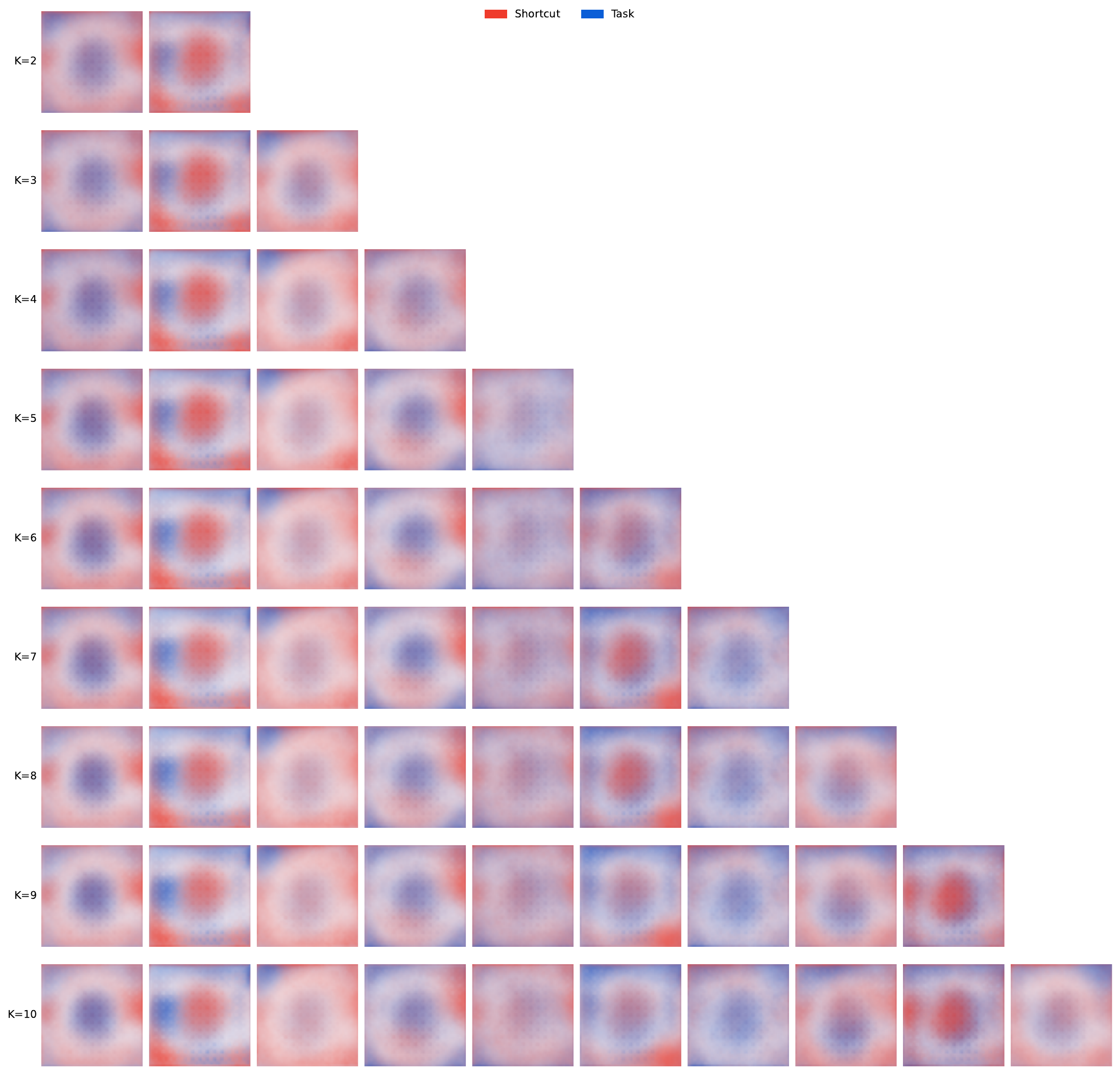}
\caption{\textbf{\textsc{ISIC2019}, ResNet, LRP: NMF shortcut group contribution maps.}}
\label{fig:supp_resnet_isic_lrp_nmf}
\end{figure}

\begin{figure}[p]
\centering
\includegraphics[width=\linewidth]{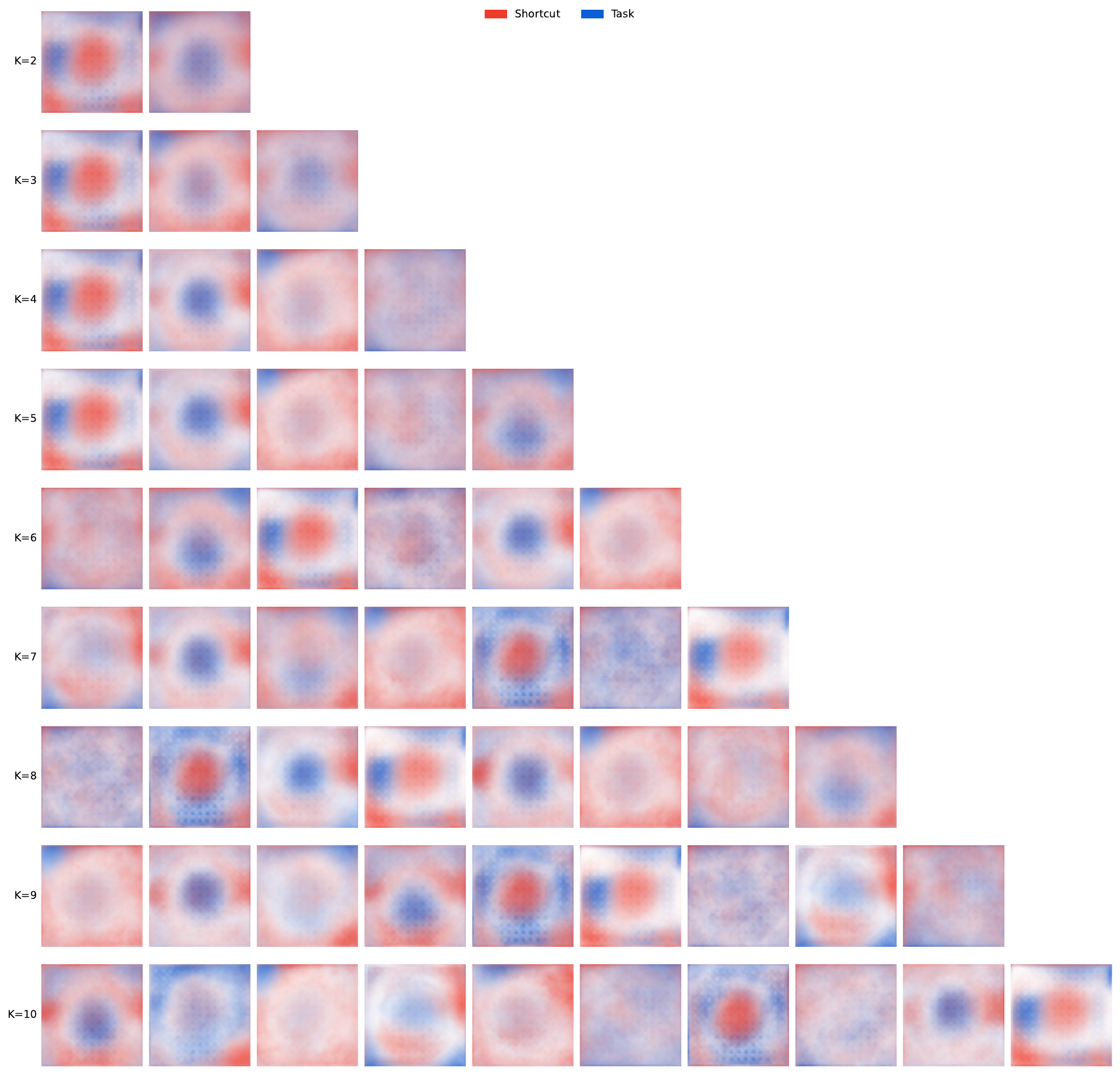}
\caption{\textbf{\textsc{ISIC2019}, ResNet, LRP: K-means shortcut group contribution maps.}}
\label{fig:supp_resnet_isic_lrp_kmeans}
\end{figure}

\begin{figure}[p]
\centering
\includegraphics[width=\linewidth]{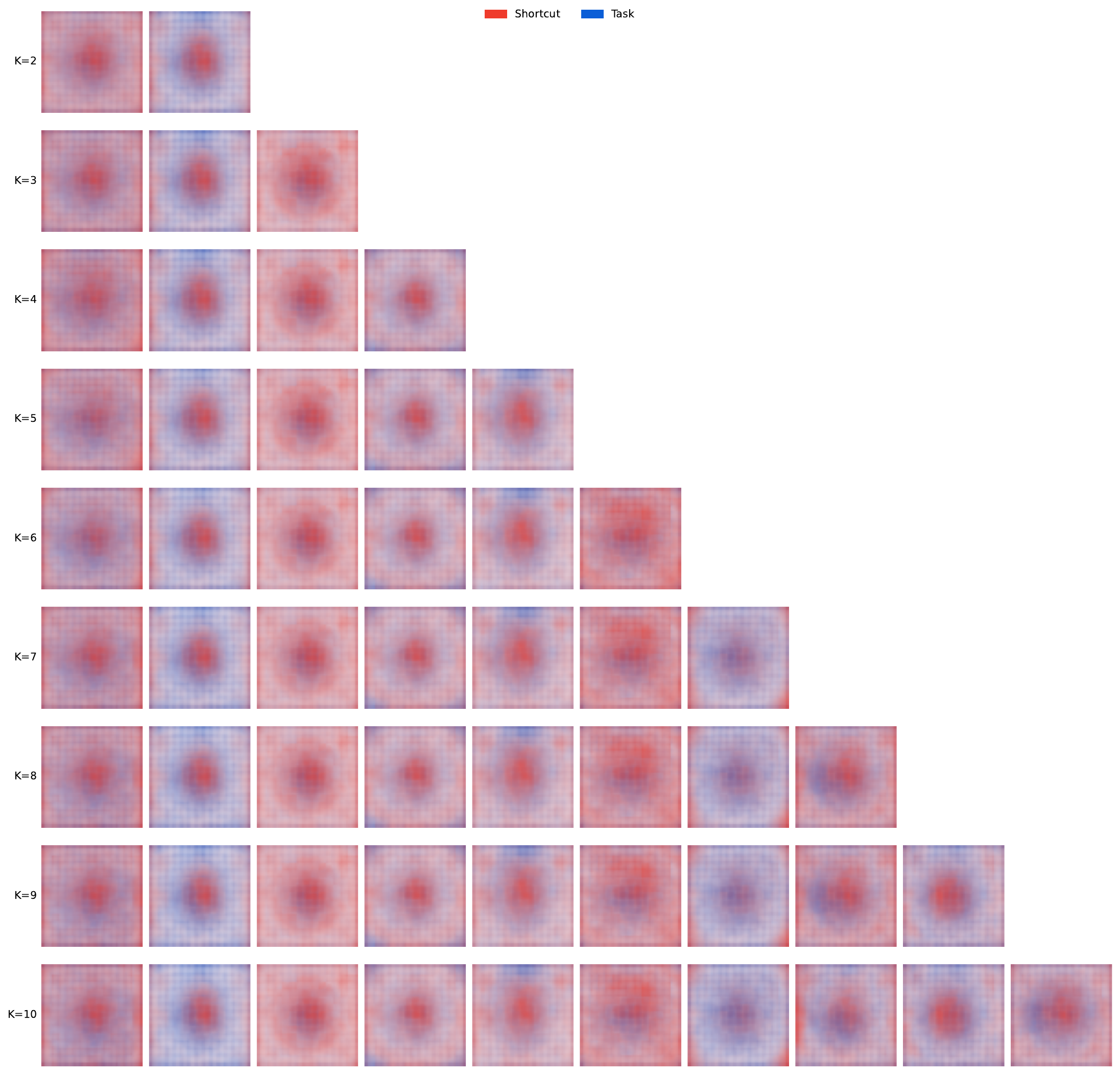}
\caption{\textbf{\textsc{ISIC2019}, ViT, AttnLRP: NMF shortcut group contribution maps.}}
\label{fig:supp_vit_isic_lrp_nmf}
\end{figure}

\begin{figure}[p]
\centering
\includegraphics[width=\linewidth]{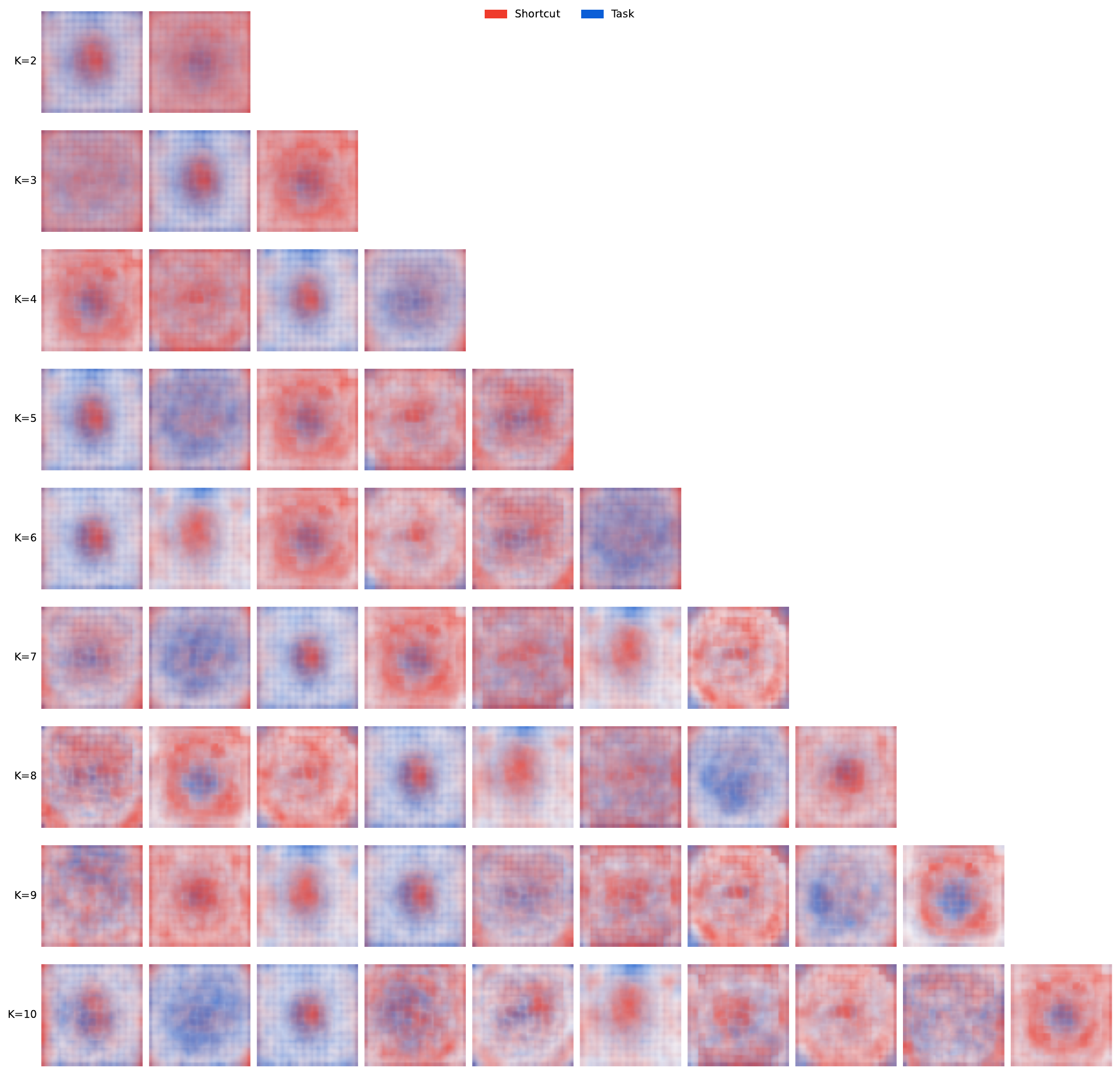}
\caption{\textbf{\textsc{ISIC2019}, ViT, AttnLRP: K-means shortcut group contribution maps.}}
\label{fig:supp_vit_isic_lrp_kmeans}
\end{figure}

\begin{figure}[p]
\centering
\includegraphics[width=\linewidth]{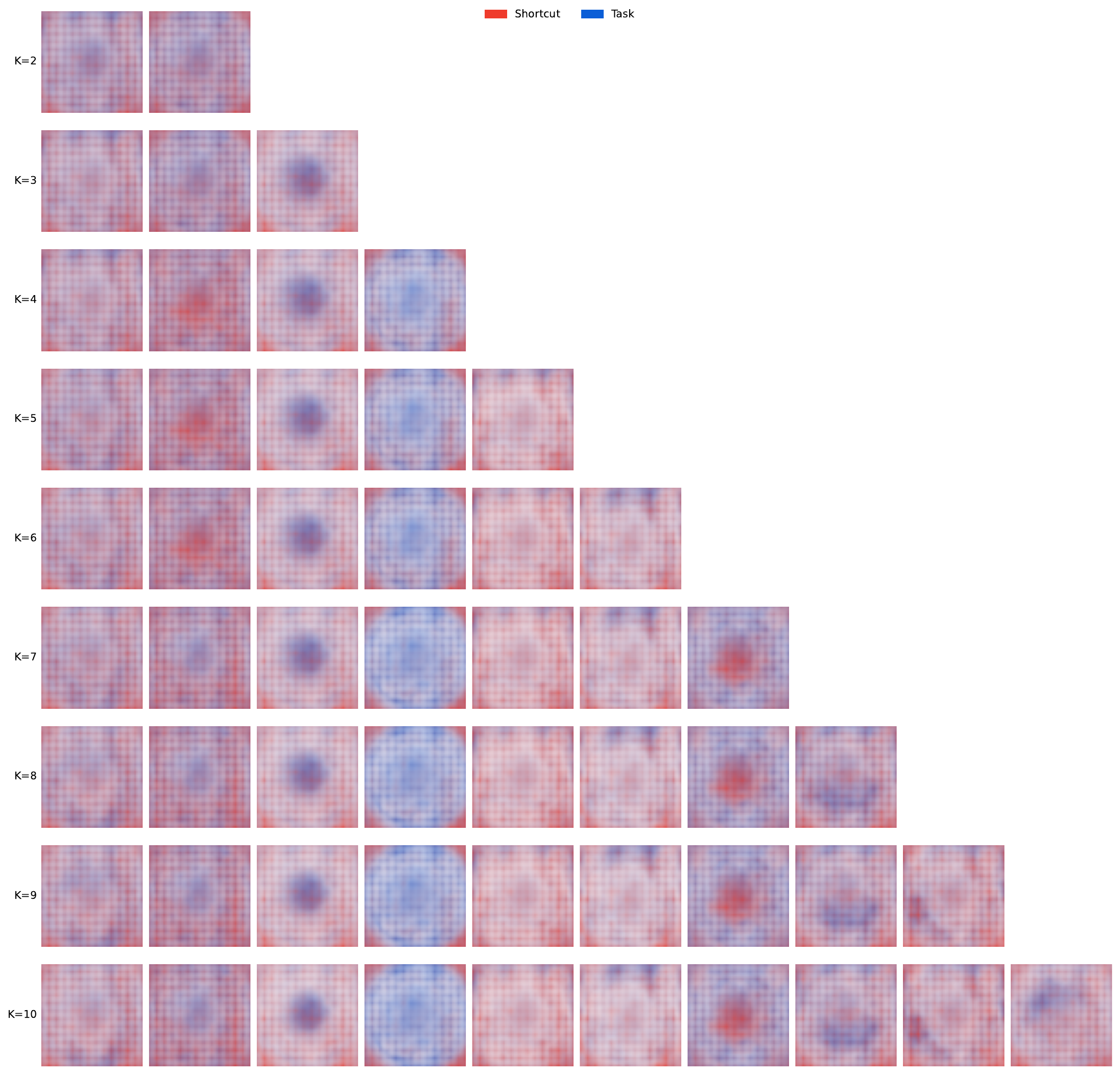}
\caption{\textbf{\textsc{ISIC2019}, ViT, TiS: NMF shortcut group contribution maps.}}
\label{fig:supp_vit_isic_tis_nmf}
\end{figure}

\begin{figure}[p]
\centering
\includegraphics[width=\linewidth]{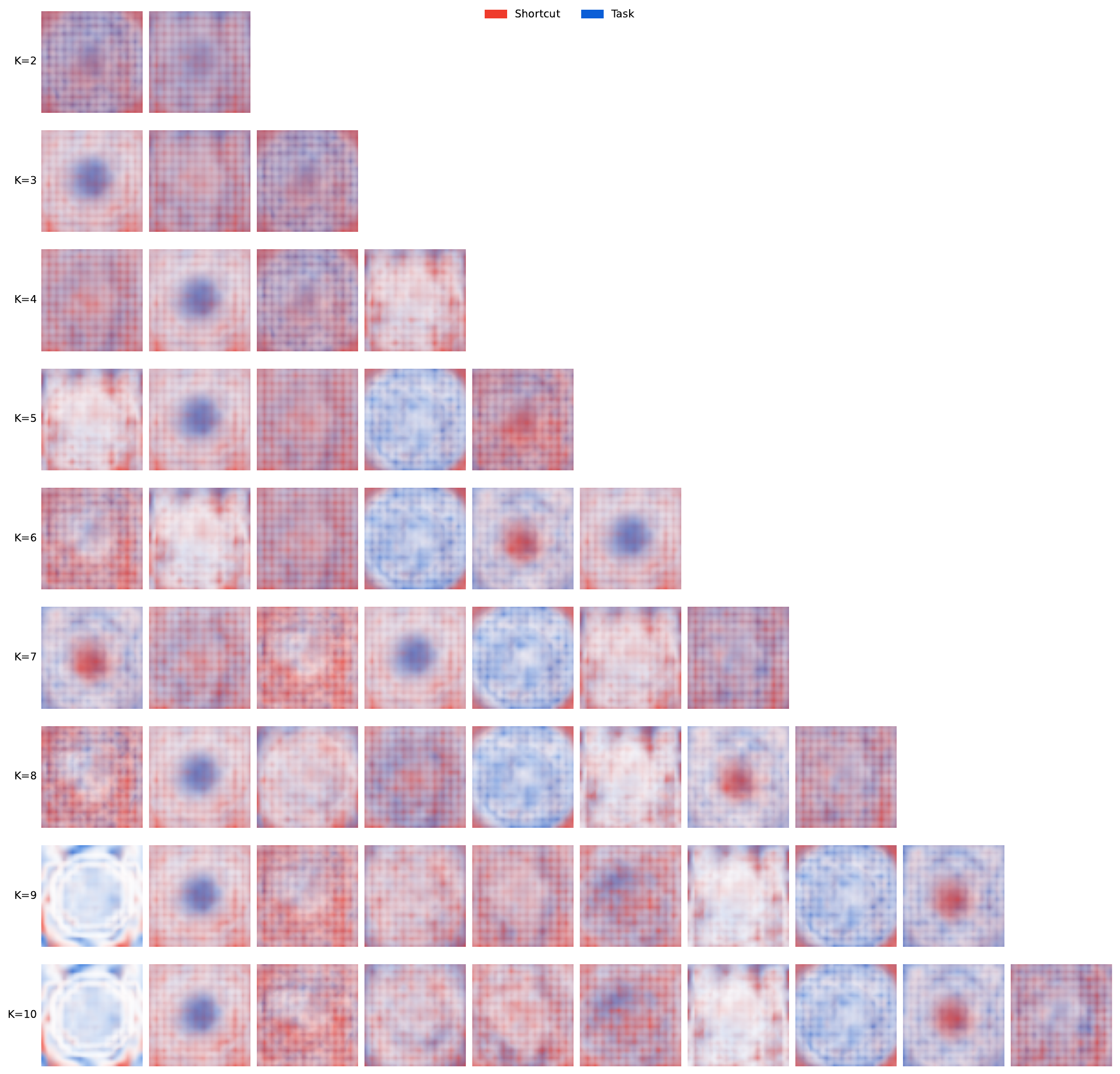}
\caption{\textbf{\textsc{ISIC2019}, ViT, TiS: K-means shortcut group contribution maps.}}
\label{fig:supp_vit_isic_tis_kmeans}
\end{figure}

\begin{figure}[p]
\centering
\includegraphics[width=\linewidth]{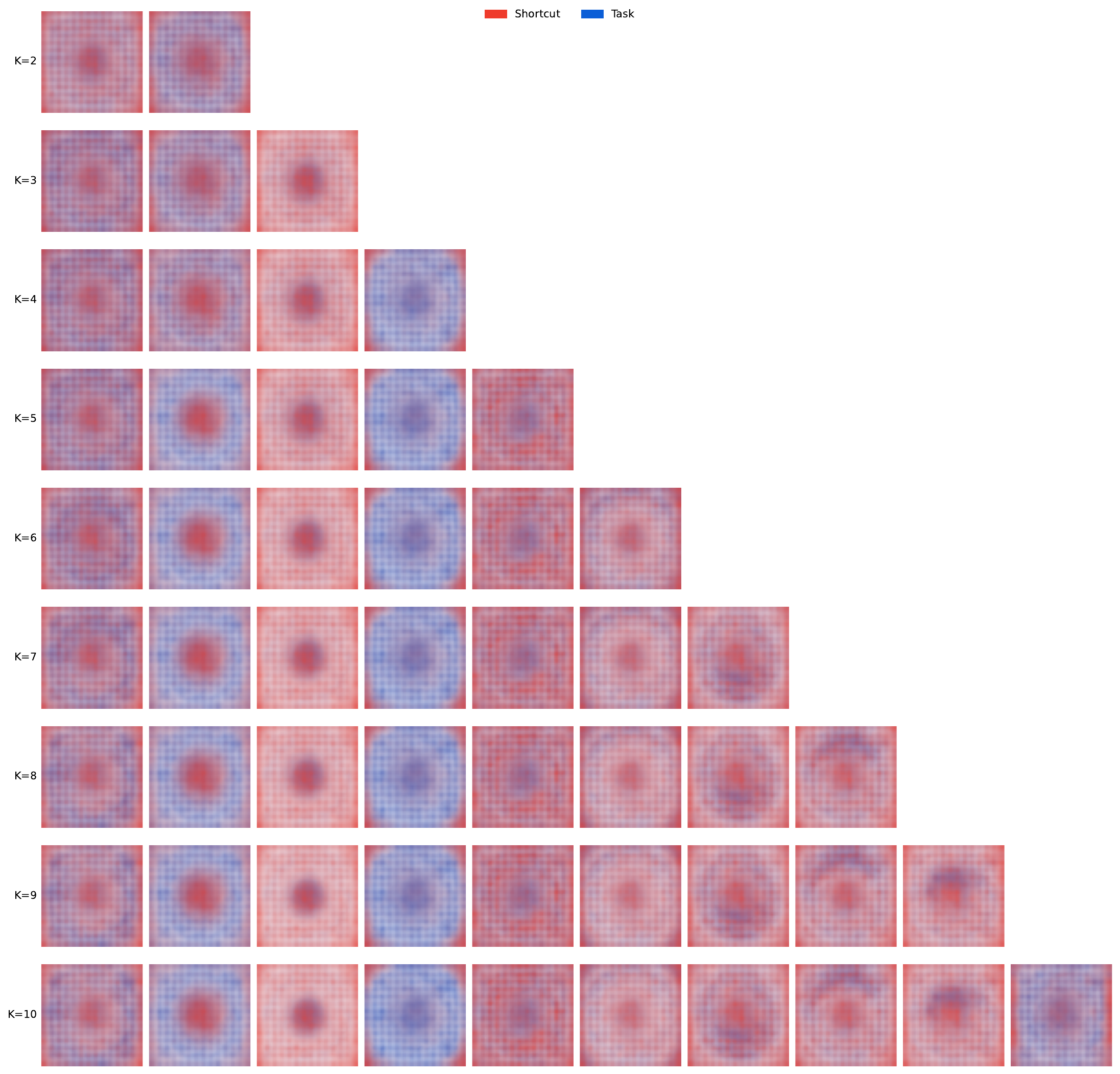}
\caption{\textbf{\textsc{ISIC2019}, ViT, CX: NMF shortcut group contribution maps.}}
\label{fig:supp_vit_isic_cx_nmf}
\end{figure}

\begin{figure}[p]
\centering
\includegraphics[width=\linewidth]{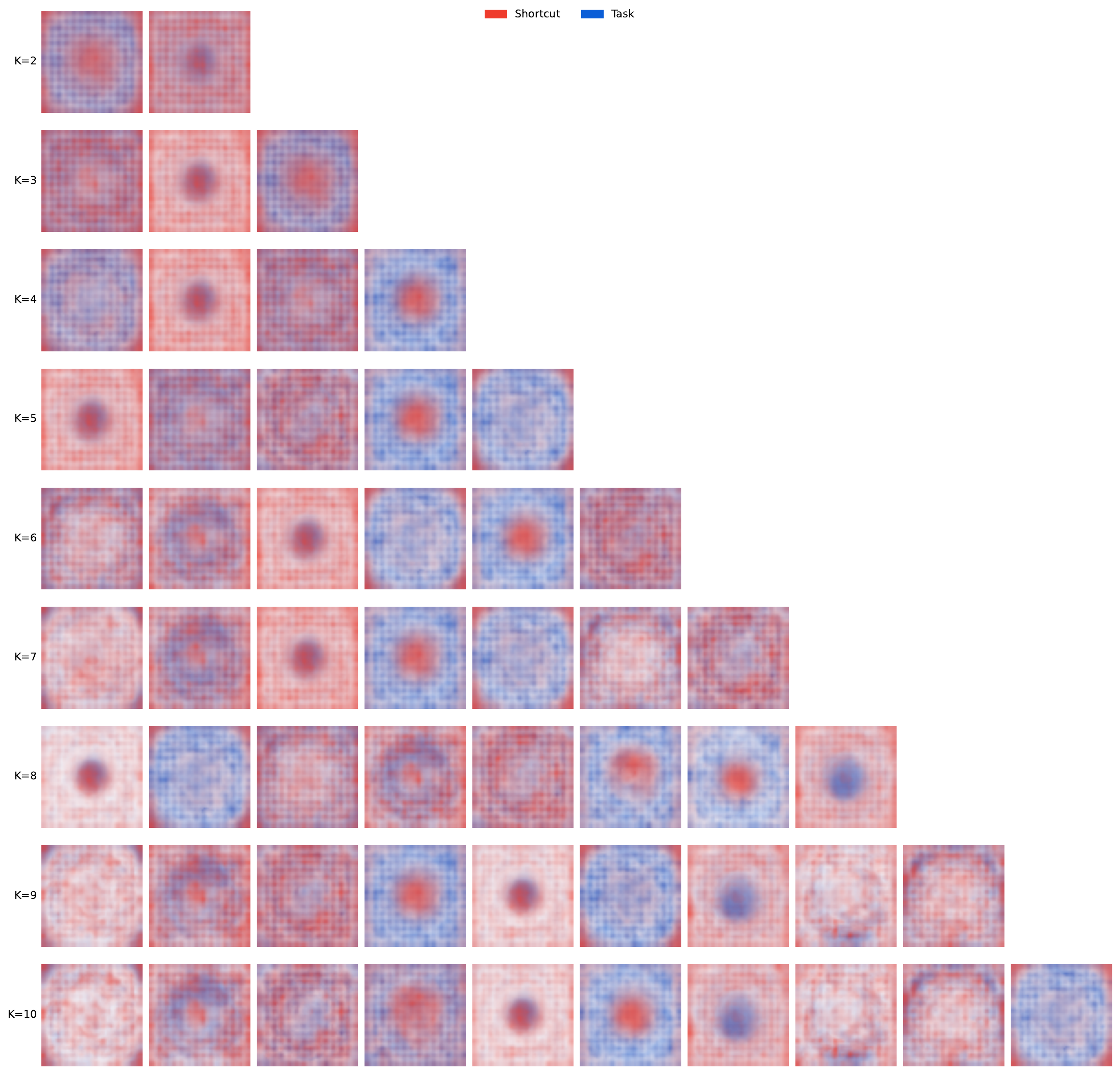}
\caption{\textbf{\textsc{ISIC2019}, ViT, CX: K-means shortcut group contribution maps.}}
\label{fig:supp_vit_isic_cx_kmeans}
\end{figure}

Figures~\ref{fig:supp_resnet_chexpert_gradcam_nmf}-\ref{fig:supp_vit_isic_cx_kmeans} provide additional shortcut group contribution maps across the medical imaging datasets, models and attribution methods with varying number of shortcut groups ($K$). We present contribution maps (using a $56\times56$ grid-based partition) for NMF and K-means-based approaches where the membership-weighted NMF figures show image-level contribution maps averaged using NMF membership weights for each shortcut group i.e. prototypes, and K-means figures show hard shortcut group contribution maps. In the following figures, each cell shows one NMF/K-means shortcut group prototype learned from the joint shortcut and task contribution representation. Orange/Red indicates shortcut contribution and blue indicates task contribution.

%% file: Tables/seed_stability.tex
\begin{table}[htbp]
\centering
\caption{\textbf{NMF derived risk score-based subsets contain more errors and are generally more consistent across independently trained models than K-means.} \emph{Errors found} is the percentage of total held-out errors among the top $20\%$ of images ranked by risk score. \emph{Intersection} is the intersection between the sets of top $20\%$ selected images, averaged across independently trained model pairs.}
\label{tab:supp_seed_stability_and_overlap_k8}
\small
\setlength{\tabcolsep}{4pt}

\begin{adjustbox}{max width=\textwidth}
\begin{tabular}{lllcc}
\toprule
Dataset & Model & Method & Errors found (\%) & Intersection \\
\midrule

\multirow{4}{*}{\textsc{CelebA}}
  & \multirow{2}{*}{ViT} & K-means & 56.9$\pm$3.6 & .365$\pm$.045 \\
  &  & NMF & \textbf{63.5$\pm$3.9} & \textbf{.402$\pm$.019} \\
  & \multirow{2}{*}{ResNet} & K-means & 55.3$\pm$3.7 & .368$\pm$.021 \\
  &  & NMF & \textbf{59.3$\pm$1.9} & \textbf{.410$\pm$.022} \\
\midrule

\multirow{4}{*}{\textsc{CheXpert}}
  & \multirow{2}{*}{ViT} & K-means & 45.3$\pm$5.8 & .368$\pm$.013 \\
  &  & NMF & \textbf{57.1$\pm$2.3} & \textbf{.526$\pm$.017} \\
  & \multirow{2}{*}{ResNet} & K-means & 34.8$\pm$1.0 & .336$\pm$.027 \\
  &  & NMF & \textbf{40.2$\pm$2.3} & \textbf{.388$\pm$.030} \\
\midrule

\multirow{4}{*}{\textsc{Waterbirds}}
  & \multirow{2}{*}{ViT} & K-means & 34.6$\pm$5.7 & .246$\pm$.030 \\
  &  & NMF & \textbf{48.1$\pm$5.8} & \textbf{.288$\pm$.017} \\
  & \multirow{2}{*}{ResNet} & K-means & 55.6$\pm$7.6 & .339$\pm$.078 \\
  &  & NMF & \textbf{73.8$\pm$2.9} & \textbf{.475$\pm$.019} \\
\midrule

\multirow{4}{*}{\textsc{Camelyon17}}
  & \multirow{2}{*}{ViT} & K-means & 31.2$\pm$3.5 & .248$\pm$.011 \\
  &  & NMF & \textbf{43.4$\pm$4.7} & \textbf{.318$\pm$.013} \\
  & \multirow{2}{*}{ResNet} & K-means & 42.7$\pm$4.9 & .287$\pm$.012 \\
  &  & NMF & \textbf{46.2$\pm$4.9} & \textbf{.311$\pm$.005} \\
\midrule

\multirow{4}{*}{\textsc{ISIC2019}}
  & \multirow{2}{*}{ViT} & K-means & 28.6$\pm$1.4 & .287$\pm$.032 \\
  &  & NMF & \textbf{40.2$\pm$4.1} & \textbf{.380$\pm$.022} \\
  & \multirow{2}{*}{ResNet} & K-means & 30.9$\pm$2.8 & \textbf{.274$\pm$.018} \\
  &  & NMF & \textbf{31.5$\pm$4.6} & .259$\pm$.012 \\
\bottomrule
\end{tabular}
\end{adjustbox}
\end{table}

%% file: Tables/auxiliary.tex
\begin{table}[t]
\centering
\caption{\textbf{Shortcut group contribution maps (prototypes) and risk scores show high stability across independently trained auxiliary models.} \emph{Non-Aux (\%)} is the \% of total held-out errors in NMF-derived, risk score-based subset when the models are unchanged, while \emph{Errors found (\%)} is the \% of total held-out errors in the top-20\% held-out samples ranked by the shortcut group risk score after auxiliary model retraining. \emph{Shortcut similarity} and \emph{task similarity} are cosine similarities between matched shortcut and task prototypes. Values are mean$\pm$standard deviation across models, except Non-Aux (fixed \(f_{\mathrm{TS}}\)), for which variation is across the five held-out splits.}
\label{tab:supp_aux_prototype_sensitivity}
\begin{adjustbox}{max width=\textwidth}
\begin{tabular}{lllcccc}
\toprule
Dataset & Model & Changed model & Non-Aux (\%) & Errors found (\%) & Shortcut similarity & Task similarity \\
\midrule
\multirow{4}{*}{\textsc{CelebA}} 
& \multirow{2}{*}{ViT} & $f_{\mathrm{BA}}$ & 68.1$\pm$4.6 & 65.4$\pm$2.9 & 0.98$\pm$0.01 & 0.98$\pm$0.01 \\
&  & $f_{\mathrm{SA}}$ & 68.1$\pm$4.6 & 67.4$\pm$1.2 & 0.98$\pm$0.01 & 0.99$\pm$0.00 \\
& \multirow{2}{*}{ResNet} & $f_{\mathrm{BA}}$ & 59.1$\pm$4.0 & 60.1$\pm$1.9 & 0.97$\pm$0.01 & 0.98$\pm$0.00 \\
&  & $f_{\mathrm{SA}}$ & 59.1$\pm$4.0 & 59.2$\pm$1.3 & 0.99$\pm$0.01 & 0.99$\pm$0.00 \\
\midrule

\multirow{4}{*}{\textsc{CheXpert}} 
& \multirow{2}{*}{ViT} & $f_{\mathrm{BA}}$ & 58.9$\pm$2.3 & 46.5$\pm$7.4 & 0.99$\pm$0.00 & 0.98$\pm$0.01 \\
&  & $f_{\mathrm{SA}}$ & 58.9$\pm$2.3 & 58.7$\pm$1.5 & 0.99$\pm$0.00 & 1.00$\pm$0.00 \\
& \multirow{2}{*}{ResNet} & $f_{\mathrm{BA}}$ & 37.8$\pm$2.8 & 41.9$\pm$2.5 & 0.94$\pm$0.03 & 0.98$\pm$0.01 \\
&  & $f_{\mathrm{SA}}$ & 37.8$\pm$2.8 & 40.5$\pm$2.7 & 0.97$\pm$0.01 & 0.96$\pm$0.02 \\
\midrule

\multirow{4}{*}{\textsc{Waterbirds}} 
& \multirow{2}{*}{ViT} & $f_{\mathrm{BA}}$ & 55.7$\pm$1.9 & 49.7$\pm$5.0 & 1.00$\pm$0.00 & 0.99$\pm$0.00 \\
&  & $f_{\mathrm{SA}}$ & 55.7$\pm$1.9 & 55.4$\pm$2.4 & 0.99$\pm$0.00 & 1.00$\pm$0.00 \\
& \multirow{2}{*}{ResNet} & $f_{\mathrm{BA}}$ & 75.9$\pm$4.5 & 74.3$\pm$1.1 & 0.99$\pm$0.00 & 1.00$\pm$0.00 \\
&  & $f_{\mathrm{SA}}$ & 75.9$\pm$4.5 & 77.4$\pm$1.0 & 0.99$\pm$0.00 & 0.99$\pm$0.00 \\
\midrule

\multirow{4}{*}{\textsc{Camelyon17}} 
& \multirow{2}{*}{ViT} & $f_{\mathrm{BA}}$ & 37.4$\pm$4.1 & 29.6$\pm$5.0 & 1.00$\pm$0.00 & 0.99$\pm$0.00 \\
&  & $f_{\mathrm{SA}}$ & 37.4$\pm$4.1 & 42.2$\pm$2.9 & 1.00$\pm$0.00 & 1.00$\pm$0.00 \\
& \multirow{2}{*}{ResNet} & $f_{\mathrm{BA}}$ & 53.9$\pm$1.7 & 38.5$\pm$9.4 & 0.99$\pm$0.01 & 0.99$\pm$0.00 \\
&  & $f_{\mathrm{SA}}$ & 53.9$\pm$1.7 & 59.4$\pm$4.1 & 0.99$\pm$0.00 & 0.99$\pm$0.00 \\
\midrule

\multirow{4}{*}{\textsc{ISIC2019}} 
& \multirow{2}{*}{ViT} & $f_{\mathrm{BA}}$ & 45.2$\pm$2.2 & 39.9$\pm$3.3 & 0.99$\pm$0.00 & 0.99$\pm$0.00 \\
&  & $f_{\mathrm{SA}}$ & 45.2$\pm$2.2 & 44.5$\pm$0.8 & 0.99$\pm$0.00 & 0.99$\pm$0.00 \\
& \multirow{2}{*}{ResNet} & $f_{\mathrm{BA}}$ & 24.7$\pm$3.1 & 28.8$\pm$4.3 & 0.96$\pm$0.01 & 0.97$\pm$0.01 \\
&  & $f_{\mathrm{SA}}$ & 24.7$\pm$3.1 & 27.6$\pm$3.3 & 0.95$\pm$0.01 & 0.96$\pm$0.01 \\
\bottomrule
\end{tabular}
\end{adjustbox}
\end{table}

%% file: Tables/quadrant_error.tex
\begin{table*}[t]
\centering
\caption{
\textbf{Combined residual-sign representations yield subsets that consistently contain high errors.} Values show the percentage of total errors captured within the top 20\% of examples ranked by shortcut-group risk score. Sign pairs $(z_1, z_2)$ denote individual quadrant regions. Values report the mean $\pm$ standard deviation across independently trained models averaged over five balanced splits.}
\label{tab:supp_quadrant_errors_found_k8}
\begin{adjustbox}{max width=\textwidth}
\begin{tabular}{llrrrrrrr}
\toprule
\textbf{Dataset} & \textbf{Model} & \textbf{$(-,-)$} & \textbf{$(-,+)$} & \textbf{$(+,-)$} & \textbf{$(+,+)$} & \textbf{Positive} & \textbf{Negative} & \textbf{Combined} \\
\midrule
\multirow{2}{*}{\textsc{CelebA}}
 & ViT & 46.4$\pm$4.6 & 46.8$\pm$6.7 & 52.2$\pm$3.3 & 45.9$\pm$3.9 & 63.5$\pm$3.9 & 66.1$\pm$3.4 & \textbf{66.5$\pm$3.8} \\
 & ResNet & 55.4$\pm$3.0 & 52.6$\pm$2.6 & 59.8$\pm$4.3 & 58.3$\pm$2.5 & 59.3$\pm$1.9 & 62.5$\pm$2.2 & \textbf{65.7$\pm$2.7} \\
\midrule
\multirow{2}{*}{\textsc{CheXpert}}
 & ViT & 40.5$\pm$6.3 & 43.0$\pm$6.7 & 43.3$\pm$4.1 & 44.6$\pm$3.4 & 57.1$\pm$2.3 & 58.9$\pm$2.5 & \textbf{61.1$\pm$2.7} \\
 & ResNet & 34.1$\pm$2.1 & 38.2$\pm$3.9 & 38.0$\pm$1.3 & 37.5$\pm$2.8 & 40.2$\pm$2.3 & 39.3$\pm$1.1 & \textbf{45.6$\pm$1.1} \\
\midrule
\multirow{2}{*}{\textsc{Waterbirds}}
 & ViT & 30.7$\pm$4.9 & 26.8$\pm$1.9 & 40.0$\pm$6.7 & 32.6$\pm$5.9 & 48.1$\pm$5.8 & 49.1$\pm$3.7 & \textbf{52.5$\pm$5.6} \\
 & ResNet & 65.4$\pm$4.4 & 74.8$\pm$4.1 & 71.7$\pm$7.0 & 69.2$\pm$3.1 & 73.8$\pm$2.9 & 79.7$\pm$3.7 & \textbf{81.9$\pm$2.1} \\
\midrule
\multirow{2}{*}{\textsc{Camelyon17}}
 & ViT & 23.2$\pm$1.2 & 21.7$\pm$1.4 & 25.1$\pm$2.7 & 34.0$\pm$3.5 & 43.4$\pm$4.7 & 41.4$\pm$3.2 & \textbf{54.8$\pm$8.6} \\
 & ResNet & 41.0$\pm$4.9 & 36.6$\pm$6.0 & 36.9$\pm$5.0 & 46.7$\pm$2.8 & 46.2$\pm$4.9 & 45.1$\pm$4.2 & \textbf{54.1$\pm$4.2} \\
\midrule
\multirow{2}{*}{\textsc{ISIC2019}}
 & ViT & 28.5$\pm$3.8 & 29.1$\pm$5.2 & 28.5$\pm$1.0 & 29.2$\pm$3.2 & 40.2$\pm$4.1 & 42.1$\pm$1.8 & \textbf{46.0$\pm$4.9} \\
 & ResNet & 28.6$\pm$3.5 & 29.6$\pm$3.7 & 29.4$\pm$3.0 & 28.4$\pm$3.9 & 31.5$\pm$4.6 & 33.9$\pm$3.5 & \textbf{38.5$\pm$3.6} \\
\bottomrule
\end{tabular}
\end{adjustbox}
\end{table*}

%% file: Tables/risk_ablation.tex
\begin{table*}[t]
\centering
\caption{
\textbf{Subsets derived from negative task alignment ($-\rho_{task}$) capture the highest percentage of errors, while positive shortcut--task representations ($m^{sc}, m^{task}$) also contain substantial errors across all datasets.} Values are the mean $\pm$ standard deviation across independently trained models averaged over five held-out sets.}
\label{tab:error_capture_score_map_baselines}
\begin{adjustbox}{max width=\textwidth}
\begin{tabular}{llrrrrrr}
\toprule
Dataset & Model & $m^{sc},m^{task}$ & $\rho_{sc}$ & $-\rho_{task}$ & $m^{sc}$ & $m^{task}$ & $\textbf{s}(f_{TS},x,\hat{y}_{TS})$ \\
\midrule
\multirow{2}{*}{\textsc{CelebA}}
 & ViT & 63.5$\pm$3.9 & 36.5$\pm$8.3 & \textbf{64.8$\pm$3.6} & 30.8$\pm$3.0 & 32.9$\pm$4.7 & 33.7$\pm$4.0 \\
 & ResNet & 59.3$\pm$1.9 & 35.0$\pm$3.2 & \textbf{63.5$\pm$1.7} & 29.0$\pm$3.8 & 42.1$\pm$4.7 & 57.4$\pm$3.7 \\
\midrule
\multirow{2}{*}{\textsc{CheXpert}}
 & ViT & 57.1$\pm$2.3 & 27.8$\pm$3.0 & \textbf{60.8$\pm$2.7} & 25.8$\pm$2.3 & 30.8$\pm$4.9 & 32.6$\pm$3.0 \\
 & ResNet & 40.2$\pm$2.3 & 25.1$\pm$4.5 & \textbf{43.0$\pm$2.7} & 23.6$\pm$4.0 & 31.9$\pm$5.2 & 30.3$\pm$3.5 \\
\midrule
\multirow{2}{*}{\textsc{Waterbirds}}
 & ViT & \textbf{48.1$\pm$5.8} & 29.8$\pm$3.8 & 46.2$\pm$8.0 & 22.6$\pm$2.4 & 24.6$\pm$3.4 & 29.2$\pm$8.2 \\
 & ResNet & 73.8$\pm$2.9 & 36.7$\pm$5.9 & \textbf{79.5$\pm$1.3} & 20.4$\pm$3.8 & 35.0$\pm$4.2 & 69.8$\pm$5.6 \\
\midrule
\multirow{2}{*}{\textsc{Camelyon17}}
 & ViT & 43.4$\pm$4.7 & 17.1$\pm$5.3 & \textbf{55.3$\pm$8.9} & 23.7$\pm$2.4 & 25.0$\pm$2.1 & 21.6$\pm$1.7 \\
 & ResNet & 46.2$\pm$4.9 & 19.7$\pm$4.9 & \textbf{53.6$\pm$4.2} & 25.0$\pm$4.6 & 26.9$\pm$2.2 & 34.5$\pm$2.9 \\
\midrule
\multirow{2}{*}{\textsc{ISIC2019}}
 & ViT & 40.2$\pm$4.1 & 20.7$\pm$2.6 & \textbf{44.2$\pm$3.5} & 24.2$\pm$1.8 & 26.8$\pm$2.3 & 24.6$\pm$2.6 \\
 & ResNet & 31.5$\pm$4.6 & 18.8$\pm$4.0 & \textbf{34.5$\pm$3.8} & 22.7$\pm$3.2 & 25.5$\pm$1.9 & 26.1$\pm$1.9 \\
\bottomrule
\end{tabular}
\end{adjustbox}
\end{table*}

%% file: Tables/occlusion_appendix.tex
\begin{table*}[t]
\centering
\caption{
\textbf{Masking task regions degrades model performance across all datasets, while masking shortcut regions either results in lower performance degradations or shows slight improvements.} Random masking values are averaged over 5 random masks for each independently trained model and reported with standard deviation across models. Blue and red indicate the highest increase (or smallest drop) and largest drop, respectively, for each dataset--model pair.
}
\label{tab:input_occlusion_56x56}
\resizebox{\textwidth}{!}{%
\begin{tabular}{lllrrrrrr}
\toprule
Dataset & Model & Source
& \multicolumn{2}{c}{Shortcut}
& \multicolumn{2}{c}{Task}
& \multicolumn{2}{c}{Random} \\
\cmidrule(lr){4-5}
\cmidrule(lr){6-7}
\cmidrule(lr){8-9}
& &
& $\Delta\mathrm{LPG}_{\mathrm{red}}$ & $\Delta$Acc.
& $\Delta\mathrm{LPG}_{\mathrm{red}}$ & $\Delta$Acc.
& $\Delta\mathrm{LPG}_{\mathrm{red}}$ & $\Delta$Acc. \\
\midrule
\multirow{6}{*}{CelebA}
& \multirow{3}{*}{ResNet} & Dataset
& \textcolor{blue}{+2.3} & \textcolor{blue}{+.9}
& -34.2 & -8.6
& $-19.6{\pm}7.4$ & $-5.1{\pm}2.3$ \\
& & Shortcut Group
& -1.1 & -.3
& \textcolor{red}{-39.4} & \textcolor{red}{-10.1}
& $-17.9{\pm}6.7$ & $-5.1{\pm}1.8$ \\
& & Image
& +1.5 & -.2
& -32.8 & -9.5
& $-16.7{\pm}6.9$ & $-4.2{\pm}2.1$ \\

\cmidrule(lr){2-9}

& \multirow{3}{*}{ViT} & Dataset
& -5.5 & -.8
& -44.5 & -10.3
& $-10.4{\pm}5.0$ & $-2.3{\pm}.8$ \\
& & Shortcut Group
& \textcolor{blue}{-4.3} & -.8
& \textcolor{red}{-49.6} & \textcolor{red}{-12.3}
& $-11.8{\pm}3.1$ & $-2.8{\pm}.5$ \\
& & Image
& -5.0 & \textcolor{blue}{-.3}
& -26.3 & -6.5
& $-9.7{\pm}2.2$ & $-2.2{\pm}.5$ \\

\midrule

\multirow{6}{*}{CheXpert}
& \multirow{3}{*}{ResNet} & Dataset
& -4.9 & -1.7
& -.2 & -3.5
& $-13.1{\pm}10.1$ & $-8.7{\pm}2.2$ \\
& & Shortcut Group
& +6.6 & \textcolor{blue}{+1.0}
& -6.2 & -5.1
& $-14.3{\pm}13.6$ & $-9.5{\pm}3.9$ \\
& & Image
& \textcolor{blue}{+9.0} & +.5
& -5.5 & -4.0
& $\textcolor{red}{-17.3{\pm}12.0}$ & $\textcolor{red}{-10.4{\pm}3.6}$ \\

\cmidrule(lr){2-9}

& \multirow{3}{*}{ViT} & Dataset
& -.4 & -2.9
& -17.4 & -8.7
& $\textcolor{red}{-36.0{\pm}8.4}$ & $-10.6{\pm}1.8$ \\
& & Shortcut Group
& \textcolor{blue}{+8.6} & \textcolor{blue}{+.7}
& -24.3 & -11.4
& $-35.0{\pm}8.3$ & $-10.6{\pm}2.3$ \\
& & Image
& -9.4 & -3.8
& -19.4 & \textcolor{red}{-13.8}
& $-34.7{\pm}8.9$ & $-10.6{\pm}2.4$ \\

\midrule

\multirow{6}{*}{Waterbirds}
& \multirow{3}{*}{ResNet} & Dataset
& -1.9 & +.9
& -52.7 & -24.2
& $-16.7{\pm}1.0$ & $-7.7{\pm}1.1$ \\
& & Shortcut Group
& \textcolor{blue}{-.8} & \textcolor{blue}{+1.4}
& \textcolor{red}{-58.9} & \textcolor{red}{-26.3}
& $-16.4{\pm}1.3$ & $-7.7{\pm}1.0$ \\
& & Image
& -5.9 & -.1
& -54.0 & -20.9
& $-15.6{\pm}.6$ & $-7.3{\pm}1.0$ \\

\cmidrule(lr){2-9}

& \multirow{3}{*}{ViT} & Dataset
& -2.6 & -.1
& -33.0 & -16.8
& $-10.6{\pm}10.0$ & $-4.7{\pm}1.4$ \\
& & Shortcut Group
& \textcolor{blue}{-1.6} & \textcolor{blue}{+.9}
& \textcolor{red}{-40.1} & \textcolor{red}{-19.6}
& $-11.8{\pm}10.0$ & $-5.2{\pm}1.3$ \\
& & Image
& -5.8 & -.9
& -38.9 & -14.5
& $-9.4{\pm}9.8$ & $-4.6{\pm}1.2$ \\

\midrule

\multirow{6}{*}{Camelyon17}
& \multirow{3}{*}{ResNet} & Dataset
& -12.1 & -5.1
& -9.4 & -3.6
& $-43.7{\pm}14.0$ & $-23.5{\pm}8.1$ \\
& & Shortcut Group
& -8.4 & \textcolor{blue}{-3.1}
& -15.1 & -6.0
& $\textcolor{red}{-45.5{\pm}14.1}$ & $\textcolor{red}{-24.2{\pm}7.9}$ \\
& & Image
& \textcolor{blue}{-6.9} & -3.1
& -21.2 & -10.2
& $-39.7{\pm}12.8$ & $-22.2{\pm}7.1$ \\

\cmidrule(lr){2-9}

& \multirow{3}{*}{ViT} & Dataset
& \textcolor{blue}{-4.8} & -2.3
& -5.6 & \textcolor{blue}{-2.0}
& $-9.6{\pm}3.5$ & $-8.2{\pm}2.2$ \\
& & Shortcut Group
& -10.7 & -5.1
& -17.2 & -7.5
& $-7.5{\pm}3.5$ & $-6.6{\pm}1.6$ \\
& & Image
& -10.4 & -3.4
& \textcolor{red}{-20.1} & \textcolor{red}{-9.1}
& $-9.3{\pm}3.3$ & $-7.6{\pm}1.9$ \\

\midrule

\multirow{6}{*}{ISIC2019}
& \multirow{3}{*}{ResNet} & Dataset
& -8.9 & -1.2
& -17.0 & -4.0
& $-24.9{\pm}9.5$ & $-9.8{\pm}4.4$ \\
& & Shortcut Group
& \textcolor{blue}{-1.5} & \textcolor{blue}{-.6}
& -6.6 & -2.4
& $\textcolor{red}{-27.3{\pm}10.8}$ & $\textcolor{red}{-10.6{\pm}4.3}$ \\
& & Image
& -9.2 & -1.4
& -14.2 & -3.5
& $-26.4{\pm}11.0$ & $-10.3{\pm}4.9$ \\

\cmidrule(lr){2-9}

& \multirow{3}{*}{ViT} & Dataset
& +1.7 & -.8
& \textcolor{red}{-3.0} & -2.0
& $-2.9{\pm}14.1$ & $\textcolor{red}{-2.8{\pm}1.8}$ \\
& & Shortcut Group
& \textcolor{blue}{+5.5} & \textcolor{blue}{+.5}
& +4.0 & -.7
& $-2.0{\pm}12.6$ & $-2.5{\pm}1.3$ \\
& & Image
& +2.5 & -.6
& +1.2 & -1.4
& $-.4{\pm}10.9$ & $-2.1{\pm}1.6$ \\
\bottomrule
\end{tabular}%
}
\end{table*}

%% file: Tables/source_comparison_intervention.tex
\begin{table*}[t]
\centering
\caption{
\textbf{Shortcut group-level combined interventions result in the strongest improvements in LPG reduction and accuracy across most datasets.} Values are the mean change in LPG reduction and accuracy relative to no-intervention with standard deviation across independently trained models.
}
\label{tab:supp_combined_source_comparison}
\resizebox{\textwidth}{!}{%
\begin{tabular}{llrrrrrr}
\toprule
Dataset & Model
& \multicolumn{2}{c}{Dataset}
& \multicolumn{2}{c}{Shortcut Group}
& \multicolumn{2}{c}{Image} \\
\cmidrule(lr){3-4}
\cmidrule(lr){5-6}
\cmidrule(lr){7-8}
& & $\Delta\mathrm{LPG}_{\mathrm{red}}$ & $\Delta$Acc.
& $\Delta\mathrm{LPG}_{\mathrm{red}}$ & $\Delta$Acc.
& $\Delta\mathrm{LPG}_{\mathrm{red}}$ & $\Delta$Acc. \\
\midrule

\multirow{2}{*}{\textsc{CelebA}}
& ResNet
& $+4.9{\pm}2.3$ & $+1.4{\pm}.2$
& $\textcolor{blue}{+7.7{\pm}5.0}$ & $\textcolor{blue}{+1.6{\pm}.5}$
& $+5.3{\pm}1.7$ & $+.8{\pm}.4$ \\
& ViT
& $+2.1{\pm}2.8$ & $+1.1{\pm}.4$
& $\textcolor{blue}{+4.0{\pm}4.4}$ & $\textcolor{blue}{+1.3{\pm}1.1}$
& $+3.1{\pm}2.4$ & $+.8{\pm}.6$ \\
\midrule

\multirow{2}{*}{\textsc{CheXpert}}
& ResNet
& $-3.0{\pm}4.4$ & $-.7{\pm}.5$
& $+.1{\pm}4.5$ & $+.3{\pm}1.3$
& $\textcolor{blue}{+5.2{\pm}3.9}$ & $\textcolor{blue}{+1.4{\pm}1.3}$ \\
& ViT
& $+5.4{\pm}4.1$ & $+.8{\pm}1.9$
& $\textcolor{blue}{+12.8{\pm}7.6}$ & $\textcolor{blue}{+3.2{\pm}.5}$
& $+6.5{\pm}1.6$ & $+.8{\pm}.1$ \\
\midrule

\multirow{2}{*}{\textsc{Waterbirds}}
& ResNet
& $+2.6{\pm}1.5$ & $+2.7{\pm}.6$
& $\textcolor{blue}{+5.4{\pm}1.5}$ & $\textcolor{blue}{+3.7{\pm}.7}$
& $+.6{\pm}.4$ & $+.8{\pm}.2$ \\
& ViT
& $+6.2{\pm}.4$ & $+2.3{\pm}.3$
& $\textcolor{blue}{+9.7{\pm}.4}$ & $\textcolor{blue}{+3.7{\pm}.4}$
& $+3.4{\pm}2.0$ & $+1.0{\pm}.9$ \\
\midrule

\multirow{2}{*}{\textsc{Camelyon17}}
& ResNet
& $-.5{\pm}4.3$ & $-.4{\pm}1.4$
& $\textcolor{blue}{+3.5{\pm}1.6}$ & $+1.0{\pm}.3$
& $+3.2{\pm}.4$ & $\textcolor{blue}{+1.1{\pm}1.0}$ \\
& ViT
& $-.3{\pm}.7$ & $-.3{\pm}.4$
& $\textcolor{blue}{+3.1{\pm}.9}$ & $\textcolor{blue}{+.7{\pm}.3}$
& $+1.8{\pm}.7$ & $+.7{\pm}.4$ \\
\midrule

\multirow{2}{*}{\textsc{ISIC2019}}
& ResNet
& $+.7{\pm}4.2$ & $+.1{\pm}1.1$
& $\textcolor{blue}{+1.0{\pm}2.7}$ & $\textcolor{blue}{+.3{\pm}.7}$
& $-.5{\pm}1.5$ & $-.1{\pm}.8$ \\
& ViT
& $+1.3{\pm}7.3$ & $-.2{\pm}.7$
& $\textcolor{blue}{+7.4{\pm}5.8}$ & $+.3{\pm}.8$
& $\textcolor{blue}{+7.4{\pm}3.2}$ & $\textcolor{blue}{+1.0{\pm}.2}$ \\

\bottomrule
\end{tabular}%
}
\end{table*}

%% file: Tables/intervention_seeds.tex
\begin{table*}[t]
\centering
\caption{
\textbf{Intervention stability across independently trained models.} Values are mean$\pm$standard deviation changes relative to the corresponding model without intervention.
}
\label{tab:supp_intervention_seed_source_stability}
\resizebox{0.95\textwidth}{!}{
\begin{tabular*}{\textwidth}{@{\extracolsep{\fill}}llrrrr@{}}
\toprule
Dataset & Model
& \multicolumn{2}{c}{Shortcut Group}
& \multicolumn{2}{c}{Image} \\
\cmidrule(lr){3-4}
\cmidrule(lr){5-6}
& & $\Delta\mathrm{LPG}_{\mathrm{red}}$ & $\Delta$Acc.
& $\Delta\mathrm{LPG}_{\mathrm{red}}$ & $\Delta$Acc. \\
\midrule
\multicolumn{6}{l}{\textit{Shortcut suppression}} \\
\midrule
\multirow{2}{*}{\textsc{CelebA}}
& ResNet & $+5.8{\pm}3.9$ & $+1.7{\pm}.4$ & $+6.3{\pm}2.7$ & $+1.1{\pm}.5$ \\
& ViT & $+1.4{\pm}4.4$ & $+.9{\pm}.9$ & $+3.7{\pm}3.1$ & $+1.2{\pm}.7$ \\
\midrule
\multirow{2}{*}{\textsc{CheXpert}}
& ResNet & $-1.0{\pm}1.9$ & $+.5{\pm}.8$ & $0.0{\pm}2.5$ & $+.1{\pm}.9$ \\
& ViT & $+13.8{\pm}4.2$ & $+3.0{\pm}.7$ & $+12.7{\pm}4.0$ & $+2.0{\pm}.8$ \\
\midrule
\multirow{2}{*}{\textsc{Waterbirds}}
& ResNet & $+5.2{\pm}1.9$ & $+2.5{\pm}.7$ & $+3.2{\pm}1.3$ & $+1.3{\pm}.2$ \\
& ViT & $+5.2{\pm}2.5$ & $+2.8{\pm}.2$ & $-2.1{\pm}3.8$ & $-.4{\pm}.8$ \\
\midrule
\multirow{2}{*}{\textsc{Camelyon17}}
& ResNet & $-.7{\pm}1.6$ & $-.2{\pm}.7$ & $-1.0{\pm}.8$ & $-.2{\pm}.5$ \\
& ViT & $+1.7{\pm}2.5$ & $+.2{\pm}.7$ & $+.6{\pm}1.4$ & $+.2{\pm}.8$ \\
\midrule
\multirow{2}{*}{\textsc{ISIC2019}}
& ResNet & $-.5{\pm}2.0$ & $-.4{\pm}.5$ & $0.0{\pm}1.4$ & $-.3{\pm}.7$ \\
& ViT & $-1.2{\pm}2.7$ & $+.6{\pm}.4$ & $+1.6{\pm}2.8$ & $+1.1{\pm}.1$ \\
\midrule
\multicolumn{6}{l}{\textit{Task amplification}} \\
\midrule
\multirow{2}{*}{\textsc{CelebA}}
& ResNet & $+3.4{\pm}3.2$ & $+1.0{\pm}.4$ & $0.0{\pm}1.5$ & $-.1{\pm}.2$ \\
& ViT & $+2.8{\pm}1.5$ & $+1.0{\pm}.9$ & $-.2{\pm}2.4$ & $-.3{\pm}.9$ \\
\midrule
\multirow{2}{*}{\textsc{CheXpert}}
& ResNet & $+1.7{\pm}3.8$ & $+.2{\pm}1.2$ & $+4.0{\pm}3.0$ & $+.4{\pm}.8$ \\
& ViT & $-3.1{\pm}4.0$ & $-.3{\pm}.5$ & $-2.1{\pm}1.5$ & $-.6{\pm}.3$ \\
\midrule
\multirow{2}{*}{\textsc{Waterbirds}}
& ResNet & $+2.8{\pm}1.5$ & $+2.8{\pm}.4$ & $-.2{\pm}1.1$ & $+.2{\pm}.2$ \\
& ViT & $+6.5{\pm}2.2$ & $+1.9{\pm}.1$ & $+2.7{\pm}1.8$ & $+.3{\pm}.5$ \\
\midrule
\multirow{2}{*}{\textsc{Camelyon17}}
& ResNet & $+4.3{\pm}.8$ & $+1.4{\pm}.3$ & $+2.9{\pm}.5$ & $+1.2{\pm}.7$ \\
& ViT & $+1.0{\pm}1.3$ & $+.3{\pm}.2$ & $+1.4{\pm}.8$ & $+.6{\pm}.3$ \\
\midrule
\multirow{2}{*}{\textsc{ISIC2019}}
& ResNet & $+.9{\pm}3.2$ & $+.4{\pm}.8$ & $-.2{\pm}1.8$ & $+.2{\pm}.5$ \\
& ViT & $+6.4{\pm}5.1$ & $0.0{\pm}.8$ & $+5.2{\pm}3.4$ & $+.1{\pm}.1$ \\

\bottomrule
\end{tabular*}
}
\end{table*}

%% file: Tables/intervention_layers.tex
\begin{table*}[t]
\centering
\caption{
\textbf{Combined shortcut suppression and task amplification-based internal intervention target layer comparison.} Values are mean changes relative to the no-intervention model, averaged over all five datasets across independently trained models.
}
\label{tab:intervention_site_appendix}

\begin{subtable}[t]{0.49\textwidth}
\centering
\caption{ResNet}
\begin{adjustbox}{max width=\linewidth}
\begin{tabular}{@{}llrrrr@{}}
\toprule
Source & Metric & Layer 4 & Last 2 & Last 3 & All \\
\midrule
\multirow{2}{*}{Shortcut Group}
& $\Delta\mathrm{LPG}_{\mathrm{red}}$ & \underline{+2.6} & \textbf{+3.5} & +1.5 & -11.9 \\
& $\Delta$Acc. & +.9 & \textbf{+1.4} & \underline{+.9} & -3.3 \\
\addlinespace
\multirow{2}{*}{Image}
& $\Delta\mathrm{LPG}_{\mathrm{red}}$ & +1.4 & \underline{+2.8} & \textbf{+3.8} & -7.5 \\
& $\Delta$Acc. & +.4 & \underline{+.8} & \textbf{+.9} & -2.1 \\
\bottomrule
\end{tabular}
\end{adjustbox}
\end{subtable}
\hfill
\begin{subtable}[t]{0.49\textwidth}
\centering
\caption{ViT}
\begin{adjustbox}{max width=\linewidth}
\begin{tabular}{@{}llrrrr@{}}
\toprule
Source & Metric & Last & Last 3 & Last 6 & All \\
\midrule
\multirow{2}{*}{Shortcut Group}
& $\Delta\mathrm{LPG}_{\mathrm{red}}$ & +1.5 & +2.8 & \underline{+6.3} & \textbf{+7.4} \\
& $\Delta$Acc. & +.4 & +1.0 & \textbf{+2.1} & \underline{+1.8} \\
\addlinespace
\multirow{2}{*}{Image}
& $\Delta\mathrm{LPG}_{\mathrm{red}}$ & +.6 & +1.9 & \underline{+3.9} & \textbf{+4.4} \\
& $\Delta$Acc. & +.2 & +.5 & \textbf{+1.0} & \underline{+.9} \\
\bottomrule
\end{tabular}
\end{adjustbox}
\end{subtable}

\end{table*}

%% file: Tables/vit_pathways.tex
\begin{table*}[t]
\centering
\caption{
\textbf{Combined intervention on value vector results in reduction in disparities and improvements in accuracy.} Values are mean changes relative to no intervention, averaged over all five datasets and independently trained models. Blue denotes the largest improvement within each contribution map source and metric.
}
\label{tab:vit_pathways_56x56}
\begin{tabular}{lrrrrrr}
\toprule
Intervention target
& \multicolumn{2}{c}{Dataset}
& \multicolumn{2}{c}{Shortcut Group}
& \multicolumn{2}{c}{Image} \\
\cmidrule(lr){2-3}
\cmidrule(lr){4-5}
\cmidrule(lr){6-7}
& $\Delta\mathrm{LPG}_{\mathrm{red}}$ & $\Delta$Acc.
& $\Delta\mathrm{LPG}_{\mathrm{red}}$ & $\Delta$Acc.
& $\Delta\mathrm{LPG}_{\mathrm{red}}$ & $\Delta$Acc. \\
\midrule
V
& \textcolor{blue}{+2.9} & \textcolor{blue}{+.7}
& \textcolor{blue}{+7.4} & \textcolor{blue}{+1.8}
& \textcolor{blue}{+4.4} & \textcolor{blue}{+.9} \\

V + MLP
& +1.0 & +.2
& +6.4 & +1.4
& +1.7 & 0.0 \\

Q + K + V + MLP
& -22.4 & -9.2
& -19.9 & -10.5
& -26.1 & -11.9 \\

Q + K + V
& -16.9 & -7.1
& -19.8 & -9.2
& -22.1 & -10.4 \\

Q + K
& -8.8 & -4.3
& -6.6 & -4.6
& -12.2 & -6.1 \\

MLP
& -3.3 & -.8
& -.3 & 0.0
& -.9 & -1.1 \\
\bottomrule
\end{tabular}
\end{table*}

%% file: Tables/resnet_iterative_interventions.tex
\begin{table*}[t]
\centering
\caption{
\textbf{Iterative interventions can decrease subgroup disparities beyond a single intervention.} Values are mean changes relative to no intervention across independently trained models. The $\bar{\rho}_{\mathrm{sc}_0}$ stage is the last iteration with a positive mean shortcut partial correlation across test images (or closest to $0$ if mean across all iterations is negative). For iteration $1$ and $5$, we also show the mean shortcut partial correlation $\bar{\rho}_{\mathrm{sc}}$. Blue denotes the largest improvement per dataset.
}
\label{tab:supp_resnet_iterative_diagnostics}
\resizebox{\textwidth}{!}{%
\begin{tabular}{llrrrrrrrr}
\toprule
Dataset & Source
& \multicolumn{3}{c}{Iteration $1$}
& \multicolumn{2}{c}{$\bar{\rho}_{\mathrm{sc}_0}$}
& \multicolumn{3}{c}{Iteration $5$} \\
\cmidrule(lr){3-5}
\cmidrule(lr){6-7}
\cmidrule(lr){8-10}
&
& $\bar{\rho}_{\mathrm{sc}}$ & $\Delta\mathrm{LPG}_{\mathrm{red}}$ & $\Delta$Acc.
& $\Delta\mathrm{LPG}_{\mathrm{red}}$ & $\Delta$Acc.
& $\bar{\rho}_{\mathrm{sc}}$ & $\Delta\mathrm{LPG}_{\mathrm{red}}$ & $\Delta$Acc. \\
\midrule

\multirow{2}{*}{CelebA}
& Shortcut Group
& -.301
& +7.7
& \textcolor{blue}{+1.6}
& +7.7
& \textcolor{blue}{+1.6}
& -.397
& \textcolor{blue}{+11.3}
& +1.5 \\
& Image
& -.240
& +5.3
& +.8
& +5.3
& +.8
& -.352
& +6.6
& +1.3 \\

\midrule

\multirow{2}{*}{CheXpert}
& Shortcut Group
& +.119
& +.1
& +.3
& -6.2
& -.8
& +.069
& -6.2
& -.8 \\
& Image
& +.018
& \textcolor{blue}{+5.2}
& \textcolor{blue}{+1.4}
& \textcolor{blue}{+5.2}
& \textcolor{blue}{+1.4}
& -.123
& -.8
& 0.0 \\

\midrule

\multirow{2}{*}{Waterbirds}
& Shortcut Group
& +.050
& +5.4
& +3.7
& \textcolor{blue}{+5.6}
& \textcolor{blue}{+4.0}
& +.024
& \textcolor{blue}{+5.6}
& \textcolor{blue}{+4.0} \\
& Image
& -.024
& +.6
& +.8
& +.6
& +.8
& -.108
& +2.0
& +.9 \\

\midrule

\multirow{2}{*}{Camelyon17}
& Shortcut Group
& +.015
& \textcolor{blue}{+3.5}
& +1.0
& \textcolor{blue}{+3.5}
& +1.0
& -.047
& +1.8
& 0.0 \\
& Image
& -.077
& +3.2
& \textcolor{blue}{+1.1}
& +3.2
& \textcolor{blue}{+1.1}
& -.191
& -.2
& -.2 \\

\midrule

\multirow{2}{*}{ISIC2019}
& Shortcut Group
& +.078
& \textcolor{blue}{+1.0}
& \textcolor{blue}{+.3}
& -.9
& +.1
& +.052
& -.9
& +.1 \\
& Image
& -.021
& -.5
& -.1
& -.5
& -.1
& -.141
& -1.0
& -.3 \\

\bottomrule
\end{tabular}
}
\end{table*}

%% file: Tables/annotated_localisation.tex
\begin{table*}[hptb]
\centering
\caption{
\textbf{Task contributions are concentrated inside bird/lung regions, while shortcut contributions are primarily in external context on aggregate and across shortcut groups.} For each shortcut group, we form membership-weighted shortcut group prototypes and annotated masks, then report the percentage of the prototype's top-$10\%$ regions overlapping the weighted mask. Values are computed from the prototype maps shown in the corresponding visualisations. Values are based on one ViT model per dataset.}
\begin{subtable}[t]{0.49\textwidth}
\centering
\caption{\textsc{Waterbirds}}
\label{tab:waterbirds_localisation}
\begin{tabular}{@{}lrrr@{}}
\toprule
\textbf{SG}
& \makecell{\textbf{Bird}\\\textbf{area}}
& \makecell{\textbf{Task on}\\\textbf{bird}}
& \makecell{\textbf{Shortcut}\\\textbf{on bird}} \\
\midrule
Dataset-level & 13.4 & 57.5 & 7.9 \\
\midrule
1 & 13.2 & 57.8 & 0.2 \\
2 & 13.5 & 57.0 & 2.1 \\
3 & 14.8 & 56.3 & 0.5 \\
4 & 11.7 & 58.7 & 0.5 \\
5 & 14.2 & 59.4 & 1.0 \\
6 & 14.2 & 60.7 & 0.6 \\
7 & 13.6 & 60.4 & 4.8 \\
8 & 12.8 & 57.1 & 0.9 \\
\bottomrule
\end{tabular}
\end{subtable}
\hfill
\begin{subtable}[t]{0.49\textwidth}
\centering
\caption{\textsc{CheXpert}}
\label{tab:chexpert_localisation}
\begin{tabular}{@{}lrrr@{}}
\toprule
\textbf{SG}
& \makecell{\textbf{Lung}\\\textbf{area}}
& \makecell{\textbf{Task inside}\\\textbf{lung}}
& \makecell{\textbf{Shortcut}\\\textbf{inside lung}} \\
\midrule
Dataset-level & 24.1 & 39.3 & 10.0 \\
\midrule
1 & 23.5 & 35.5 & 0.4 \\
2 & 24.7 & 29.2 & 6.2 \\
3 & 25.5 & 27.1 & 1.5 \\
4 & 23.3 & 43.3 & 0.7 \\
5 & 23.2 & 55.0 & 0.5 \\
6 & 23.0 & 47.7 & 1.5 \\
7 & 23.9 & 36.4 & 4.1 \\
8 & 25.2 & 26.6 & 4.0 \\
\bottomrule
\end{tabular}
\end{subtable}

\medskip

\begin{minipage}{\textwidth}
\centering
\includegraphics[width=\linewidth]{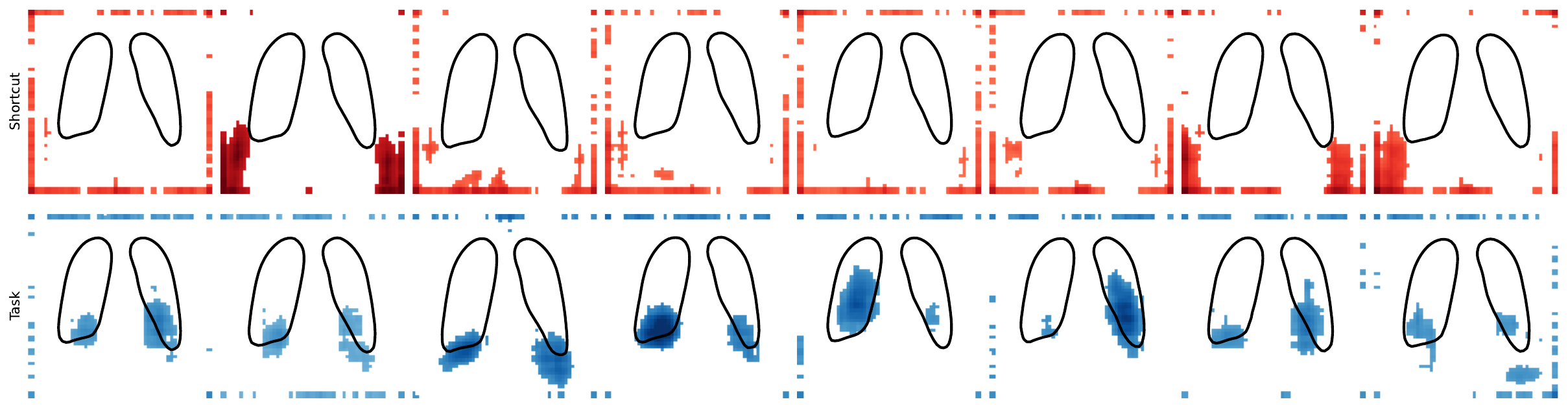}
\captionof{figure}{\textbf{Across shortcut groups, shortcut contribution regions are non-lung whereas task contribution regions are generally overlapping with lung regions across different shortcut groups.} Each column corresponds to one of the shortcut groups. \textsc{CheXmask} atlas overlaid on shortcut group-level prototypes using ViT.}
\label{fig:chexmask}
\end{minipage}

\end{table*}